%% file: thesis.tex
\documentclass[12pt,a4paper]{report}

\usepackage{colortbl}
\usepackage[dvips]{graphicx}
\usepackage{fancyhdr}
\usepackage{latexsym}
\usepackage{setspace}
\usepackage{lingmacros}
\usepackage{pstricks,pst-node,pst-tree}
\usepackage{natbib}
\usepackage[safe]{tipa}

\usepackage{mvz}

\usepackage{makeidx} % remove showidx to remove index info in margins
\makeindex

\begin{document}

% BEGIN INTRO
\include{cover}       % This is the front page
  \pagestyle{plain}
  \pagenumbering{roman}
  \setcounter{page}{1}
\include{abstract}    % Abstract
\include{acknowledge} % Acknowledgements
\include{declare}     % (optional) - declarations of previous papers
\tableofcontents
\listoffigures
\listoftables
\newpage
% END INTRO

% Set up page numbering
\pagenumbering{arabic}
% Preface is still without fancyheadings
\include{preface}

\pagestyle{fancy}
\fancyhead[L]{\textsl{\chaptername\ \thechapter}}
\fancyhead[R]{\textsl{\leftmark}}
\fancyhead[C]{\thepage}
\fancyfoot[C]{}
\renewcommand{\headrulewidth}{0.4pt}
\renewcommand{\chaptermark}[1]{\markboth{#1}{}}

% BEGIN MAIN PART
\include{introduction}
\include{firstattempt}
\include{overview}
\include{instantiations}
\include{results}
\include{compare}
\include{extensions}
\include{conclusion}
% END MAIN PART

% BEGIN APPENDICES
%\appendix
%\fancyhead[L]{\textsl{Appendix\ \thechapter}}
% END APPENDICES

% BEGIN END PART
\fancyhead[L]{}
\bibliographystyle{apalike}
\bibliography{abl}
\printindex
% END END PART

\end{document}

%% file: cover.tex
%%%%%%%%%%%%%%%%%%%%%%%%%%%%%%%%%%%%%%%%%%%%%%%%%%%%%%%%%%%%%%%%%%%%%%%%%%%%%%%%
%% Menno van Zaanen                                                           %%
%% menno@comp.leeds.ac.uk                                                     %%
%%%%%%%%%%%%%%%%%%%%%%%%%%%%%%%%%%%%%%%%%%%%%%%%%%%%%%%%%%%%%%%%%%%%%%%%%%%%%%%%
%% Filename: cover.tex                                                        %%
%%%%%%%%%%%%%%%%%%%%%%%%%%%%%%%%%%%%%%%%%%%%%%%%%%%%%%%%%%%%%%%%%%%%%%%%%%%%%%%%

\pagestyle{empty}
\begin{center}
\LARGE
{\bf Bootstrapping Structure into Language:
Alignment-Based Learning
} \\
\ \\
\Large
{\bf by} \\
\vspace{0.3in}
\Large
{\it Menno M. van Zaanen} \\
\normalsize
\vspace{1in}
{\bf Submitted in accordance with the requirements \\ for the degree of Doctor
of Philosophy.}\\
\vspace{0.5 in}
\mbox{\includegraphics[width=2.5cm]{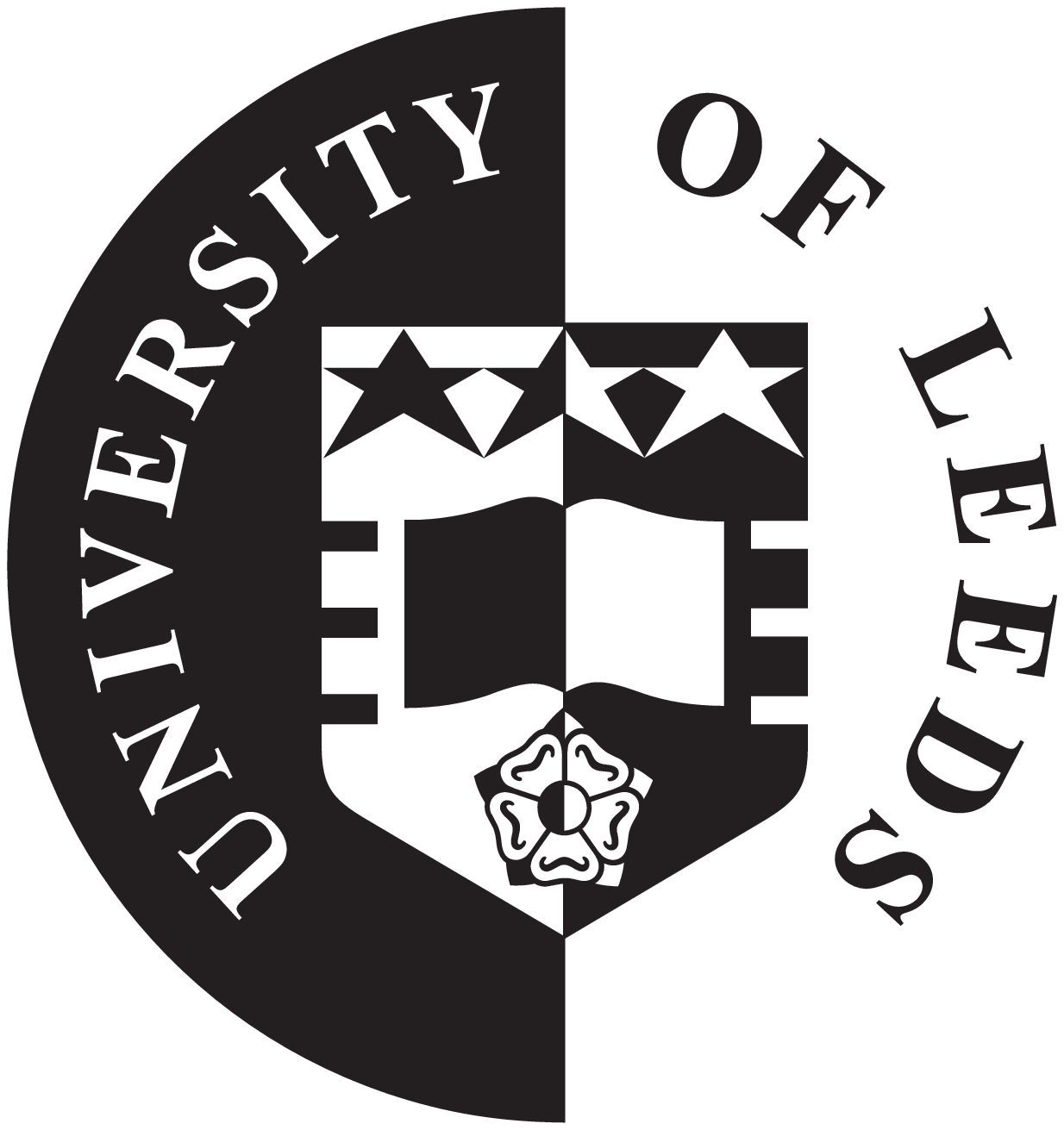} }  \\
{\bf The University of Leeds \\
   School of Computing} \\
\ \\
\ \\
\vspace{0.2in}
{\bf September 2001} \\

\vspace{0.6in}
{\bf The candidate confirms that the work submitted is his own and the
appropriate credit has been given where reference has been made to the
work of others.}
\end{center}

\newpage

%% end of file: cover.tex

%% file: abstract.tex
%%%%%%%%%%%%%%%%%%%%%%%%%%%%%%%%%%%%%%%%%%%%%%%%%%%%%%%%%%%%%%%%%%%%%%%%%%%%%%%%
%% Menno van Zaanen                                                           %%
%% menno@comp.leeds.ac.uk                                                     %%
%%%%%%%%%%%%%%%%%%%%%%%%%%%%%%%%%%%%%%%%%%%%%%%%%%%%%%%%%%%%%%%%%%%%%%%%%%%%%%%%
%% Filename: abstract.tex                                                     %%
%%%%%%%%%%%%%%%%%%%%%%%%%%%%%%%%%%%%%%%%%%%%%%%%%%%%%%%%%%%%%%%%%%%%%%%%%%%%%%%%

%%%%%%%%%%%%%%%%%%%%%%%%%%%%%%%%%%%%%%%%%%%%%%%%%%%%%%%%%%%%%%%%%%%%%%%%%%%%%%%%
%%%%%%%%%%%%%%%%%%%%%%%%%%%%%%%%%%%%%%%%%%%%%%%%%%%%%%%%%%%%%%%%%%%%%%%%%%%%%%%%
%%%%%%%%%%%%%%%%%%%%%%%%%%%%%%%%%%%%%%%%%%%%%%%%%%%%%%%%%%%%%%%%%%%%%%%%%%%%%%%%
\emptythesischapter{Abstract}
{\ldots\ refined and abstract meanings largely grow out of more
concrete meanings.}
{\citet{Bloomfield:L-33}}
\label{ch:abstract}
%%%%%%%%%%%%%%%%%%%%%%%%%%%%%%%%%%%%%%%%%%%%%%%%%%%%%%%%%%%%%%%%%%%%%%%%%%%%%%%%
%%%%%%%%%%%%%%%%%%%%%%%%%%%%%%%%%%%%%%%%%%%%%%%%%%%%%%%%%%%%%%%%%%%%%%%%%%%%%%%%
%%%%%%%%%%%%%%%%%%%%%%%%%%%%%%%%%%%%%%%%%%%%%%%%%%%%%%%%%%%%%%%%%%%%%%%%%%%%%%%%
\index{Bloomfield, L.}%
%approximately 300 words

This thesis introduces a new unsupervised learning framework,
called Alignment-Based Learning, which is based on the alignment
of sentences and Harris's \citeyearpar{Harris:SL-51} notion of
\index{Harris, Z.S.}%
substitutability. Instances of the framework can be applied to an
untagged, unstructured corpus of natural language sentences,
resulting in a labelled, bracketed version of that corpus.

Firstly, the framework aligns all sentences in the corpus in
pairs, resulting in a partition of the sentences consisting of
parts of the sentences that are equal in both sentences and parts
that are unequal. Unequal parts of sentences can be seen as being
substitutable for each other, since substituting one unequal part
for the other results in another valid sentence. The unequal
parts of the sentences are thus considered to be possible
(possibly overlapping) constituents, called hypotheses.

Secondly, the selection learning phase considers all hypotheses
found by the alignment learning phase and selects the best of
these. The hypotheses are selected based on the order in which
they were found, or based on a probabilistic function.

The framework can be extended with a grammar extraction phase.
This extended framework is called \parseabl.  Instead of
returning a structured version of the unstructured input corpus,
like the \abl system, this system also returns a stochastic
context-free or tree substitution grammar.

Different instances of the framework have been tested on the
English ATIS corpus, the Dutch OVIS corpus and the Wall Street
Journal corpus. One of the interesting results, apart from the
encouraging numerical results, is that all instances can (and do)
learn recursive structures.

%% end of file: abstract.tex

%% file: acknowledge.tex
%%%%%%%%%%%%%%%%%%%%%%%%%%%%%%%%%%%%%%%%%%%%%%%%%%%%%%%%%%%%%%%%%%%%%%%%%%%%%%%%
%% Menno van Zaanen                                                           %%
%% menno@comp.leeds.ac.uk                                                     %%
%%%%%%%%%%%%%%%%%%%%%%%%%%%%%%%%%%%%%%%%%%%%%%%%%%%%%%%%%%%%%%%%%%%%%%%%%%%%%%%%
%% Filename: acknowledge.tex                                                  %%
%%%%%%%%%%%%%%%%%%%%%%%%%%%%%%%%%%%%%%%%%%%%%%%%%%%%%%%%%%%%%%%%%%%%%%%%%%%%%%%%

%%%%%%%%%%%%%%%%%%%%%%%%%%%%%%%%%%%%%%%%%%%%%%%%%%%%%%%%%%%%%%%%%%%%%%%%%%%%%%%%
%%%%%%%%%%%%%%%%%%%%%%%%%%%%%%%%%%%%%%%%%%%%%%%%%%%%%%%%%%%%%%%%%%%%%%%%%%%%%%%%
%%%%%%%%%%%%%%%%%%%%%%%%%%%%%%%%%%%%%%%%%%%%%%%%%%%%%%%%%%%%%%%%%%%%%%%%%%%%%%%%
\emptythesischapter{Acknowledgements}
{``Yacc'' owes much to a most stimulating collection of users,\\
 who have goaded me beyond my inclination,\\
 and frequently beyond my ability in their endless search for "one more
 feature".\\
 Their irritating unwillingness to learn how to do things my way\\
 has usually led to my doing things their way;\\
 most of the time, they have been right.}
{\citet[p.\ 376]{Johnson:UPM-79}}
%%%%%%%%%%%%%%%%%%%%%%%%%%%%%%%%%%%%%%%%%%%%%%%%%%%%%%%%%%%%%%%%%%%%%%%%%%%%%%%%
%%%%%%%%%%%%%%%%%%%%%%%%%%%%%%%%%%%%%%%%%%%%%%%%%%%%%%%%%%%%%%%%%%%%%%%%%%%%%%%%
%%%%%%%%%%%%%%%%%%%%%%%%%%%%%%%%%%%%%%%%%%%%%%%%%%%%%%%%%%%%%%%%%%%%%%%%%%%%%%%%
\index{Johnson, S.C.}%

There are many people who helped me (in many different ways) to
start, continue and finish this thesis and the research described
in it. Rens Bod, my supervisor for this project, has been of
great help, revising my writing, solving problems when I was
stuck and giving me back my enthusiasm for the research when
times were difficult (even though he might not even have noticed
doing it).

Especially in the beginning of this research I have had
discussions with many people, especially the \dop group,
consisting of Arjen Poutsma and Lars Hoogweg among others, but
also Mila Groot, Rob Freeman, Alexander Clark, and Gerardo Sierra
(who pointed the Edit-Distance algorithm out to me) need
mentioning.

People at three universities have allowed me to work there. First
of all, the University of Leeds, where Dan Black, Vincent Devin,
and John Elliott made me feel welcome every time. Eric Atwell, my
advisor, was always interested in my current research and could
point out some interesting features and applications I had never
even thought about. At the faculty of Humanities at the
University of Amsterdam, Remko Scha initially showed me the use
of Data-Oriented Parsing in error correction (where it all
started) and Khalil Sima'an helped me with his efficient \dop
parser. With Pieter Adriaans and Marco Vervoort (who generated
the EMILE results for me) from the faculty of Science I had many
interesting discussions when comparing their EMILE system to
\abl.  People at the University of Groningen, especially Wouter
Jansen (who allowed me to use his workspace), Wilbert Heeringa,
Roberto Bolognesi (who both had to cope with me working in their
room), and John Nerbonne (who helped me with many things) made
life a lot easier. 

Whenever I needed programming help or computer power, I could ask
Lo van den Berg, while Kurt Brauchli (working at the department
of Pathology at the University of Basel) explained the ins and
outs of DNA/RNA structures and Rogier Blokland (University of
Groningen) supplied me with the Hungarian example sentences.

In the final stage of the process, Yorick Wilks of the University of
Sheffield accepted the task of external examiner of this thesis.  I
would like to thank him for his useful comments, suggestions, and
discussions.

Finally, I would like to thank my parents, who supported me
throughout my life (including the years I have done this
research) and Tanja Gaustad who helped me with many fruitful
discussions and allowed me to spend time on my research even
though it meant me spending less time with her.

I could not have done it without you all.

%% end of file: acknowledge.tex

%% file: declare.tex
%%%%%%%%%%%%%%%%%%%%%%%%%%%%%%%%%%%%%%%%%%%%%%%%%%%%%%%%%%%%%%%%%%%%%%%%%%%%%%%%
%% Menno van Zaanen                                                           %%
%% menno@comp.leeds.ac.uk                                                     %%
%%%%%%%%%%%%%%%%%%%%%%%%%%%%%%%%%%%%%%%%%%%%%%%%%%%%%%%%%%%%%%%%%%%%%%%%%%%%%%%%
%% Filename: declare.tex                                                      %%
%%%%%%%%%%%%%%%%%%%%%%%%%%%%%%%%%%%%%%%%%%%%%%%%%%%%%%%%%%%%%%%%%%%%%%%%%%%%%%%%

%%%%%%%%%%%%%%%%%%%%%%%%%%%%%%%%%%%%%%%%%%%%%%%%%%%%%%%%%%%%%%%%%%%%%%%%%%%%%%%%
%%%%%%%%%%%%%%%%%%%%%%%%%%%%%%%%%%%%%%%%%%%%%%%%%%%%%%%%%%%%%%%%%%%%%%%%%%%%%%%%
%%%%%%%%%%%%%%%%%%%%%%%%%%%%%%%%%%%%%%%%%%%%%%%%%%%%%%%%%%%%%%%%%%%%%%%%%%%%%%%%
\emptythesischapter{Declarations}
{How do you think he does it?\\
 I don't know!\\
 What makes him so good?}
{The Who (Tommy, Pinball Wizard)}
%%%%%%%%%%%%%%%%%%%%%%%%%%%%%%%%%%%%%%%%%%%%%%%%%%%%%%%%%%%%%%%%%%%%%%%%%%%%%%%%
%%%%%%%%%%%%%%%%%%%%%%%%%%%%%%%%%%%%%%%%%%%%%%%%%%%%%%%%%%%%%%%%%%%%%%%%%%%%%%%%
%%%%%%%%%%%%%%%%%%%%%%%%%%%%%%%%%%%%%%%%%%%%%%%%%%%%%%%%%%%%%%%%%%%%%%%%%%%%%%%%

Parts of the work presented in this thesis have been published in
the following articles:

\nocite{vanZaanen:CLIN99-235}
\nocite{vanZaanen:CLUK99-1}
\nocite{vanZaanen:COLING00-961}
\nocite{vanZaanen:ICML00-1063}
\nocite{vanZaanen:CLUK00-75}
\nocite{vanZaanen:BTG-01}
\nocite{vanZaanen:CTU-01}
\nocite{vanZaanen:BNAIC01-??}
\nocite{vanZaanen:DOP02-??}

\begin{description}
%\addtolength{\itemindent}{-5em}
\item[van Zaanen, 1999a] \hfill\\
\index{van Zaanen, M.M.}%
\newblock Bootstrapping structure using similarity.
\newblock In Monachesi, P., editor, {\em Computational
Linguistics in the Netherlands 1999---Selected Papers from the
Tenth {CLIN} Meeting}, pages 235--245, Utrecht, the Netherlands.
Universteit Utrecht.

\item[van Zaanen, 1999b] \hfill\\
\index{van Zaanen, M.M.}%
\newblock Error correction using {DOP}.
\newblock In \uppercase{d}e Roeck, A., editor, {\em Proceedings
of the Second
  UK Special Interest Group for Computational Linguistics
({CLUK2}) (Second
  Issue)}, pages 1--12, Colchester, UK. University of Essex.

\item[van Zaanen, 2000a] \hfill\\
\index{van Zaanen, M.M.}%
\newblock {ABL}: {A}lignment-{B}ased {L}earning.
\newblock In {\em Proceedings of the 18th International
Conference on
  Computational Linguistics ({COLING}); Saarbr{\"u}cken,
Germany}, pages
  961--967. Association for Computational Linguistics ({ACL}).

\item[van Zaanen, 2000b] \hfill\\
\index{van Zaanen, M.M.}%
\newblock Bootstrapping syntax and recursion using
{A}lignment-{B}ased
  {L}earning.
\newblock In Langley, P., editor, {\em Proceedings of the
Seventeenth
  International Conference on Machine Learning}, pages
1063--1070, Stanford:CA,
  USA. Stanford University.

\item[van Zaanen, 2000c] \hfill\\
\index{van Zaanen, M.M.}%
\newblock Learning structure using alignment based learning.
\newblock In Kilgarriff, A., Pearce, D., and Tiberius, C.,
editors, {\em
  Proceedings of the Third Annual Doctoral Research Colloquium
({CLUK})}, pages
  75--82. Universities of Brighton and Sussex.

\item[van Zaanen, 2001] \hfill\\
\index{van Zaanen, M.M.}%
\newblock Building treebanks using a grammar induction system.
\newblock Technical Report TR2001.06, University of Leeds, Leeds,
UK.

\item[van Zaanen and Adriaans, 2001a] \hfill\\
\index{van Zaanen, M.M.}%
\index{Adriaans, P.W.}%
\newblock Alignment-based learning versus {EMILE}: A comparison.
\newblock In {\em Proceedings of the Belgian-Dutch Conference on
Artificial
  Intelligence ({BNAIC}); Amsterdam, the Netherlands}.
\newblock to be published.

\item[van Zaanen and Adriaans, 2001b] \hfill\\
\index{van Zaanen, M.M.}%
\index{Adriaans, P.W.}%
\newblock Comparing two unsupervised grammar induction systems:
  {A}lignment-{B}ased {L}earning vs. {EMILE}.
\newblock Technical Report TR2001.05, University of Leeds, Leeds,
UK.

\item[van Zaanen, 2002] \hfill\\
\index{van Zaanen, M.M.}%
\newblock Alignment-based learning versus data-oriented parsing.
\newblock In Bod, R., Sima'an, K., and Scha, R., editors, {\em
{D}ata {O}riented {P}arsing}.
  Center for Study of Language and Information (CSLI)
Publications,
  Stanford:CA, USA.
\newblock to be published.
\end{description}

%\newpage

%% end of file: declare.tex

%% file: preface.tex
%%%%%%%%%%%%%%%%%%%%%%%%%%%%%%%%%%%%%%%%%%%%%%%%%%%%%%%%%%%%%%%%%%%%%%%%%%%%%%%%
%% Menno van Zaanen                                                           %%
%% menno@comp.leeds.ac.uk                                                     %%
%%%%%%%%%%%%%%%%%%%%%%%%%%%%%%%%%%%%%%%%%%%%%%%%%%%%%%%%%%%%%%%%%%%%%%%%%%%%%%%%
%% Filename: preface.tex                                                      %%
%%%%%%%%%%%%%%%%%%%%%%%%%%%%%%%%%%%%%%%%%%%%%%%%%%%%%%%%%%%%%%%%%%%%%%%%%%%%%%%%

%%%%%%%%%%%%%%%%%%%%%%%%%%%%%%%%%%%%%%%%%%%%%%%%%%%%%%%%%%%%%%%%%%%%%%%%%%%%%%%%
%%%%%%%%%%%%%%%%%%%%%%%%%%%%%%%%%%%%%%%%%%%%%%%%%%%%%%%%%%%%%%%%%%%%%%%%%%%%%%%%
%%%%%%%%%%%%%%%%%%%%%%%%%%%%%%%%%%%%%%%%%%%%%%%%%%%%%%%%%%%%%%%%%%%%%%%%%%%%%%%%
\emptythesischapter{Preface}
{The White Rabbit put on his spectacles.\\
`Where shall I begin, please your Majesty?' he asked.\\
`Begin at the beginning,' the King said, very gravely,\\
`and go on till you come to the end: then stop.'}
{\citet[p.\ 109]{Carroll:CIW-82-17}}
%%%%%%%%%%%%%%%%%%%%%%%%%%%%%%%%%%%%%%%%%%%%%%%%%%%%%%%%%%%%%%%%%%%%%%%%%%%%%%%%
%%%%%%%%%%%%%%%%%%%%%%%%%%%%%%%%%%%%%%%%%%%%%%%%%%%%%%%%%%%%%%%%%%%%%%%%%%%%%%%%
%%%%%%%%%%%%%%%%%%%%%%%%%%%%%%%%%%%%%%%%%%%%%%%%%%%%%%%%%%%%%%%%%%%%%%%%%%%%%%%%
\index{Carroll, L.}%
\vspace{-.3cm}
Some years ago, I had a meeting with Remko Scha at the University
of Amsterdam. He mentioned an interesting research topic where
the Data-Oriented Parsing (\dop) framework \citep{Bod:ELS-95} was
\index{Bod, R.}%
to be used in error correction. During the research for my
Master's thesis at the Vrije Universiteit, I implemented an
error correction system based on \dop and incorporated it in a C
compiler \citep{vanZaanen:ECD-97}.
\index{van Zaanen, M.M.}%

When I was nearly done with writing my thesis, Rens Bod contacted
me and asked me if I would be interested in a PhD position at the
University of Leeds, which would possibly allow me to continue
the research I was doing. A short while later, I got accepted and
the work, I have done there, has led to the thesis now lying in
front of you.

The original idea for the research described in this thesis was
to transfer the error correction system from the field of
computer science back into computational linguistics (where the
\dop framework originally came from).

The main disadvantage of using such an error correcting system to
correct errors in natural language sentences, however, is that
the underlying grammar should be fixed beforehand; the errors are
corrected according to the grammar. In practice, with natural
language, it is often the case that the grammar itself is
incorrect or incomplete and the sentence \emph{is} correct.

The topic of the research shifted from building an error
correction system to building a grammar correction system.
Taking things to extremes, the final system should be able to
bootstrap a grammar from scratch. I started wondering how
people are able to learn grammars and I wondered how I could
get a computer to do the same.\footnote{It must be absolutely
clear that the system in no way claims to be cognitively
modelling human language learning, even though some aspects may
be cognitively plausible.}

The result of this process is described in the rest of this
thesis.

\nocite{Knuth:TT-86,Lamport:LADPS-94,Goossens:LC-94,Goossens:LGC-97}
\nocite{Stroustrup:CPL-97,Lippman:CP-91,Kernighan:CPL-88,Kay:CPU-89}

%% end of file: preface.tex

%% file: introduction.tex
%%%%%%%%%%%%%%%%%%%%%%%%%%%%%%%%%%%%%%%%%%%%%%%%%%%%%%%%%%%%%%%%%%%%%%%%%%%%%%%%
%% Menno van Zaanen                                                           %%
%% menno@comp.leeds.ac.uk                                                     %%
%%%%%%%%%%%%%%%%%%%%%%%%%%%%%%%%%%%%%%%%%%%%%%%%%%%%%%%%%%%%%%%%%%%%%%%%%%%%%%%%
%% Filename: introduction.tex                                                 %%
%%%%%%%%%%%%%%%%%%%%%%%%%%%%%%%%%%%%%%%%%%%%%%%%%%%%%%%%%%%%%%%%%%%%%%%%%%%%%%%%

%%%%%%%%%%%%%%%%%%%%%%%%%%%%%%%%%%%%%%%%%%%%%%%%%%%%%%%%%%%%%%%%%%%%%%%%%%%%%%%%
%%%%%%%%%%%%%%%%%%%%%%%%%%%%%%%%%%%%%%%%%%%%%%%%%%%%%%%%%%%%%%%%%%%%%%%%%%%%%%%%
%%%%%%%%%%%%%%%%%%%%%%%%%%%%%%%%%%%%%%%%%%%%%%%%%%%%%%%%%%%%%%%%%%%%%%%%%%%%%%%%
\thesischapter{Introduction}
{Are you ready for a new sensation?}
{David Lee Roth (Eat 'em and smile)}
%%%%%%%%%%%%%%%%%%%%%%%%%%%%%%%%%%%%%%%%%%%%%%%%%%%%%%%%%%%%%%%%%%%%%%%%%%%%%%%%
%%%%%%%%%%%%%%%%%%%%%%%%%%%%%%%%%%%%%%%%%%%%%%%%%%%%%%%%%%%%%%%%%%%%%%%%%%%%%%%%
%%%%%%%%%%%%%%%%%%%%%%%%%%%%%%%%%%%%%%%%%%%%%%%%%%%%%%%%%%%%%%%%%%%%%%%%%%%%%%%%

The increase of computer storage and processing power has opened
the way for new, more resource intensive linguistic applications
that used to be unreachable. The trend in increase of resources
also creates new uses for structured corpora or treebanks.
\index{treebank}%
\index{treebank!increase of use}%
\index{structured corpus}%
\index{structured corpus!increase of use}%
In the mean time, wider availability of treebanks will account
for new types of applications. These new applications can already
be found in several fields, for example:
\begin{itemize}
\item Natural language parsing
\index{natural language parsing}%
\citep{Allen:NLU-95,Bod:BG-98,Charniak:NCAI97-598,Jurafsky:SLP-00},
\index{Allen, J.}%
\index{Bod, R.}%
\index{Charniak, E.}%
\index{Jurafsky, D.}%
\index{Martin, J.H.}%
\item Evaluation of natural language grammars
\index{evaluation!grammar}%
\index{grammar!evaluation}%
\citep{Black:SNL91-306,Sampson:00-5-53},
\index{Black, E.}%
\index{Abney, S.}%
\index{Flickinger, D.}%
\index{Gdaniec, C.}%
\index{Grishman, R.}%
\index{Harrison, P.}%
\index{Hindle, D.}%
\index{Ingria, R.}%
\index{Jelinek, F.}%
\index{Klavans, J.}%
\index{Liberman, M.}%
\index{Marcus, M.}%
\index{Roukos, S.}%
\index{Santorini, B.}%
\index{Strzalkowski, T.}%
\item Machine translation
\index{machine translation}%
\citep{Poutsma:DOT-00,Sadler:COLING90-449,Way:99-11-??},
\index{Poutsma, A.}%
\index{Sadler, V.}%
\index{Vendelmands, R.}%
\index{Way, A.}%
\item Investigating unknown scripts \citep{Knight:ULNLP99-37}.
\index{Knight, K.}%
\index{Yamada, K.}%
\index{unknown script}%
\end{itemize}

Even though the applications rely heavily on the availability of
treebanks, in practice it is often hard to find one that is
suitable for the specific task. The main reason for this is
that building treebanks is costly.

Language learning systems\footnote{Language learning systems are
sometimes called structure bootstrapping, grammar
induction, or grammar learning systems.  These names are used
interchangeably throughout this thesis.  However, when the
emphasis is on \emph{grammar} learning, bootstrapping, or
induction, the system should at least return a grammar as output,
in contrast to language learning systems which only need to build
a structured corpus.}
\index{grammar induction system}%
\index{language learning system}%
\index{grammar learning system}%
may help to solve the
above mentioned problem. These systems structure plain sentences
(without the use of a grammar) or learn grammars, which can then
be used to parse sentences.  Parsing indicates possible
\index{parsing}%
structures or completely structures the corpus, making annotation
less time and expertise intensive and therefore less costly.
Furthermore, language learning systems can be reused on corpora
in different domains.

Because of the many uses and advantages of a language learning
system, one might try to build such a system.  Unsupervised
learning of syntactic structure, however, is one of the hardest
problems in NLP. Although people are adept at learning
grammatical structure, it is difficult to model this process and
therefore it is hard to make a computer learn structure similarly
to humans.

The goal of the algorithms described in this thesis is not to
\index{goal}%
model the human process of language learning, even though the
idea originated from trying to model human language acquisition.
\index{human language acquisition}%
Instead, the algorithm should, given unstructured sentences, find
the best structure. This means that the algorithm should assign a
structure to sentences which is similar to the structure people
would give to those sentences, but not necessarily in the same
way humans do this, nor in the same time or space restrictions.

The rest of this thesis is subdivided as follows. First, in
chapter~\ref{ch:learningalignment}, the underlying ideas of the
system are discussed, followed by a more formal description of
the framework in chapter~\ref{ch:overview}. Next,
chapter~\ref{ch:instantiations} introduces several possible
instantiations of the different phases of the system, and
chapter~\ref{ch:results} contains the results of the
instantiations on different corpora. Since at that point the
entire system has been described and evaluated, it is then
compared against other systems in the field in
chapter~\ref{ch:compare}.  Possible extensions of the basic
systems are described in chapter~\ref{ch:extensions} and
finally, chapter~\ref{ch:conclusion} concludes this thesis.

%% end of file: introduction.tex

%% file: firstattempt.tex
%%%%%%%%%%%%%%%%%%%%%%%%%%%%%%%%%%%%%%%%%%%%%%%%%%%%%%%%%%%%%%%%%%%%%%%%%%%%%%%%
%% Menno van Zaanen                                                           %%
%% menno@comp.leeds.ac.uk                                                     %%
%%%%%%%%%%%%%%%%%%%%%%%%%%%%%%%%%%%%%%%%%%%%%%%%%%%%%%%%%%%%%%%%%%%%%%%%%%%%%%%%
%% Filename: firstattempt.tex                                                 %%
%%%%%%%%%%%%%%%%%%%%%%%%%%%%%%%%%%%%%%%%%%%%%%%%%%%%%%%%%%%%%%%%%%%%%%%%%%%%%%%%

%%%%%%%%%%%%%%%%%%%%%%%%%%%%%%%%%%%%%%%%%%%%%%%%%%%%%%%%%%%%%%%%%%%%%%%%%%%%%%%%
%%%%%%%%%%%%%%%%%%%%%%%%%%%%%%%%%%%%%%%%%%%%%%%%%%%%%%%%%%%%%%%%%%%%%%%%%%%%%%%%
%%%%%%%%%%%%%%%%%%%%%%%%%%%%%%%%%%%%%%%%%%%%%%%%%%%%%%%%%%%%%%%%%%%%%%%%%%%%%%%%
\thesischapter{Learning by Alignment}
{Thus, the fiction of interchangeability is inhumane and inherently wasteful.}
{\citet[p.\ 716]{Stroustrup:CPL-97}}
%%%%%%%%%%%%%%%%%%%%%%%%%%%%%%%%%%%%%%%%%%%%%%%%%%%%%%%%%%%%%%%%%%%%%%%%%%%%%%%%
%%%%%%%%%%%%%%%%%%%%%%%%%%%%%%%%%%%%%%%%%%%%%%%%%%%%%%%%%%%%%%%%%%%%%%%%%%%%%%%%
%%%%%%%%%%%%%%%%%%%%%%%%%%%%%%%%%%%%%%%%%%%%%%%%%%%%%%%%%%%%%%%%%%%%%%%%%%%%%%%%
\label{ch:learningalignment}
\index{Stroustrup, B.}%

\vspace{-1.5em}
This chapter will informally describe step by step how one might
build a system that finds the syntactic structure of a sentence
without knowing the grammar beforehand. First, the main goals are
reviewed, followed by the description of a method for finding
constituents. The method is then extended to be used for
multiple sentences. This method, however, introduces
ambiguities, which will be resolved in the following section.
Finally, some criticism on the applied methods will be discussed.

%%%%%%%%%%%%%%%%%%%%%%%%%%%%%%%%%%%%%%%%%%%%%%%%%%%%%%%%%%%%%%%%%%%%%%%%%%%%%%%%
%%%%%%%%%%%%%%%%%%%%%%%%%%%%%%%%%%%%%%%%%%%%%%%%%%%%%%%%%%%%%%%%%%%%%%%%%%%%%%%%
\thesissection{Goals}
%%%%%%%%%%%%%%%%%%%%%%%%%%%%%%%%%%%%%%%%%%%%%%%%%%%%%%%%%%%%%%%%%%%%%%%%%%%%%%%%
%%%%%%%%%%%%%%%%%%%%%%%%%%%%%%%%%%%%%%%%%%%%%%%%%%%%%%%%%%%%%%%%%%%%%%%%%%%%%%%%
\label{s:goals}
\index{goal}%

It is widely acknowledged that the principal goal in linguistics
is to characterise the set of sentence-meaning pairs. Considering
\index{sentence-meaning pair}%
\index{meaning}%
that linguistics deals with production and perception of
\index{production}%
\index{perception}%
language, using sentence-meaning pairs corresponds to
converting from sentence to meaning in perception and from
meaning to sentence in the production process.

It may be obvious that directly finding these sentence-meaning
mappings is difficult. Taking the (syntactic)
structure\footnote{In this thesis, ``structure'' and ``syntactic
structure'' are used interchangeably.} of the sentence into
\index{syntactic structure}%
\index{structure!syntactic}%
account simplifies this process. A system that finds sentence to
meaning mappings (i.e.\ in the perception process), first analyses
the sentence, generating the structure of the sentence
as an intermediate. Using this structure, the meaning of the
sentence is computed \citep{Montague:FP-74-247}.
\index{Montague, R.}%
\index{sentence!meaning}%
\index{meaning}%

In this thesis, the structure of a sentence is considered to be
in the form of a tree structure. This is not an arbitrary choice.
\index{structure!tree}%
\index{tree!structure}%
Apart from the fact that trees are rather uncontroversial in
linguistics, it is also true in general that ``complex entities
produced by any process of unplanned evolution, \ldots, will have
tree structuring as a matter of statistical necessity''
\citep{Sampson:EE-97}.  Another reason is that ``hierarchies have
\index{Sampson, G.}%
\index{hierarchy}%
a property, \ldots, that greatly simplifies their behaviour''
\citep{Simon:SA-69}, which will be illustrated in
\index{Simon, H.A.}%
section~\ref{s:findconstituent}.

If the sentences conform to a language, described by a known
grammar, several techniques exist to generate the syntactic
structure of these sentences (see for example the (statistical)
parsing techniques in
\index{parsing}%
\citep{Allen:NLU-95,Charniak:SLL-93,Jurafsky:SLP-00}). However,
\index{Allen, J.}%
\index{Charniak, E.}%
\index{Jurafsky, D.}%
\index{Martin, J.H.}%
if the underlying grammar of the sentences is not known, these
techniques cannot be used, since they rely on knowledge of the
grammar.
\nocite{Grune:PT-90}

This thesis will describe a method that generates the syntactic
structure of a sentence when the underlying grammar of the
language (or the set of possible grammars\footnote{The
empiricist/nativist distinction will be discussed in
section~\ref{s:minimum}.}) is \emph{not} known. This type of
system is called a \emph{structure bootstrapping system}.

Following the discussion on the goals of linguistics in general,
let us now concentrate on the goals of a structure bootstrapping
system.  The system described here is developed with two goals in
mind: \emph{usefulness} and \emph{minimum of information}. Both
goals will be described next.

%%%%%%%%%%%%%%%%%%%%%%%%%%%%%%%%%%%%%%%%%%%%%%%%%%%%%%%%%%%%%%%%%%%%%%%%%%%%%%%%
\thesissubsection{Usefulness}
%%%%%%%%%%%%%%%%%%%%%%%%%%%%%%%%%%%%%%%%%%%%%%%%%%%%%%%%%%%%%%%%%%%%%%%%%%%%%%%%

The first goal of a structure bootstrapping system is to find
structure.  However, arbitrary, random, incomplete or incorrect
structure is not very useful. The main goal of the system is to
find \emph{useful} structure.
\index{goal!usefulness}%
\index{usefulness}%

Remember that the goal in linguistics is to find sentence-meaning
\index{sentence-meaning pair}%
pairs, using structure as an intermediate.  Useful structure,
therefore, should help us find these pairs. As a (very simple)
example how structure can help, consider
figure~\ref{fig:sentencemeaning}. When a sentence in the left
column has the (partial) structure as shown in the middle column,
the meaning (shown in the right column)\footnote{In this example,
the meaning of a sentence is represented in an overly simple type
of predicate logic, where words in small caps represent the
meaning of the word in the regular font. For example, \mean{like}
is the meaning of the word \sent{likes}.} can be computed by
combining the meaning of the separate parts.\footnote{How the
transition from structure to meaning is found is outside the
scope of this thesis. It will simply be assumed that such a
transition is possible.}

\newpage
Based on the principle of compositionality of meaning
\index{compositionality of meaning}%
\index{meaning!compositionality of}%
\index{meaning}%
\citep{Frege:BS-79}, the meaning of a sentence can be computed by
\index{Frege, G.}%
combining the meaning of its constituents. In general, the
constituent on position \sent{X} may be more than one word. If,
for example, \sent{the old man} is found on position \sent{X},
then the meaning of the sentence would be \mean{likes(oscar, the
old man)}. Of course, this example is too simple to be practical,
but it illustrates how structured sentences can help in finding the
meaning of sentences.

\begin{fig}{Sentence-meaning pairs}
\label{fig:sentencemeaning}
\vspace{1em}
\begin{tabular}
{|l@{\hspace{.5cm}$\Rightarrow$\hspace{.5cm}}
  l@{\hspace{.5cm}$\Rightarrow$\hspace{.5cm}}
  l|}
\hline
\textbf{Sentence} &
\textbf{Structure} &
\textbf{Meaning}\\
\hline\hline
Oscar likes trash &
Oscar likes [trash] &
\mean{like(oscar, trash)}\\
Oscar likes biscuits &
Oscar likes [biscuits] &
\mean{like(oscar, biscuits)}\\
Oscar likes \sent{X} &
Oscar likes [\sent{X}] &
\mean{like(oscar, \sent{X})}\\
\hline
\end{tabular}
\end{fig}

Usefulness can be tested based on a predefined, manually
structured corpus, such as the Penn Treebank
\citep{Marcus:93-19-313}, or the Susanne Corpus
\index{Marcus, M.}%
\index{Santorini, B.}%
\index{Marcinkiewicz, M.}%
\citep{Sampson:EC-95}. The structures of the sentences in such a
\index{Sampson, G.}%
corpus are considered completely correct (i.e.\ each tree
corresponds to the structure as it was perceived for the
particular sentence uttered in a certain context). The structure
learned by the structure bootstrapping system is then compared
against this \emph{true} structure. First, plain sentences are
extracted from a given structured corpus. These plain sentences
are the input of the structure bootstrapping system. The output
(structured sentences) can be compared to the structured
sentences of the original corpus and the completeness (recall)
\index{goal!recall}%
\index{recall}%
and correctness (precision)
\index{goal!precision}%
\index{precision}%
of the learned structure can be computed. Details of this
evaluation method can be found in section~\ref{s:metrics}.

%%%%%%%%%%%%%%%%%%%%%%%%%%%%%%%%%%%%%%%%%%%%%%%%%%%%%%%%%%%%%%%%%%%%%%%%%%%%%%%%
\thesissubsection{Minimum of information}
%%%%%%%%%%%%%%%%%%%%%%%%%%%%%%%%%%%%%%%%%%%%%%%%%%%%%%%%%%%%%%%%%%%%%%%%%%%%%%%%
\label{s:minimum}
\index{goal!minimum of information}%
\index{minimum of information}%

Structure bootstrapping systems can be grouped (like other
learning methods) into \emph{supervised} and \emph{unsupervised}
\index{supervised system}%
\index{unsupervised system}%
systems. Supervised methods are initialised with structured
sentences, while unsupervised methods only get to see plain
sentences. In practice, supervised methods outperform
unsupervised methods, since they can adapt their output based on
the structured examples in the initialisation phase whereas
unsupervised methods cannot.

Even though in general the performance of unsupervised methods is
\index{supervised system!performance of}%
\index{unsupervised system!performance of}%
less than that of supervised methods, it \emph{is} worthwhile to
investigate unsupervised grammar learning methods. Supervised
methods need structured sentences to initialise, but ``the costs
of annotation are prohibitively time and expertise intensive, and
the resulting corpora may be too susceptible to restriction to a
particular domain, application, or genre''
\citep{Kehler:ULNLP99-0}. Thus annotated corpora may not always
\index{Kehler, A.}%
\index{Stolcke, A.}%
be available for a language.  In contrast, unsupervised methods
do not need these structured sentences.

The second goal of the bootstrapping system can be described as
learning using a \emph{minimum of information}. The system should
\index{goal!minimum of information}%
\index{minimum of information}%
try to minimise the amount of information it needs to learn
structure. Supervised systems receive structured examples, which
contain more information than their unstructured counterparts, so
the second goal restricts the system described in this thesis
from being supervised.

In general, unsupervised systems may still use additional
information. This additional information may be for example in
the form of a lexicon, part-of-speech tags
\citep{Klein:ACL/CNLL01-113,Pereira:ACL92-128}, many adjustable
\index{Klein, D.}%
\index{Manning, C.D.}%
\index{Pereira, F.}%
\index{Schabes, Y.}%
language dependent settings in the system
\citep{Adriaans:LLC-92,Vervoort:GWG-00} or a combination of
\index{Adriaans, P.W.}%
\index{Vervoort, M.R.}%
language dependent information sources \citep{Osborne:LUB-94}.
\index{Osborne, M.}%

However, since the goal of the system described here is to use a
minimum of information, the system must refrain from using extra
information. The only language dependent information the system
may use is a corpus of plain sentences (for example in the form
of transcribed acoustic utterances). Using this information it
outputs the same corpus augmented with structure (or a compact
representation of this structure, for example in the form of a
grammar).

The advantages of using a minimum of information are legion.
Since no language specific knowledge is needed, it can be used on
\index{information!language specific}%
languages for which no structured corpora or dictionaries exist.
It can even be used on unknown languages. Furthermore, it does
not need extensive tuning (since the system does not have many
adjustable settings).

By assuming the goal of minimum of information, the learning
method should be classified as an empiricist approach. According
\index{empiricism}%
\index{approach!empiricist}%
to \citet[pp.\ 47--48]{Chomsky:ATS-65}:
\index{Chomsky, N.}%
\begin{quote}
The empiricist approach has assumed that the structure of the
acquisition device is limited to certain elementary ``peripheral
processing mechanisms''\ldots Beyond this, it assumes that the
device has certain analytical data-processing mechanisms or
inductive principles of a very elementary sort, \ldots
\end{quote}

\newpage
In contrast to the empiricist approach, there is the nativist (or
rationalist) approach:
\index{nativism}%
\index{approach!nativist}%
\index{rationalism}%
\index{approach!rationalist}%
\index{rationalism}%
\begin{quote}
The rationalist approach holds that beyond the peripheral
processing mechanisms, there are innate ideas and principles of
various kinds that determine the form of the acquired knowledge
in what may be a rather restricted and highly organised way.
\end{quote}

The nativist approach assumes innate ideas and principles (for
\index{innateness}%
example an innate universal grammar describing all possible
\index{universal grammar}%
languages). The empiricist approach, however, is more closely
linked to the idea of minimum of information.

Note that instead of refuting the nativist approach, the work
described in this thesis is an attempt to show how much can be
learned using an empiricist approach.

Now that the goals of the system are set, a first attempt will be
made to meet these goals. The rest of the chapter develops a
method that adheres to the first goal (usefulness), while keeping
the second goal in mind.

%%%%%%%%%%%%%%%%%%%%%%%%%%%%%%%%%%%%%%%%%%%%%%%%%%%%%%%%%%%%%%%%%%%%%%%%%%%%%%%%
%%%%%%%%%%%%%%%%%%%%%%%%%%%%%%%%%%%%%%%%%%%%%%%%%%%%%%%%%%%%%%%%%%%%%%%%%%%%%%%%
\thesissection{Finding constituents}
%%%%%%%%%%%%%%%%%%%%%%%%%%%%%%%%%%%%%%%%%%%%%%%%%%%%%%%%%%%%%%%%%%%%%%%%%%%%%%%%
%%%%%%%%%%%%%%%%%%%%%%%%%%%%%%%%%%%%%%%%%%%%%%%%%%%%%%%%%%%%%%%%%%%%%%%%%%%%%%%%
\label{s:findconstituent}
\index{constituent!find}%

Starting with the first goal of the system, usefulness,
constituents need to be found in unstructured text. To get an
idea of the exact problem, imagine seeing text in an unknown
language (unknown to you). How would you try to find out which
\index{language!unknown}%
\index{unknown language}%
words belong to the same syntactic type (for example, nouns) or
which words group together to form, for example, a noun phrase?
(Remember that the goal is \emph{not} to find a model of human
\index{human language acquisition}%
language learning. However, thinking about how one searches for
structure might also help in finding a way to automatically
structure text.)

Let us start with the simplest case. If you see one sentence in
an unknown language then what can you conclude from that? (Try
for example sentence~\ref{unknown1}.) If you do not know anything
about the language, it is very hard if not
impossible\footnote{Using for example universal rules or
\index{universal rules}%
expected distributions of word classes, it may be possible to
\index{distribution of word class}%
find some structure in one plain sentence.} to say anything about
the structure of the sentence (but you can conclude that it
\emph{is} a sentence).

However, if two sentences are available, it is possible to find
parts of the sentences that are the same in both and parts that
are not (provided that some words are the same and some words are
different in both sentences). The comparison of two sentences
falls into one of three different categories:
\begin{enumerate}
\item All words in the two sentences are the same (and so is the
order of the words).\footnote{Previous
publications mentioned ``similar'' and ``dissimilar'' instead of
``equal'' and ``unequal'' in this context. However, apart from
section~\ref{s:weakening} where the exact match is weakened,
sentences are only considered equal if the words in the two
sentences are exactly the same (and not just similar).}%
\index{sentence!equal}\index{sentence!similar}%
\item The sentences are completely unequal (i.e.\ there is not one
word that can be found in both sentences).
\index{sentence!unequal}\index{sentence!dissimilar}%
\item Some words in the sentences are the same in both and some are not.
\end{enumerate}

It may be clear that the third case is the most interesting one.
The first case does not yield any additional information. It was
already known that the sentence was a sentence. No new knowledge
can be induced from seeing it another time. The second case
illustrates that there are more sentences, but since there is no
relation between the two, it is impossible to extract any further
\index{relation!between sentences}%
useful information.

The sentences contained in the third case give more information.
They show different contexts of the same words. These different
\index{context}%
contexts of the words might help find structure in the two
sentences.

Let us now consider the pairs of sentences in~\ref{unknownsim}
and~\ref{unknowndis}. It is possible to group words that are
equal and words that are unequal in both sentences. The word
groups
\index{word group}%
that are equal in both sentences are underlined.

\eenumsentence{
\toplabel{unknownsim}
\item Bert s\"ut egy \align{kekszet}
\label{unknown1}
\item Ernie eszi a \align{kekszet}
}

\eenumsentence{
\toplabel{unknowndis}
\item \align{Bert s\"ut egy} kekszet
\item \align{Bert s\"ut egy} kal\'acsot
}

These sentences are simple cases of the more complex sentences
\index{sentence!complex}%
where there are more groups of equal and unequal words. The more
complex examples can be seen as concatenations of the simple
cases. Therefore, these simple sentences can be used without loss
\index{sentence!simple}%
of generality. 

Although it is clear which words are the same in both sentences
(and which are not), it is still unclear which parts are
constituents. The Hungarian sentences in~\ref{unknownsim}
translate to the English sentences as shown
in~\ref{transsim}.\footnote{For the sake of the argument, we may
assume that the word order of the two languages is the same, but
this need not necessarily be so, i.e.\ our argument does not
depend on this (language dependent) assumption.} In this case,
it is clear that the underlined word groups (consisting of one
\index{constituent}%
word) should be constituents.\footnote{\nt{X_1} and \nt{X_2}
denote non-terminal types.} \sent{Biscuit} is a constituent, but
\sent{Bert is baking a} and \sent{Ernie is eating the} are not.

\eenumsentence{
\toplabel{transsim}
\item \shortex{4}
{ Bert & s\"ut & egy & [\align{kekszet}]\snt{X_1} }
{ Bert-nom-sg & to bake-pres-3p-sg-indef & a-indef &
biscuit-acc-sg }
{ Bert is baking a [\align{biscuit}]\snt{X_1} }
\item \shortex{4}
{ Ernie & eszi & a & [\align{kekszet}]\snt{X_1} }
{ Ernie-nom-sg & to eat-pres-3p-sg-def & the-def & biscuit-acc-sg }
{ Ernie is eating the [\align{biscuit}]\snt{X_1} }
}

However, if we conclude that equal parts in sentences are always
\index{constituent!equal part}%
\index{equal parts as constituents}%
\index{constituent!unequal part}%
\index{unequal parts as constituents}%
constituents, the sentences in~\ref{unknowndis} will be
structured incorrectly. These sentences are translated as shown
in~\ref{transdis}. In this case, the unequal parts of the
sentences are constituents. \sent{Biscuit} and \sent{cake} are
constituents, but \sent{Bert is baking a} is not.

\eenumsentence{
\toplabel{transdis}
\item \shortex{4}
{ \align{Bert} & \align{s\"ut} & \align{egy} & [kekszet]\snt{X_2} }
{ Bert-nom-sg & to bake-pres-3p-sg-indef & a-indef &
biscuit-acc-sg }
{ \align{Bert is baking a} [biscuit]\snt{X_2} }
\item \shortex{4}
{ \align{Bert} & \align{s\"ut} & \align{egy} & [kal\'acsot]\snt{X_2} }
{ Bert-nom-sg & to bake-pres-3p-sg-indef & a-indef &
cake-acc-sg }
{ \align{Bert is baking a} [cake]\snt{X_2} }
}

Intuitively, choosing constituents by treating \emph{unequal}
parts of sentences as constituents (as shown in the sentences
of~\ref{transdis}) is preferred over constituents of \emph{equal}
parts of sentences (as indicated by the sentences
in~\ref{transsim}).  This will be shown (in two ways) by
considering the underlying grammar.
\index{grammar!underlying}%

\begin{fig}{Constituents induce compression}
\label{fig:compression}
\vspace{1em}
\begin{tabular}
{|l|
  l|
  l|}
\hline
\textbf{Method} & \textbf{Structure}        & \textbf{Grammar}\\
\hline\hline
Equal parts
 & [[Bert s\"ut egy]\snt{X} kekszet]\snt{S}
  & \nt{S}\der \nt{X} kekszet \\
 & [[Bert s\"ut egy]\snt{X} kal\'acsot]\snt{S}
  & \nt{S}\der \nt{X} kal\'acsot \\
 && \nt{X}\der Bert s\"ut egy \\
\hline
Unequal parts
 & [Bert s\"ut egy [kekszet]\snt{X}]\snt{S}
  & \nt{S}\der Bert s\"ut egy \nt{X} \\
 & [Bert s\"ut egy [kal\'acsot]\snt{X}]\snt{S}
  & \nt{X}\der kekszet \\
 && \nt{X}\der kal\'acsot \\
\hline
\end{tabular}
\end{fig}

The main argument for choosing unequal parts of sentences instead
of equal parts of sentences as constituents is that the resulting
grammar is smaller. When unequal parts of sentences are taken to
\index{grammar!smaller}%
be constituents, this results in more compact grammars than when
equal parts of sentences are taken to be constituents. In other
words, the grammar is more compressed.

An example of the stronger compression power of the unequal parts
\index{compression}%
of sentences can be found in figure~\ref{fig:compression}.
If the length of a grammar is defined as the number of symbols in
the grammar, then the length of the first grammar is $10$.
However, the length of the second grammar is $9$.

In general, it is the case that making constituents of unequal
parts of sentences creates a smaller grammar. Imagine aligning
two sentences (and, to keep things simple, assume there is one
equal part and one unequal part in both sentences). The equal
parts of the two sentences is defined as $E$ and the unequal part
of the first sentence is $U_1$ and of the second sentence $U_2$.

\begin{fig}{Grammar size based on equal and unequal parts}
\label{fig:grammarsize}
\vspace{1em}
\begin{tabular}{|l|l|l|}
\hline
\textbf{Method} & \textbf{Grammar}        & \textbf{Size}\\
\hline\hline
Equal parts
& \nt{S}\der\nt{X} $U_1$ & $1+1+|U_1|$ \\
& \nt{S}\der\nt{X} $U_2$ & $1+1+|U_2|$ \\
& \nt{X}\der$E$          & $1+|E|$ \\
\hline
& total                  & $5+|U_1|+|U_2|+|E|$ \\
\hline\hline
Unequal parts
& \nt{S}\der$E$ \nt{X}   & $1+|E|+1$ \\
& \nt{X}\der$U_1$        & $1+|U_1|$ \\
& \nt{X}\der$U_2$        & $1+|U_2|$ \\
\hline
& total                  & $4+|U_1|+|U_2|+|E|$ \\
\hline
\end{tabular}
\end{fig}

Figure~\ref{fig:grammarsize} shows that taking unequal parts of
sentences as constituents create slightly smaller
grammars.\footnote{The order of the non-terminals and the
sentence parts may vary. This does not change the resulting
grammar size.} In the rightmost column, the grammar size is
computed. Each non-terminal counts as $1$ and the sizes of the
equal and unequal parts are also incorporated.

\begin{fig}{Unequal parts of sentences generated by same non-terminal}
\label{fig:derivation}
\begin{multicols}{2}
\psset{nodesep=2pt,levelsep=35pt}
\pstree[linestyle=dashed]{\TR{\nt{S}}}
{\skiplevel{\Tfan[linestyle=solid,fansize=2cm]}
 \pstree[linestyle=solid]{\TR{\nt{X}}}{\TR{\sent{a}}}
}

\pstree[linestyle=dashed]{\TR{\nt{S}}}
{\skiplevel{\Tfan[linestyle=solid,fansize=2cm]}
 \pstree[linestyle=solid]{\TR{\nt{X}}}{\TR{\sent{b}}}
}
\end{multicols}
\end{fig}

For the second argument, assume the sentences are generated from a
context-free grammar.  This means that for each of the two
\index{CFG}%
\index{context-free grammar}%
sentences there is a derivation that leads to that sentence.
Figure~\ref{fig:derivation} shows the derivations of two
\index{derivation}%
sentences, where \sent{a} and \sent{b} are the unequal parts of
the sentences (the rest is the same in both). The idea is now
that the unequal parts of the sentences are both generated from
the same non-terminal, which would indicate that they are
constituents of the same type. Note that this is not necessarily
the case, as is shown in the example sentences in~\ref{transsim}.
However, changing one of the sentences in~\ref{transsim} to the
other one, would require several non-terminals to have different
yields\footnote{If changing the sentences would only require one
non-terminal to be changed, the situation in
figure~\ref{fig:derivation} arises again.}, whereas taking
unequal parts of sentences as constituents means that only one
non-terminal needs to have a different yield.

These two arguments indicate a preference for the method that
takes unequal parts of sentences as constituents. Additionally,
the idea closely resembles the linguistically motivated and
language independent notion of
\emph{substitutability}.
\index{substitutability}%
\index{segment!substitutable}%
\citet[p.\ 30]{Harris:SL-51}
\index{Harris, Z.S.}%
describes freely substitutable segments as follows:
\begin{quote}
\label{substitutability}
If segments are \emph{freely substitutable} for each other they are
descriptively equivalent, in that everything we henceforth say about one of
them will be equally applicable to the others.
\end{quote}

Harris continues by describing how substitutable segments can be found:
\index{find substitutable segment}%
\begin{quote}
We take an utterance whose segments are recorded as \textsl{DEF}. We now
construct an utterance composed of the segments \textsl{DA'F}, where
\textsl{A'} is a repetition of a segment \textsl{A} in an utterance which we
had represented as \textsl{ABC}. If our informant accepts \textsl{DA'F} as a
repetition of \textsl{DEF}, or if we hear an informant say \textsl{DA'F}, and
if we are similarly able to obtain \textsl{E'BC} (\textsl{E'} being a
repetition of \textsl{E}) as equivalent to \textsl{ABC}, then we say that
\textsl{A} and \textsl{E} (and \textsl{A'} and \textsl{E'}) are mutually
substitutable (or equivalent), as free variants of each other, and write
\textsl{A}=\textsl{E}. If we fail in these tests, we say that \textsl{A} is
different from \textsl{E} and not substitutable for it.
\end{quote}

When Harris's test is simplified or reduced (removing the need for
repetitions), the following test remains: ``If the occurrences \textsl{DEF},
\textsl{DAF}, \textsl{ABC} and \textsl{EBC} are found, \textsl{A} and
\textsl{E} are substitutable.''

The simplified test is instantiated with constituents as
segments.\footnote{\citet{Harris:SL-51} considers segments
\index{Harris, Z.S.}%
mainly as parts of words, i.e.\ phonemes and morphemes.} We
conclude that if constituents
\index{constituent!find}%
are found in the same context, they are substitutable and thus of
the same type. This is equivalent to finding constituents as is
shown in the sentences in~\ref{transdis}.

The test for finding constituents is intuitively assumed correct
(however, see section~\ref{s:criticism} for criticism of this
approach). A constituent of a certain type can be replaced by
another constituent of the same type, while still retaining a
valid sentence. Therefore, if two sentences are the same except
for a certain part, it might be the case that these two sentences
were generated by replacing a certain constituent by another one
of the same type.

%%%%%%%%%%%%%%%%%%%%%%%%%%%%%%%%%%%%%%%%%%%%%%%%%%%%%%%%%%%%%%%%%%%%%%%%%%%%%%%%
%%%%%%%%%%%%%%%%%%%%%%%%%%%%%%%%%%%%%%%%%%%%%%%%%%%%%%%%%%%%%%%%%%%%%%%%%%%%%%%%
\thesissection{Multiple sentences}
%%%%%%%%%%%%%%%%%%%%%%%%%%%%%%%%%%%%%%%%%%%%%%%%%%%%%%%%%%%%%%%%%%%%%%%%%%%%%%%%
%%%%%%%%%%%%%%%%%%%%%%%%%%%%%%%%%%%%%%%%%%%%%%%%%%%%%%%%%%%%%%%%%%%%%%%%%%%%%%%%
\label{s:multiplesent}%
\index{multiple sentences}%
\index{sentence!multiple}%

The previous section showed how constituents can be found by looking for
unequal parts of sentences. Of course, one pair of sentences can introduce
more than one constituent-pair
\index{constituent!find}%
\index{constituent!multiple}%
(see for example the sentences in~\ref{moreconst}).

\eenumsentence{
\toplabel{moreconst}
\item {}[Oscar]\snt{X_1} \align{sees the} [large, green]\snt{X_2}
  \align{apple}
\item {}[Cookie monster]\snt{X_1} \align{sees the} [red]\snt{X_2}
  \align{apple}
}

Even so, the system is highly limited if only two sentences can
be used to find constituents. If more sentences can be used
simultaneously, more constituents can be found.

When a third sentence is used to learn, it must be compared to
the first two sentences. For example, using the sentence
\sent{Big Bird sees a pear} it is possible to learn more
structure in the sentences in~\ref{moreconst}. Preferably, the
result should be the structured sentences as shown
in~\ref{threesent}. The ``old'' structure is augmented with the
new structure found by comparing \sent{Big bird sees a pear} to
the two sentences.

\eenumsentence{
\toplabel{threesent}
\item {}[Oscar]\snt{X_1} \align{sees} [the
[large, green]\snt{X_2} apple]\snt{X_3}
\label{threesent:a}
\item {}[Cookie monster]\snt{X_1} \align{sees} [the [red]\snt{X_2}
apple]\snt{X_3}
\label{threesent:b}
\item {}[Big Bird]\snt{X_1} \align{sees} [a pear]\snt{X_3}
\label{threesent:c}
}

Learning structure by finding the unequal parts of sentences is
easy if no structure is present yet. However, it is unclear what
exactly the system would do when some constituents are already
present in the sentence. The easiest way of handling this is to
compare the plain sentences (temporarily forgetting any structure
already present). When some structure is found it is then added
to the already existing structure.

In example~\ref{threesent}, sentences~\ref{threesent:a}
and~\ref{threesent:b} are compared first (as shown
in~\ref{moreconst}). The plain sentence~\ref{threesent:c} is then
compared against the plain sentence of~\ref{threesent:a}. This
adds the constituents with type \nt{X_3} to
sentences~\ref{threesent:a} and~\ref{threesent:c} and the
constituent of type \nt{X_1} to sentence~\ref{threesent:c}.

Sentence~\ref{threesent:c} then already has the structure as
shown. However, it is still compared against
sentence~\ref{threesent:b}, since some more structure might be
induced. Indeed, when comparing the plain
sentences~\ref{threesent:b} and~\ref{threesent:c},
sentence~\ref{threesent:b} also receives the constituents with
types \nt{X_3}.

It might happen that a new constituent is added to the structure,
that overlaps with
\index{constituent!overlap}%
a constituent that already existed. As an example, consider the
sentences in~\ref{overthree}. When the first sentence is compared
against the second, \sent{the apple} is equal in both sentences,
so the \nt{X_1} constituents are introduced.\footnote{Whenever
necessary, opening brackets are also annotated with their
non-terminals.} At a later time, the second sentence is compared
against the third sentence. This time, \sent{Big Bird} is equal
in both sentences (indicated by a double underlining). This
introduces the \nt{X_2} constituents. At that point, the second
sentence contains two constituents that overlap.

\eenumsentence{
\toplabel{overthree}
\item {}[Oscar sees]\snt{X_1} \align{the apple}
\label{overthree:a}
\item {}\brover{4.35cm}[\snt{X_1}\dalign{Big Bird}
\brunder{4.5cm}[\snt{X_2}throws]\snt{X_1} \align{the apple}]\snt{X_2}
\label{overthree:b}
\item \dalign{Big Bird} [walks]\snt{X_2}
\label{overthree:c}
}

As the example shows, the algorithm can generate overlapping
constituents. This happens when an incorrect constituent is
introduced. Since the structure is assumed to be generated from a
context-free grammar, the structure of sentence~\ref{overthree:b}
is clearly incorrect.\footnote{Overlapping constituents can also
be seen as a richly structured version of the sentences.  From
this viewpoint, the assumed underlying context-free grammar
restricts the output. It delimits the structure of the sentences
into a version that could have been generated by the less
powerful context-free type of grammar.}
\index{context-free grammar}%
\index{CFG}%

%%%%%%%%%%%%%%%%%%%%%%%%%%%%%%%%%%%%%%%%%%%%%%%%%%%%%%%%%%%%%%%%%%%%%%%%%%%%%%%%
%%%%%%%%%%%%%%%%%%%%%%%%%%%%%%%%%%%%%%%%%%%%%%%%%%%%%%%%%%%%%%%%%%%%%%%%%%%%%%%%
\thesissection{Removing overlapping constituents}
%%%%%%%%%%%%%%%%%%%%%%%%%%%%%%%%%%%%%%%%%%%%%%%%%%%%%%%%%%%%%%%%%%%%%%%%%%%%%%%%
%%%%%%%%%%%%%%%%%%%%%%%%%%%%%%%%%%%%%%%%%%%%%%%%%%%%%%%%%%%%%%%%%%%%%%%%%%%%%%%%
\label{s:removeoverlapping}
\index{constituent!overlap}%

If the system is trying to learn structure that can be generated
by a context-free (or mildly context-sensitive) grammar,
\index{context-free grammar}%
\index{CFG}%
\index{grammar!underlying}%
overlaps should never occur within one tree structure. However,
the system so far \emph{can} (and does) introduce overlapping
constituents.

Assuming that the underlying grammar is context-free is not an
arbitrary choice. To evaluate learning systems (amongst others),
a structured corpus is taken to compute the recall and precision.
\index{recall}%
\index{precision}%
Most structured corpora are built on a context-free (or weakly
\index{treebank}%
\index{structured corpus}%
context-sensitive) grammar and thus do not contain overlapping
constituents within a sentence.

The system described so far needs to be modified so that the
final structured sentences do not have overlapping constituents.
This will be the second phase in the system. There are many
different, possible ways of accomplishing this, but as a first
instantiation, the system will make the (very) simple assumption
\index{constituent!correct}%
that constituents learned earlier are always correct.\footnote{It
must be completely clear that assuming that older constituents
are correct is not a feature of the general framework (which will
be described in the next chapter), but merely a particular
instantiation of this phase.} This means that when a constituent
is introduced that overlaps with an already existing constituent,
the newer constituent is considered incorrect. If this happens,
the new constituent is not introduced into the structure. This
will make sure that no overlapping constituents are stored.
Other instantiations, based on a probabilistic evaluation method,
will be explained in chapter~\ref{ch:instantiations}.

The next section will describe overall problems of the method
described so far (including the problems of this approach of
solving overlapping constituents). Improved methods that remove
overlapping constituents will be discussed in
section~\vref{s:selectioninst}.

%%%%%%%%%%%%%%%%%%%%%%%%%%%%%%%%%%%%%%%%%%%%%%%%%%%%%%%%%%%%%%%%%%%%%%%%%%%%%%%%
%%%%%%%%%%%%%%%%%%%%%%%%%%%%%%%%%%%%%%%%%%%%%%%%%%%%%%%%%%%%%%%%%%%%%%%%%%%%%%%%
\thesissection{Problems with the approach}
%%%%%%%%%%%%%%%%%%%%%%%%%%%%%%%%%%%%%%%%%%%%%%%%%%%%%%%%%%%%%%%%%%%%%%%%%%%%%%%%
%%%%%%%%%%%%%%%%%%%%%%%%%%%%%%%%%%%%%%%%%%%%%%%%%%%%%%%%%%%%%%%%%%%%%%%%%%%%%%%%

There seem to be two problems with the structure bootstrapping
approach as described so far. First, removing overlapping
constituents as described in the previous section, although
solving the problem, is clearly incorrect. Second, the underlying
idea of the method, Harris's notion of substitutability, has been
heavily criticised. Both problems will be described in some
detail next.

%%%%%%%%%%%%%%%%%%%%%%%%%%%%%%%%%%%%%%%%%%%%%%%%%%%%%%%%%%%%%%%%%%%%%%%%%%%%%%%%
\thesissubsection{Incorrectly removing overlapping constituents}
%%%%%%%%%%%%%%%%%%%%%%%%%%%%%%%%%%%%%%%%%%%%%%%%%%%%%%%%%%%%%%%%%%%%%%%%%%%%%%%%

The method of finding constituents as described in
section~\vref{s:findconstituent} may, at some point, find
overlapping constituents. Since overlapping constituents are
unwanted, as discussed above, the system takes the older
constituents as correct, removing the newer constituents.

Even though there is evidence that ``analyses that a person has
experienced before are preferred to analyses that must be newly
constructed'' \citep[p.\ 3]{Bod:BG-98}, it is clear that applying
\index{Bod, R.}%
this idea directly will generate incorrect results. It is easily
imagined that when the order of the sentences is different, the
final results will be different, since different constituents are
seen earlier.

Let us reflect on why exactly constituents are removed. If
overlapping constituents are unwanted, then clearly the method of
finding constituents, which introduces these overlapping
constituents, is incorrect.

Before discarding the work done so far, it may be helpful to
reconsider the used terminology. Another way of looking at the
approach is to say that the method which finds constituents does
not really introduce \emph{constituents}, but instead it
introduces \emph{hypotheses} about constituents.

``Finding constituents'' really builds a hypothesis space, where
possible constituents in the sentences are stored. ``Removing
overlapping constituents'' searches this hypothesis space,
removing hypotheses until the best remain.

Clearly, a better system would keep (i.e.\ not remove) hypotheses
if there is more evidence for them. In this case, evidence can be
defined in terms of frequency. Hypotheses that have a higher
frequency are more likely to be correct. Older hypotheses can now
be overruled by newer hypotheses if the latter have a higher
frequency. Section~\vref{s:selectioninst} contains two
hypothesis selection methods based on this idea. Using
probabilities to select hypotheses makes the system described in
this thesis a Bayesian learning method. The goal is to maximise
the probability of a set of (non-overlapping) hypotheses for a
sentence given that sentence.

Selecting constituents based on their frequency is an intuitively
correct solution. But, apart from that, the notion of selecting
structure based on frequencies is uncontroversial in psychology.
``More frequent analyses are preferred to less frequent ones''
\citep[p.\ 3]{Bod:BG-98}.
\index{Bod, R.}%

%%%%%%%%%%%%%%%%%%%%%%%%%%%%%%%%%%%%%%%%%%%%%%%%%%%%%%%%%%%%%%%%%%%%%%%%%%%%%%%%
\thesissubsection{Criticism on Harris's notion of substitutability}
%%%%%%%%%%%%%%%%%%%%%%%%%%%%%%%%%%%%%%%%%%%%%%%%%%%%%%%%%%%%%%%%%%%%%%%%%%%%%%%%
\label{s:criticism}

Harris's notion of substitutability has been heavily criticised.
However, most criticism is similar in nature to: ``\ldots,
although there is frequent reference in the literature of
linguistics, psychology, and philosophy of language to inductive
procedures, methods of abstraction, analogy and analogical
synthesis, generalisation, and the like, the fundamental
inadequacy of these suggestions is obscured only by their
unclarity''\citep[p.\ 31]{Chomsky:LSLT-75} or ``Structuralist
\index{Chomsky, N.}%
theories, both in the European and American traditions, did
concern themselves with analytic procedures for deriving aspects
of grammar from data, as in the procedural theories of \ldots\@
Zellig Harris, \ldots\@ primarily in the areas of phonology and
morphology. The procedures suggested were seriously inadequate
\ldots''\citep[p.\ 7]{Chomsky:KL-86}.  There are only a few
\index{Chomsky, N.}%
places where Harris's notion of substitutability is really
questioned.  See \citep{Redington:98-22-425} for a discussion of
\index{Redington, M.}%
\index{Chater, N.}%
\index{Finch, S.}%
the problems described in~\citep{Pinker:LLLD-84}.
\index{Pinker, S.}%

The next two sections will discuss serious objections by
\citeauthor{Chomsky:LSLT-75} and \citeauthor{Pinker:LI-94}
respectively.

%%%%%%%%%%%%%%%%%%%%%%%%%%%%%%%%%%%%%%%%%%%%%%%%%%%%%%%%%%%%%%%%%%%%%%%%%%%%%%%%
\thesissubsubsection{Chomsky's objections to substitutability}
%%%%%%%%%%%%%%%%%%%%%%%%%%%%%%%%%%%%%%%%%%%%%%%%%%%%%%%%%%%%%%%%%%%%%%%%%%%%%%%%

\citet[pp.\ 129--145]{Chomsky:LSLT-75} gives a nice overview of
\index{Chomsky, N.}%
problems when the notion of substitutability is used. In his
argumentation, he introduces four problems. Three of these
problems are relevant to the system described in this thesis, so
these will be discussed in some detail.\footnote{The fourth
problem deals with a measure of grammaticality of sentences.}

\citeauthor{Chomsky:LSLT-75} describes the first problem as:
\begin{quote}
In any sample of linguistic material, no two words can be
expected to have exactly the same set of contexts. On the other
hand, many words which should be in different contexts will have
some context in common. \ldots Thus substitution is either too
narrow, if we require complete mutual substitutability for
co-membership in a syntactic category \ldots, or too broad, if we
require only that some context be shared.
\end{quote}

The structure bootstrapping system uses the ``broad'' method for
substitutability. Hypotheses are introduced when ``some context''
is shared. This will introduce too many hypotheses and some of
the introduced hypotheses might have an incorrect type (as
\citeauthor{Chomsky:LSLT-75} rightly points out). However, when
overlapping hypotheses are removed, the more likely hypotheses
will remain.

So far, \citeauthor{Chomsky:LSLT-75} talked about
substitutability of words. In the second problem he states: ``We
cannot freely allow substitution of word sequences for one
another.'' According to \citeauthor{Chomsky:LSLT-75}, this cannot
be done, since it will introduce incorrect constituents.

The solution to this problem is similar to the solution of the
first problem. Since \emph{hypotheses} about constituents are
stored and afterwards the best hypotheses are selected, we
conjecture that probably no incorrect constituents will be
contained in the final structure. This conjecture will be tested
extensively in chapter~\ref{ch:results}.

\citeauthor{Chomsky:LSLT-75}'s last problem deals with homonyms:
``[Homonyms] are best understood as belonging simultaneously to
several categories of the same order.'' \citeauthor{Pinker:LI-94}
discussed this problem extensively. His discussion will be dealt
with next.

%%%%%%%%%%%%%%%%%%%%%%%%%%%%%%%%%%%%%%%%%%%%%%%%%%%%%%%%%%%%%%%%%%%%%%%%%%%%%%%%
\thesissubsubsection{Pinker's objections to substitutability}
%%%%%%%%%%%%%%%%%%%%%%%%%%%%%%%%%%%%%%%%%%%%%%%%%%%%%%%%%%%%%%%%%%%%%%%%%%%%%%%%
\label{s:pinker}

\citet[pp.\ 283--288]{Pinker:LI-94} discusses two problems that
\index{Pinker, S.}%
deal with the notion of substitutability. The first problem deals
with words that receive an incorrect type. For the second
problem, \citeauthor{Pinker:LI-94} shows that finding one-word
constituents is not enough. Instead of classifying words, phrases
need to be classified. He then shows that there are too many ways
of doing this, concluding that the problem is too difficult to
solve in an unsupervised way.

When describing the first problem, \citeauthor{Pinker:LI-94}
shows how structure can be learned by considering the sentences
in~\ref{pinkersent1}.
\eenumsentence{
\toplabel{pinkersent1}
\item Jane eats chicken
\item Jane eats fish
\label{fish1}
\item Jane likes fish
\label{fish2}
}
From this, he concludes that sentences contain three words,
\sent{Jane}, followed by \sent{eats} or \sent{likes}, again
followed by \sent{chicken} or \sent{fish}. This is exactly what
the structure bootstrapping system described in this thesis does.
He then continues by giving the sentences as
in~\ref{pinkersent2}.
\eenumsentence{
\toplabel{pinkersent2}
\item Jane eats slowly
\item Jane might fish
\label{fish3}
}
However, this introduces inconsistencies, since \sent{might} may
now appear in the second position and \sent{slowly} may appear in
the third position, rendering sentences like \sent{Jane might
slowly}, \sent{Jane likes slowly} and \sent{Jane might chicken}
correct.

This is a complex problem and the current system, which selects
hypotheses based on the chronological order of learning hypotheses,
cannot cope with it. However, a probabilistic system (that assigns
types to hypotheses based on probabilities) will be able to solve this
problem. In section~\vref{s:duallevel} a solution to this problem is
discussed in more detail, but the main line of the solution is briefly
described here.

The problem with \citeauthor{Pinker:LI-94}'s approach (and
actually \citeauthor{Harris:SL-51} makes the same mistake) is
that he does not allow his system to recognise that words may
belong to different classes. In other words, his approach will
assign one class to a word (or in general, a phrase), for example
a word is a noun and nothing else.  This is clearly incorrect, as
can be seen in sentences~\ref{fish1} and~\ref{fish2} (\sent{fish}
is a noun) and sentence~\ref{fish3} (\sent{fish} is a verb).

However, the contexts of a word that does not have one clear type
help to distinguish between the different types. A noun like
\sent{fish} can occur in places in sentences where the verb
\sent{fish} cannot. Consider for example the sentences
in~\ref{fishsent}. The noun \sent{fish} can never occur in the
context of the first sentence, while the verb \sent{fish} cannot
occur in the context of the second sentence.
\vspace{-.15em}
\eenumsentence{
\toplabel{fishsent}
\item We fish for trout
\item Jane eats fish
}
\vspace{-.15em}

Using these differences in contexts, a word may be classified as
having a certain type in one context and another type in another
context. For example, verb-like words occur in verb contexts and
noun-like words occur in noun contexts. The frequencies of the
word in the different contexts indicate which type the word has
in a specific context.

\citeauthor{Pinker:LI-94} continues with the second problem. He
wonders what \emph{word} could occur before the word
\sent{bother} when he shows the sentences
in~\ref{pinkersentences3}. This introduces a problem, since there
are many different types of words that may occur before
\emph{bother}. From this, he concludes that looking for a
\emph{phrase} is the solution (a noun phrase in this particular
case).
\vspace{-.15em}
\eenumsentence{
\toplabel{pinkersentences3}
\item That dog bothers me [\sent{dog}, a noun]
\item What she wears bothers me [\sent{wears}, a verb]
\item Music that is too loud bothers me [\sent{loud}, an adjective]
\item Cheering too loudly bothers me [\sent{loudly}, an adverb]
\item The guy she hangs out with bothers me [\sent{with}, a preposition]
}

\citeauthor{Pinker:LI-94} then suggests considering all possible
ways to group words into phrases. This results in $2^{n-1}$
possibilities if the sentence has length $n$. Since there are too
many possibilities, the child (in our case, the structure
bootstrapping system) needs additional guidance. This additional
guidance clashes with the goal of minimum of information, so
\citeauthor{Pinker:LI-94} implies that an unsupervised bootstrapping
system is not feasible.

We believe that \citeauthor{Pinker:LI-94} missed the point here.
It is clear that applying the system that has been described
earlier in this chapter to the sentences
in~\ref{pinkersentences3} will find exactly the correct
constituents.  In all sentences the words before \sent{bothers
me} are grouped in constituents of the same type. In other words,
the system does not need any guiding as \citeauthor{Pinker:LI-94}
wants us to believe.

%% end of file: firstattempt.tex

%% file: overview.tex
%%%%%%%%%%%%%%%%%%%%%%%%%%%%%%%%%%%%%%%%%%%%%%%%%%%%%%%%%%%%%%%%%%%%%%%%%%%%%%%%
%% Menno van Zaanen                                                           %%
%% menno@comp.leeds.ac.uk                                                     %%
%%%%%%%%%%%%%%%%%%%%%%%%%%%%%%%%%%%%%%%%%%%%%%%%%%%%%%%%%%%%%%%%%%%%%%%%%%%%%%%%
%% Filename: overview.tex                                                     %%
%%%%%%%%%%%%%%%%%%%%%%%%%%%%%%%%%%%%%%%%%%%%%%%%%%%%%%%%%%%%%%%%%%%%%%%%%%%%%%%%

%%%%%%%%%%%%%%%%%%%%%%%%%%%%%%%%%%%%%%%%%%%%%%%%%%%%%%%%%%%%%%%%%%%%%%%%%%%%%%%%
%%%%%%%%%%%%%%%%%%%%%%%%%%%%%%%%%%%%%%%%%%%%%%%%%%%%%%%%%%%%%%%%%%%%%%%%%%%%%%%%
%%%%%%%%%%%%%%%%%%%%%%%%%%%%%%%%%%%%%%%%%%%%%%%%%%%%%%%%%%%%%%%%%%%%%%%%%%%%%%%%
\thesischapter{The \abl Framework}
{One or two homologous sequences whisper \ldots\\
 a full multiple alignment shouts out loud.}
{\citet{Hubbard:96-4-313}}
\label{ch:overview}
%%%%%%%%%%%%%%%%%%%%%%%%%%%%%%%%%%%%%%%%%%%%%%%%%%%%%%%%%%%%%%%%%%%%%%%%%%%%%%%%
%%%%%%%%%%%%%%%%%%%%%%%%%%%%%%%%%%%%%%%%%%%%%%%%%%%%%%%%%%%%%%%%%%%%%%%%%%%%%%%%
%%%%%%%%%%%%%%%%%%%%%%%%%%%%%%%%%%%%%%%%%%%%%%%%%%%%%%%%%%%%%%%%%%%%%%%%%%%%%%%%
\index{Hubbard, T.J.P.}%
\index{Lesk, A.M.}%
\index{Tramontano, A.}%

The structure bootstrapping system described informally in the previous
chapter is one of many possible instances of a more general framework. This
framework is called \emph{Alignment-Based Learning (\abl)}
\index{Alignment-Based Learning}%
\index{ABL}%
and will be described more formally in this chapter.

Specific instances of \abl attempt to find structure using a
corpus of plain (unstructured) sentences. They do not assume a
structured training set to initialise, nor are they based on any
other language dependent information. All structural information
is gathered from the unstructured sentences only.  The output of
the algorithm is a labelled, bracketed version of the input
corpus. This corresponds to the goals as described in section
\ref{s:goals}. 

The \abl framework consists of two distinct phases:
\index{phase}%
\begin{enumerate}
\item \emph{alignment learning}
\index{phase!alignment learning}%
\index{alignment learning}%
\item \emph{selection learning}
\index{phase!selection learning}%
\index{selection learning}%
\end{enumerate}

The alignment learning phase is the most important, in that it
finds hypotheses about constituents by aligning sentences from
the corpus. The selection learning phase selects constituents
from the possibly overlapping hypotheses that are found by the
alignment learning phase.

Although the \abl framework consists of these two phases, it is
possible (and useful) to extend this framework with another
phase:
\begin{enumerate}
\setcounter{enumi}{2}
\item \emph{grammar extraction}
\index{phase!grammar extraction}%
\index{grammar!extraction}%
\end{enumerate}
As the name suggests, this phase extracts a grammar from the
structured corpus (as created by the alignment and selection
learning phases). This extended system is called
\emph{\parseabl.}\footnote{Pronounce \parseabl as parsable.}
\index{parseABL}%

Figure~\ref{fig:overviewabl} gives a graphical description of the
\abl and \parseabl frameworks. The parts surrounded by a dashed
line depict data structures, while the parts with solid lines
mark phases in the system. The two parts surrounded by two solid
lines are the output data structures.  The first chain describes
the \abl framework. Continuing the first chain with the second
yields the \parseabl framework. All different parts in this
figure will be described in more detail next.

\begin{fig}{Overview of the \abl framework}
\label{fig:overviewabl}
\vspace{1em}
\psset{framearc=.2,arrowscale=2}
\begin{psmatrix}[colsep=25pt]
[name=corpus]\psframebox[linestyle=dashed]{\parbox{2cm}{Corpus\\}} &
[name=alignment]\psframebox{\parbox{2cm}{Alignment\\Learning}} &
[name=hypothesis]\psframebox[linestyle=dashed]{\parbox{2cm}{Hypothesis\\Space}} &
[name=selection]\psframebox{\parbox{2cm}{Selection\\Learning}} &
[name=structureda]\psframebox[doubleline=true]{\parbox{2cm}{Structured\\Corpus}}\\
\empty[name=structuredb]\psframebox[linestyle=dashed]{\parbox{2cm}{Structured\\Corpus}} &
[name=extract]\psframebox{\parbox{2cm}{Grammar\\Extraction}} &
[name=grammar]\psframebox[doubleline=true]{\parbox{2cm}{Grammar\\}}
\psset{arrows=->}
\ncline{corpus}{alignment}
\ncline{alignment}{hypothesis}
\ncline{hypothesis}{selection}
\ncline{selection}{structureda}
\ncline{structuredb}{extract}
\ncline{extract}{grammar}
\ncangle[linestyle=dashed,angleA=270,angleB=90]{structureda}{structuredb}
\end{psmatrix}
\end{fig}

%%%%%%%%%%%%%%%%%%%%%%%%%%%%%%%%%%%%%%%%%%%%%%%%%%%%%%%%%%%%%%%%%%%%%%%%%%%%%%%%
%%%%%%%%%%%%%%%%%%%%%%%%%%%%%%%%%%%%%%%%%%%%%%%%%%%%%%%%%%%%%%%%%%%%%%%%%%%%%%%%
\thesissection{Input}
%%%%%%%%%%%%%%%%%%%%%%%%%%%%%%%%%%%%%%%%%%%%%%%%%%%%%%%%%%%%%%%%%%%%%%%%%%%%%%%%
%%%%%%%%%%%%%%%%%%%%%%%%%%%%%%%%%%%%%%%%%%%%%%%%%%%%%%%%%%%%%%%%%%%%%%%%%%%%%%%%
\index{input}%
\label{formgoals}%

As described in the previous chapter, the main goal of \abl is to find useful
structure using plain input sentences only. To describe this input in a more
formal way, let us define a sentence.

\begin{definition}[Sentence]
\index{sentence}%
\index{sentence!plain}%
A \emph{sentence} or \emph{plain sentence} $S$ of length $|S|=n$ is a
non-empty list of words $[w_1, w_2, \ldots, w_n]$. The words are considered
elementary. A word $w_i$ in sentence $S$ is written as $S[i]=w_i$.
\end{definition}

\abl cannot learn using only one sentence, it uses more sentences to find
structure. The sentences it uses are stored in a list called a corpus. Note
that according to the definition, a corpus can never contain structured
sentences.

\begin{definition}[Corpus]
\index{corpus}%
A \emph{corpus} $U$ of size $|U|=n$ is a list of sentences
$[S_1, S_2, \ldots S_n]$.
\end{definition}

%%%%%%%%%%%%%%%%%%%%%%%%%%%%%%%%%%%%%%%%%%%%%%%%%%%%%%%%%%%%%%%%%%%%%%%%%%%%%%%%
%%%%%%%%%%%%%%%%%%%%%%%%%%%%%%%%%%%%%%%%%%%%%%%%%%%%%%%%%%%%%%%%%%%%%%%%%%%%%%%%
\thesissection{Alignment learning}
%%%%%%%%%%%%%%%%%%%%%%%%%%%%%%%%%%%%%%%%%%%%%%%%%%%%%%%%%%%%%%%%%%%%%%%%%%%%%%%%
%%%%%%%%%%%%%%%%%%%%%%%%%%%%%%%%%%%%%%%%%%%%%%%%%%%%%%%%%%%%%%%%%%%%%%%%%%%%%%%%
\label{s:alignmentlearning}
\index{alignment learning}%

A corpus of sentences is used as (unstructured) input. The
framework attempts to find structure in this corpus. The basic
unit of structure is a constituent, which describes a group of
words.

\begin{definition}[Constituent]
\index{constituent}%
A \emph{constituent} in sentence $S$ is a tuple $c^S=\tuple{$b$,
$e$, $n$}$ where $0\leq b\leq e\leq|S|$. $b$ and $e$ are indices
in $S$ denoting respectively the beginning and end of the
constituent. $n$ is the non-terminal of the constituent and is
taken from the set of non-terminals. $S$ may be omitted when its
value is clear from the context.
\end{definition}

The goal of the \abl framework is to introduce constituents in
the unstructured input sentences. The alignment learning phase
indicates where in the input sentences constituents \emph{may}
occur. Instead of introducing constituents, the alignment
learning phase indicates \emph{possible} constituents. These
possible constituents are called hypotheses.

\begin{definition}[Hypothesis]
A \emph{hypothesis} describes a possible constituent. It indicates where a
constituent may (but not necessarily needs to) occur. The
structure of a hypothesis is exactly the same as the structure of
a constituent.
\end{definition}

Now we can describe a sentence and hypotheses about constituents. Both are
combined in a fuzzy tree.

\begin{definition}[Fuzzy tree]
\index{fuzzy tree}%
\index{tree!fuzzy}%
A \emph{fuzzy tree} is a tuple $F=\tuple{$S$, $H$}$, where $S$ is a
sentence and $H$ a set of hypotheses $\{h^S_1, h^S_2, \ldots\}$.
\end{definition}

Similarly to storing sentences in a corpus, one can store fuzzy trees in a
hypothesis space.

\begin{definition}[Hypothesis space]
\label{def:hypothesisspace}
A \emph{hypothesis space} $D$ is a list of fuzzy trees.
\end{definition}

The process of alignment learning
\index{goal!alignment learning}%
converts a corpus (of sentences) into a hypothesis space (of fuzzy trees).
Section~\ref{s:findconstituent} on page~\pageref{substitutability} informally
showed how hypotheses can be found using Harris's notion of substitutable
\index{substitutability}%
segments, what he called freely substitutable segments. Applying this notion
to our problem yields: \emph{constituents of the same type can be substituted
by each other}.

Harris also showed how substitutable segments can be found. Informally this
can be described as: if two segments occur in the same context, they are
substitutable. In our problem, the notion of substitutability can be defined
as follows (using the auxiliary definition of a subsentence).

\begin{definition}[Subsentence or word group]
\index{subsentence}%
\index{word group}%
A \emph{subsentence} or \emph{word group} of sentence $S$ is a list of words
$v^S_{i\ldots j}$ such that $S=u+v^S_{i\ldots j}+w$ (the $+$ is defined to be
the concatenation operator on lists), where $u\mbox{ and }w$ are lists of
words and $v^S_{i\ldots j}$ with $i\leq j$ is a list of $j-i$ elements where
for each $k$ with $1\leq k\leq j-i:v^S_{i\ldots j}[k]=S[i+k]$.  A
subsentence may be empty (when $i=j$) or it may span the entire
sentence (when $i=0$ and $j=|S|$). $S$ may be omitted if its
meaning is clear from the context.
\end{definition}

\begin{definition}[Substitutability]
\index{substitutability}%
\index{subsentence!substitutable}%
\label{def:substitutability}%
Subsentences $u$ and $v$ are \emph{substitutable} for each other if
\begin{enumerate}
\item the sentences $S_1=t+u+w$ and $S_2=t+v+w$ (with $t$ and $w$
subsentences) are both valid, and
\item for each $k$ with $1\leq k\leq |u|$ it holds that $u[k]\not\in v$ and
for each $l$ with $1\leq l\leq |v|$ it holds that $v[l]\not\in u$.
\end{enumerate}
\end{definition}

Note that this definition of substitutability allows for the
substitution of empty subsentences. In the rest of the thesis we
assume that for two subsentences to be substitutable, at least
one of the two subsentences needs to be non-empty.

Consider the sentences in~\ref{subexample}. In this case, the
words \sent{Bert} and \sent{Ernie} are the unequal parts of the
sentences. These words are the only words that are substitutable
according to the definition. The word groups \sent{sees Bert} and
\sent{Ernie} are not substitutable, since the first condition in
the definition does not hold ($t={}$\sent{Oscar} in
\ref{subexample1} and $t={}$\sent{Oscar sees} in \ref{subexample2})
On the other hand, the word groups \sent{sees Bert} and \sent{sees
Ernie} are not substitutable, since these clash with the second
condition. The word \sent{sees} is present in both word groups.

\newpage
\eenumsentence{
\toplabel{subexample}
\item \underline{Oscar sees} Bert
\label{subexample1}
\item \underline{Oscar sees} Ernie
\label{subexample2}
}

The advantage of this notion of substitutability is that the substitutable
word groups can be found easily by searching for unequal parts of sentences.
Section~\ref{s:alignmentinst} will show how exactly the unequal parts (and
thus the substitutable parts) between two sentences can be found.

In Harris's definition of substitutability it is unclear whether
equal words may occur in substitutable word groups.
Definition~\ref{def:substitutability} clearly states that the two
substitutable subsentences may not have any words in common. This
definition is equal to Harris's if he meant to exclude
substitutable subsentences with words in common, or
definition~\ref{def:substitutability} is much stronger than
Harris's if he did mean to allow equal words in substitutable
word groups.

Harris used an informant to test whether a sentence is valid or
not: ``[I]f our informant accepts [the sentence] or if we hear an
informant say [the sentence], \ldots, then we say [the word
groups] are mutually substitutable''\citep[p.\ 31]{Harris:SL-51}.
\index{Harris, Z.S.}%
However, in an unsupervised structure bootstrapping system there
is no informant. The only information about the language is
stored in the corpus.  Therefore, we consider the validity of a
sentence as follows.

\begin{theorem}[Validity]
\index{validity!of a sentence}%
\index{sentence!valid}%
A sentence $S$ is \emph{valid} if an occurrence of $S$ can be found in the
corpus.
\end{theorem}

The definition of substitutability allows us to test if two subsentences are
substitutable. If two subsentences are substitutable, they may be replaced by
each other and still retain a valid sentence. With this in mind, a more
general version of the definition of substitutability can be given. This
version can test for multiple substitutable subsentences simultaneously.
\begin{definition}[Substitutability (general case)]
\index{substitutability!general case}%
Subsentences $u_1, u_2, \ldots, u_n$ and $v_1, v_2, \ldots, v_n$
are pairwise \emph{substitutable} for each other if the sentences
$S_1=s_1+u_1+s_2+u_2+s_3+\cdots+s_n+u_n+s_{n+1}$ and
$S_2=s_1+v_1+s_2+v_2+s_3+\cdots+s_n+v_n+s_{n+1}$ are both valid
and for each $k$ with $1\leq k\leq |n|$ the sentences
$T_1=s_k+u_k+s_{k+1}$ and $T_2=s_k+v_k+s_{k+1}$ indicate that
$u_k$ and $v_k$ are substitutable for each other if $T_1$ and
$T_2$ would be valid.
\end{definition}

The idea behind substitutability is that two substitutable
subsentences can be replaced by each other. This is directly
reflected in the general definition of substitutability. Sentence
$S_1$ can be transformed into sentence $S_2$ by replacing
substitutable subsentences. This transformation is accomplished
by substituting pairs of subsentences exactly as in the simple
case.

The assumption on how to find hypotheses can now be rephrased as:
\begin{theorem}[Hypotheses as subsentences]
\index{subsentence!is constituent}%
If subsentences $v_{i\ldots j}$ and $u_{k\ldots l}$ are
substitutable for each other then this yields hypotheses
$h_1=\tuple{$i$, $j$, $n$}$ and $h_2=\tuple{$k$, $l$, $n$}$ with
$n$ denoting a type label.
\end{theorem}

The goal of the alignment learning phase is to convert a corpus
into a hypothesis space. Algorithm~\ref{alg:al} gives pseudo code
of a function that takes a corpus and outputs its corresponding
hypothesis space. It first converts the plain sentences in the
corpus into fuzzy trees. Each fuzzy tree consists of the sentence
and a hypothesis indicating that the sentence can be reached from
the start symbol (of the underlying grammar). It then compares
the fuzzy tree to all fuzzy trees that are already present in the
hypothesis space. Comparing the two fuzzy trees yields
substitutable subsentences (if present) and from that it infers
new hypotheses. Finally, the fuzzy tree is added to the
hypothesis space.

\begin{alg}{Alignment learning}
\label{alg:al}
\index{alignment learning}%
\begin{programbox}
\keyword{func} \texttt{AlignmentLearning}%
   ($U$: corpus): hypothesis space\\
\mbox{\# The sentences in $U$ will be used to find hypotheses}\\
\keyword{var} $S$: sentence,\\
\>$F$, $G$: fuzzy tree,\\
\>$H$: set of hypotheses,\\
\>$SS$: list of pairs of pairs of indices in a sentence,\\
\>$PSS$: pair of pairs of indices in a sentence,\\
\>$B_F$, $E_F$, $B_G$, $E_G$: indices in a sentence,\\
\>$N$: non-terminal,\\
\>$D$: hypothesis space\\
\keyword{begin}\\
\>\keyword{foreach} $S\in U$ \keyword{do}\\
\>\>$H:=\{\tuple{$0$, $|S|$, startsymbol}\}$\\
\>\>$F:=\tuple{$S$, $H$}$\\
\>\>\keyword{foreach} $G\in D$ \keyword{do}\\
\>\>\>$SS:={}$\texttt{FindSubstitutableSubsentences}($F$, $G$)\\
\>\>\>\keyword{foreach} $PSS\in SS$ \keyword{do}\\
\>\>\>\>$\tuple{\tuple{$B_F$, $E_F$}, \tuple{$B_G$, $E_G$}}:=PSS$\\
\>\>\>\>$N:={}$\texttt{NewNonterminal}() %
        \`\mbox{\# Return a new (unused) non-terminal}\\
\>\>\>\>\texttt{AddHypothesis}(\tuple{$B_F$, $E_F$, $N$}, $F$)
        \`\mbox{\# Add to set of hypotheses of $F$}\\
\>\>\>\>\texttt{AddHypothesis}(\tuple{$B_G$, $E_G$, $N$}, $G$)\\
\>\>\>\keyword{od}\\
\>\>\keyword{od}\\
\>\>$D:=D+F$\`\mbox{\# Add $F$ to $D$}\\
\>\keyword{od}\\
\>\keyword{return} $D$\\
\keyword{end.}
\end{programbox}
\end{alg}

In the algorithm there are two undefined functions and one undefined
procedure:
\begin{enumerate}
\item \texttt{NewNonterminal},
\item \texttt{FindSubstitutableSubsentences}, and
\item \texttt{AddHypothesis}.
\end{enumerate}
The first function, \texttt{NewNonterminal} simply returns a new
(unused) non-terminal. The other function and procedure are more
complex and will be described in more detail next.

%%%%%%%%%%%%%%%%%%%%%%%%%%%%%%%%%%%%%%%%%%%%%%%%%%%%%%%%%%%%%%%%%%%%%%%%%%%%%%%%
\subsection{Find the substitutable subsentences}
%%%%%%%%%%%%%%%%%%%%%%%%%%%%%%%%%%%%%%%%%%%%%%%%%%%%%%%%%%%%%%%%%%%%%%%%%%%%%%%%
\index{subsentence!substitutable}%

The function \texttt{FindSubstitutableSubsentences} finds
substitutable subsentences in the sentences of its arguments. The
arguments of the function, $F$ and $G$, are both fuzzy trees. A
subsentence in the sentence of a fuzzy tree is stored as a pair
\tuple{$B$, $E$}. $B$ denotes the begin index of the subsentence
and $E$ refers to the end index (as if describing a subsentence
$v_{B\ldots E}$). Substitutable subsentences in $F$ and $G$ are
stored in pairs of subsentences, for example \tuple{\tuple{$B_F$,
$E_F$}, \tuple{$B_G$, $E_G$}}, where \tuple{$B_F$, $E_F$} is the
substitutable subsentence in the sentence of fuzzy tree $F$ and
similarly \tuple{$B_G$, $E_G$} in $G$.

Using different methods to find the substitutable subsentences results in
different instances of the alignment learning phase. Three different methods
will be described in section~\vref{s:alignmentinst}. For now, let us assume
that there exists a method that can find substitutable subsentences.

%%%%%%%%%%%%%%%%%%%%%%%%%%%%%%%%%%%%%%%%%%%%%%%%%%%%%%%%%%%%%%%%%%%%%%%%%%%%%%%%
\subsection{Insert a hypothesis in the hypothesis space}
%%%%%%%%%%%%%%%%%%%%%%%%%%%%%%%%%%%%%%%%%%%%%%%%%%%%%%%%%%%%%%%%%%%%%%%%%%%%%%%%
\label{s:inserthypothesis}

The procedure \texttt{AddHypothesis} adds its first argument, a
hypothesis in the form \tuple{$b$, $e$, $n$} to the set of
hypotheses of its second argument (a fuzzy tree). However, there
are some cases in which simply adding the hypothesis to the set
does not exactly result in the expected structure.

In total, three distinct cases have to be considered. Assume that
the procedure is called to insert hypothesis $h_F=\tuple{$B_F$,
$E_F$, $N$}$ into fuzzy tree $F$ and next, $h_G=\tuple{$B_G$,
$E_G$, $N$}$ is inserted in $G$. In the algorithm, hypotheses are
always added in pairs, since substitutable subsentences always
occur in pairs. The three cases will be described with help from
the following definition.

\begin{definition}[Equal and equivalent hypotheses]
Two hypotheses $h_1=\tuple{$b_1$, $e_1$, $n_1$}$ and $h_2=\tuple{$b_2$, $e_2$,
$n_2$}$ are called \emph{equal} when $b_1=b_2$, $e_1=e_2$, and $n_1=n_2$. The
hypotheses $h_1$ and $h_2$ are \emph{equivalent} when $b_1=b_2$ and $e_1=e_2$,
but $n_1=n_2$ need not be true.
\end{definition}

\begin{enumerate}
\item The sets of hypotheses of both $F$ and $G$ do not contain
hypotheses equivalent to $h_F$ and $h_G$ respectively.
\item The set of hypotheses of $F$ already contains a hypothesis
equivalent to $h_F$ or the set of hypotheses of $G$ already
contains a hypothesis equivalent to $h_G$. 
\item The sets of hypothesis of both $F$ and $G$ already contain
hypotheses equivalent to hypotheses $h_F$ and $h_G$ respectively.
\end{enumerate}

Let us consider the first case. Both $F$ and $G$ receive completely new
hypotheses. This occurs for example with the fuzzy trees
in~\ref{insertcase1}.\footnote{In this thesis, non-terminals are
natural numbers starting from $1$, which is also the start
symbol. New non-terminals are introduced by taking the next
lowest, unused natural number.}
\eenumsentence{
\toplabel{insertcase1}
\item {}[\align{Oscar sees} Bert]\snt{1}
\label{insertcase1a}
\item {}[\align{Oscar sees} Big Bird]\snt{1}
\label{insertcase1b}
}
In this case, the subsentences denoting \sent{Bert} and \sent{Big Bird} are
substitutable for each other, so a new non-terminal is chosen ($\nt{2}$ in
this case) and the hypotheses \tuple{$2$, $3$, $\nt{2}$} in fuzzy
tree~\ref{insertcase1a} and \tuple{$2$, $4$, $\nt{2}$} in fuzzy
tree~\ref{insertcase1b} are inserted in the respective sets of hypotheses of
the fuzzy trees. This results in the fuzzy trees as shown
in~\ref{insertedcase1} as expected.
\eenumsentence{
\toplabel{insertedcase1}
\item {}[Oscar sees [Bert]\snt{2}]\snt{1}
\item {}[Oscar sees [Big Bird]\snt{2}]\snt{1}
}

The second case is slightly more complex. Consider the fuzzy trees
in~\ref{insertcase2}.
\eenumsentence{
\toplabel{insertcase2}
\item {}[\align{Oscar sees} [Bert]\snt{2}]\snt{1}
\label{insertcase2a}
\item {}[\align{Oscar sees} Big Bird]\snt{1}
\label{insertcase2b}
}
Since hypotheses are found by considering the plain sentences only, the
hypotheses \tuple{$2$, $3$, $\nt{3}$} and \tuple{$2$, $4$, $\nt{3}$} should be
inserted respectively in the sets of hypotheses of fuzzy
trees~\ref{insertcase2a} and ~\ref{insertcase2b}. However, the first fuzzy tree
already has a hypothesis equivalent to the new one.

Finding the hypotheses in fuzzy trees~\ref{insertcase2} indicates that
\sent{Bert} and \sent{Big Bird} might (in the end) be constituents of the same
type. Therefore, both should receive the same non-terminal. This can be
achieved in two similar ways. One way is by adding the hypothesis \tuple{$2$,
$4$, $\nt{2}$} to fuzzy tree~\ref{insertcase2b} instead of \tuple{$2$, $4$,
$\nt{3}$} (its non-terminal is equal to the non-terminal of the existing
hypothesis) and no hypothesis is added to~\ref{insertcase2a}. This yields
fuzzy trees:
\eenumsentence{
\toplabel{insertcaseexample}
\item {}[\align{Oscar sees} [Bert]\snt{2}]\snt{1}
\item {}[\align{Oscar sees} [Big Bird]\snt{2}]\snt{1}
}

The other way is to insert the hypotheses in the regular way (overriding the
existing \tuple{$2$, $3$, $\nt{2}$} in the fuzzy tree of~\ref{insertcase2a})
\eenumsentence{
\item {}[\align{Oscar sees} [Bert]\snt{2+3}]\snt{1}
\item {}[\align{Oscar sees} [Big Bird]\snt{3}]\snt{1}
}
Then all occurrences of non-terminal \nt{2} in the entire hypothesis
space are replaced by non-terminal \nt{3}, again resulting in the fuzzy trees
of~\ref{insertcaseexample}.

The third case (where hypotheses are found that are equivalent to
existing hypotheses in both fuzzy trees), falls into two
subcases.

The first subcase is when the two original hypotheses already have the same
type. This is depicted in fuzzy trees~\ref{insertcase3a}.
\eenumsentence{
\toplabel{insertcase3a}
\item {}[\align{Oscar sees} [Bert]\snt{2}]\snt{1}
\item {}[\align{Oscar sees} [Big Bird]\snt{2}]\snt{1}
}
Since it is already known that \sent{Bert} and \sent{Big Bird} are hypotheses
of the same type, nothing has to be changed and the hypotheses do not
even need to be inserted in the sets of hypotheses.

However, the second subcase is more difficult. Consider the following fuzzy
trees:
\eenumsentence{
\toplabel{insertcase3b}
\item {}[\align{Oscar sees} [Bert]\snt{2}]\snt{1}
\item {}[\align{Oscar sees} [Big Bird]\snt{3}]\snt{1}
}
It is now known that \sent{Bert} and \sent{Big Bird} should have the same type,
because the tuples \tuple{$2$, $3$, $\nt{4}$} and \tuple{$2$, $4$, $\nt{4}$}
are being inserted in the respective sets of hypotheses. In other words, it is
known that the non-terminals \nt{2} and \nt{3} both describe the same type.
Therefore, types \nt{2} and \nt{3} can be merged
\index{constituent!merge}%
\index{merge types}%
and all occurrences of both types in the entire hypothesis space can be
updated.

%%%%%%%%%%%%%%%%%%%%%%%%%%%%%%%%%%%%%%%%%%%%%%%%%%%%%%%%%%%%%%%%%%%%%%%%%%%%%%%%
\thesissubsection{Clustering}
%%%%%%%%%%%%%%%%%%%%%%%%%%%%%%%%%%%%%%%%%%%%%%%%%%%%%%%%%%%%%%%%%%%%%%%%%%%%%%%%
\label{s:clustering}

In the previous section, it was assumed that word groups in the
same context are always of the same type.
The assumption that if there is some evidence that word groups
occur in the same context then they also have the same non-terminal
type, might be too strong (and indeed it is, as will be shown
below).  However, this is just one of the many ways of merging
non-terminal types.

In fact, the merging of non-terminal types can be seen as a
sub-phase of the alignment learning phase. An instantiation based
on the assumption made in the previous section, namely that
hypotheses which occur in the same context always have the same
non-terminal type, is only one possible instantiation of this
phase.  In section~\ref{s:duallevel}, another instantiation is
discussed.  It is important to remember that the assumption of
the previous section is in no way a feature of the \abl
framework, it is merely a feature of one of its instantiations.

All systems in this thesis use the instantiation that merges
types when two hypotheses with different types occur in the same
context. This assumption influences one of the selection learning
instantiations and additionally, the qualitative evalution relies
on this assumption, although similar results may be found when
other cluster methods are used.

The assumption that hypotheses in the same context always have
the same non-terminal type is not necessarily always true, as has
been shown by Pinker (see section~\vref{s:pinker}). Consider the
following fuzzy trees:
\eenumsentence{
\toplabel{samecontext}
\item {}[\align{Ernie eats} well]\snt{1}
\item {}[\align{Ernie eats} biscuits]\snt{1}
}
In this case, \sent{biscuits} is a noun and \sent{well} is an
adjective. However, \abl will conclude that both are of the same
type. Since \sent{well} and \sent{biscuits} are substitutable
subsentences, they will be contained in two hypotheses with the
same non-terminal.

Merging non-terminals as described in the previous section
assumes that when there is evidence that two non-terminals
describe the same type, they are merged. However, finding
\emph{hypotheses} merely indicates that the possibility exists
that a constituent occurs there.  Furthermore, it indicates that
a constituent with a certain type occupies that context. Simply
merging non-terminals assumes that hypotheses are actually
constituents and that the evidence is completely correct.

A better method of merging non-terminals, which also solves
Pinker's problem, would only merge types when enough evidence,
for example in the form of frequencies of contexts, is found. For
example, types could be clustered when the hypotheses are chosen
to be correct and all hypotheses having one of the two types
occur mostly in the same contexts.  Section~\ref{s:duallevel}
will discuss this in more detail.

%%%%%%%%%%%%%%%%%%%%%%%%%%%%%%%%%%%%%%%%%%%%%%%%%%%%%%%%%%%%%%%%%%%%%%%%%%%%%%%%
%%%%%%%%%%%%%%%%%%%%%%%%%%%%%%%%%%%%%%%%%%%%%%%%%%%%%%%%%%%%%%%%%%%%%%%%%%%%%%%%
\thesissection{Selection learning}
%%%%%%%%%%%%%%%%%%%%%%%%%%%%%%%%%%%%%%%%%%%%%%%%%%%%%%%%%%%%%%%%%%%%%%%%%%%%%%%%
%%%%%%%%%%%%%%%%%%%%%%%%%%%%%%%%%%%%%%%%%%%%%%%%%%%%%%%%%%%%%%%%%%%%%%%%%%%%%%%%
\index{selection learning}%

The goal of the selection learning phase
\index{goal!selection learning}%
is to remove overlapping hypotheses from the hypothesis space.
Now that a more formal framework of hypotheses is available, it
is easy to define exactly what overlapping hypotheses are:

\begin{definition}[Overlapping hypotheses]
\index{constituent!overlap}%
\index{\overlap (overlap)}%
Two hypotheses $h_i=\tuple{$b_i$, $e_i$, $n_i$}\mbox{ and }
h_j=\tuple{$b_j$, $e_j$, $n_j$}$ \emph{overlap} (written as $h_i\overlap h_j$)
iff $(b_i<b_j<e_i<e_j)\mbox{ or }(b_j<b_i<e_j<e_i)$. (Overlap between
hypotheses from different fuzzy trees is undefined.)
\end{definition}

An example of overlapping hypotheses can be found
in~\ref{overrepeat}, which is the same fuzzy tree as
in~\ref{overthree:b}. It contains two hypotheses ($\tuple{0, 3,
$X_1$}$ and $\tuple{$\underbar{2}$, $\underbar{5}$, $X_2$}$) which
overlap, since $0<\underbar{2}<3<\underbar{5}$.

\enumsentence{
\label{overrepeat}
\brover{4.35cm}[\snt{X_1}\dalign{Big Bird}
\brunder{4.5cm}[\snt{X_2}throws]\snt{X_1} \align{the
apple}]\snt{X_2}
}

The fuzzy trees generated by the alignment learning phase closely resemble the
normal notion of tree structures. The only difference is that fuzzy trees
can have overlapping hypotheses. Regular trees never have this property.
\begin{definition}[Tree]
\index{tree}%
\index{tree!non-fuzzy}%
A \emph{tree} $T=\tuple{$S$, $C$}$ is a fuzzy tree such that for
each $h_i, h_j\in C: \neg(h_i\overlap h_j)$. The hypotheses in a
tree are called \emph{constituents}.
\end{definition}

Similarly, where a collection of fuzzy trees is called a hypothesis space, a
collection of trees is a treebank.
\index{treebank}%

\begin{definition}[Treebank]
\index{treebank}%
A \emph{treebank} $B$ is a list of trees.
\end{definition}

The goal of the selection learning phase
\index{goal!selection learning}%
can now be rephrased as transforming
a hypothesis space to a treebank. This means that each fuzzy tree in the
hypothesis space should be converted into a tree. In other words,
all overlapping hypotheses in the fuzzy trees in the hypothesis
space should be removed.

The previous chapter described a simple instantiation of the
phase that converts a hypothesis space into a treebank. The
method assumed that older hypotheses are always correct. In the
formal framework so far, this can be implemented in two different
ways:
\begin{description}
\item [change \texttt{AddHypothesis}] The system described
earlier adds hypotheses by calling the procedure
\texttt{AddHypothesis}. This procedure, therefore, can test if
there exists another (older) hypothesis in the fuzzy tree that
conflicts with the new hypothesis. If there exists one, the new
hypothesis is not inserted in the set of hypotheses. If no
overlapping hypothesis exists, the new hypothesis is added
normally.
\vspace{-.5em}
\item [change the way fuzzy trees store hypotheses] Storing
hypotheses in a \emph{list} of hypotheses, instead of storing
them in a set, allows the algorithm to keep track of which
hypotheses were inserted first. For example, if a hypothesis is
always appended to the end of the list (of hypotheses), it must
be true that the first hypothesis was the oldest and the last
hypothesis in the list is the newest.  Therefore, the selection
learning phase searches for pairs of overlapping hypotheses and
removes the one closer to the end of the list. This leaves the
oldest non-overlapping hypothesis in the list.
\end{description}

Albeit very simple, this method is crummy (as will also be shown
in the results in chapter~\ref{ch:results}). Fortunately, there
are many other methods to remove the overlapping hypotheses.
Section~\ref{s:selectioninst} describes two of them. The emphasis
of these methods lies on the selection of hypotheses by computing
the probability of each hypothesis.

%%%%%%%%%%%%%%%%%%%%%%%%%%%%%%%%%%%%%%%%%%%%%%%%%%%%%%%%%%%%%%%%%%%%%%%%%%%%%%%%
%%%%%%%%%%%%%%%%%%%%%%%%%%%%%%%%%%%%%%%%%%%%%%%%%%%%%%%%%%%%%%%%%%%%%%%%%%%%%%%%
\thesissection{Grammar extraction}
%%%%%%%%%%%%%%%%%%%%%%%%%%%%%%%%%%%%%%%%%%%%%%%%%%%%%%%%%%%%%%%%%%%%%%%%%%%%%%%%
%%%%%%%%%%%%%%%%%%%%%%%%%%%%%%%%%%%%%%%%%%%%%%%%%%%%%%%%%%%%%%%%%%%%%%%%%%%%%%%%
\index{grammar!extraction}%

Apart from the two phases (alignment learning and selection
learning) described above, \abl can be extended with a third
phase. The combination of alignment learning and selection
learning generates a treebank, a list of trees. These two phases
combined are called a \emph{structure bootstrapping system}.
Adding the third phase, which extracts a stochastic grammar from
the treebank, expands the systems into \emph{grammar
bootstrapping systems}.

It is possible to extract different types of stochastic grammars
from a treebank. Section~\vref{s:grammarextract} will describe
how two different types of grammar can be extracted. The next two
sections go into the advantages of extracting such a grammar from
the treebank.

%%%%%%%%%%%%%%%%%%%%%%%%%%%%%%%%%%%%%%%%%%%%%%%%%%%%%%%%%%%%%%%%%%%%%%%%%%%%%%%%
\thesissubsection{Comparing grammars}
%%%%%%%%%%%%%%%%%%%%%%%%%%%%%%%%%%%%%%%%%%%%%%%%%%%%%%%%%%%%%%%%%%%%%%%%%%%%%%%%
\index{grammar!comparing}%

Evaluating grammars can be done by comparing constituents
\index{constituent!compare}%
in sentences \emph{parsed} by the different grammars
\citep{Black:SNL91-306}.\footnote{Grammars can also be evaluated
by formally comparing the grammars in terms of generative power.
However, in this thesis, the emphasis will be on indirectly
comparing grammars.} The grammars are compared against a given
\index{Black, E.}%
\index{Abney, S.}%
\index{Flickinger, D.}%
\index{Gdaniec, C.}%
\index{Grishman, R.}%
\index{Harrison, P.}%
\index{Hindle, D.}%
\index{Ingria, R.}%
\index{Jelinek, F.}%
\index{Klavans, J.}%
\index{Liberman, M.}%
\index{Marcus, M.}%
\index{Roukos, S.}%
\index{Santorini, B.}%
\index{Strzalkowski, T.}%
treebank, which is considered completely correct, a gold
standard. From the treebank, a corpus is extracted. The
sentences in the corpus are parsed using both grammars. This
results in two new treebanks (one generated by each grammar) and
the original treebank. The constituents in the new treebanks that
can also be found in the gold standard treebank are counted.
Using these numbers from both parsed treebanks, evaluation metrics
such as precision and recall of the two grammars can be computed
and the grammars are (indirectly) compared against each other.

A similar approach will be taken when evaluating the different
alignment and selection learning phases in
chapter~\ref{ch:results}. Instead of parsing the sentences with a
grammar, the trees are generated by the alignment and selection
learning phases. In this case, there is never actually a grammar
present. The structure is built during the two phases.

When a grammar is extracted from the trees generated by alignment
and selection learning, this grammar can be evaluated in the
normal way.  Furthermore, the grammar can be evaluated against
other grammars (for example, grammars generated by other grammar
bootstrapping systems).

The \parseabl system, which is the \abl system extended with a
grammar extraction phase, returns a grammar as well as a
structured version of the plain corpus. The availability of a
grammar is not the only advantage of this extended system. The
selection learning phase also benefits from the grammar
extraction phase, as will be explained in the next section.

%%%%%%%%%%%%%%%%%%%%%%%%%%%%%%%%%%%%%%%%%%%%%%%%%%%%%%%%%%%%%%%%%%%%%%%%%%%%%%%%
\thesissubsection{Improving selection learning}
%%%%%%%%%%%%%%%%%%%%%%%%%%%%%%%%%%%%%%%%%%%%%%%%%%%%%%%%%%%%%%%%%%%%%%%%%%%%%%%%

The main idea behind the probabilistic selection learning methods
is that the most probable constituents (which are to be selected
from the set of hypotheses) should be introduced in the tree. The
probability of the combination of constituents is computed from
the probabilities of its parts (extracted from the hypothesis
space).

The probability of each possible combination of (overlapping
\emph{and} non-overlapping) hypotheses should be computed, but
the next chapter will show that this is difficult. In practice,
only the \emph{overlapping} hypotheses are considered for
deletion.  This assumes that non-overlapping hypotheses are
always correct. However, even non-overlapping hypotheses may be
incorrect.

Reparsing the plain sentences with an extracted grammar may find
other, more probable parses. This implies that more probable
constituents will be inserted in the sentences. Although the
(imperfect) selection learning phase still takes place, the
grammar extraction/parsing phase simulates selection learning,
reconsidering \emph{all} hypotheses.
\index{selection learning!with all constituents}%

\vspace{-.35em}
\eenumsentence{
  \toplabel{npleftright}
  \item \ldots \align{from} [Bert's house]\snt{2} \align{to}
[Sesame Street]\snt{3}
  \item \ldots \align{from} [Sesame Street]\snt{2} \align{to}
  [Ernie's room]\snt{3}
}
\vspace{-.35em}

Another problem that can be solved by extracting a grammar from the treebank
and then reparsing the plain sentences is depicted in
example~\ref{npleftright}. In these sentences \sent{\align{from}} and
\sent{\align{to}} serve as ``boundaries''.
\index{constituent!boundaries}%
Hypotheses are found between these boundaries and hypotheses between
\sent{\align{from}} and \sent{\align{to}} are always of type
\nt{2} and hypotheses after \sent{\align{to}} are of type \nt{3}.
However, \sent{Sesame Street} can now have two types depending on
the context. This may be correct if type \nt{2} is considered as
a ``from-noun phrase'' and type \nt{3} as a ``to-noun phrase'',
but normally \sent{Sesame Street} should have only one type
(e.g.\ a noun phrase).

Extracting a grammar and reparsing the plain sentences with this
\index{grammar!extraction}%
\index{parsing}%
grammar may solve this problem. When for example \sent{Sesame
Street} occurs more often with type \nt{2} than with type \nt{3},
reparsing the subsentence \sent{Sesame Street} will receive type
\nt{2} in both cases.\footnote{This occurs when for example
\sent{to} and \sent{from} both have the same type label and there
is a grammar rule describing a simple prepositional phrase.}
Finding more probable constituents can also happen on higher
levels with non-lexicalised grammar rules.

Note that reparsing not always solves this problem. If for
example hypotheses with type \nt{2} are more probable in one
context and hypotheses with type \nt{3} are more probable in the
other context, the parser will still find the parses
of~\ref{npleftright}.

It can be expected that reparsing the sentences will improve the
resulting treebank. A stochastic grammar contains probabilistic
\index{grammar!stochastic}%
information about the possible contexts of hypotheses, in
contrast to the selection learning phase which only uses local
probabilistic information (i.e.\ counts).

%% end of file: overview.tex

%% file: instantiations.tex
%%%%%%%%%%%%%%%%%%%%%%%%%%%%%%%%%%%%%%%%%%%%%%%%%%%%%%%%%%%%%%%%%%%%%%%%%%%%%%%%
%% Menno van Zaanen                                                           %%
%% menno@comp.leeds.ac.uk                                                     %%
%%%%%%%%%%%%%%%%%%%%%%%%%%%%%%%%%%%%%%%%%%%%%%%%%%%%%%%%%%%%%%%%%%%%%%%%%%%%%%%%
%% Filename: instantiations.tex                                              %%
%%%%%%%%%%%%%%%%%%%%%%%%%%%%%%%%%%%%%%%%%%%%%%%%%%%%%%%%%%%%%%%%%%%%%%%%%%%%%%%%

%%%%%%%%%%%%%%%%%%%%%%%%%%%%%%%%%%%%%%%%%%%%%%%%%%%%%%%%%%%%%%%%%%%%%%%%%%%%%%%%
%%%%%%%%%%%%%%%%%%%%%%%%%%%%%%%%%%%%%%%%%%%%%%%%%%%%%%%%%%%%%%%%%%%%%%%%%%%%%%%%
%%%%%%%%%%%%%%%%%%%%%%%%%%%%%%%%%%%%%%%%%%%%%%%%%%%%%%%%%%%%%%%%%%%%%%%%%%%%%%%%
\thesischapter{Instantiating the Phases}
{Do you still think you can control the game by brute force?}
{Shiwan Khan\\
 (tilt message in ``the Shadow'' pinball machine)}
%%%%%%%%%%%%%%%%%%%%%%%%%%%%%%%%%%%%%%%%%%%%%%%%%%%%%%%%%%%%%%%%%%%%%%%%%%%%%%%%
%%%%%%%%%%%%%%%%%%%%%%%%%%%%%%%%%%%%%%%%%%%%%%%%%%%%%%%%%%%%%%%%%%%%%%%%%%%%%%%%
%%%%%%%%%%%%%%%%%%%%%%%%%%%%%%%%%%%%%%%%%%%%%%%%%%%%%%%%%%%%%%%%%%%%%%%%%%%%%%%%
\label{ch:instantiations}

This chapter will discuss several instances for each of \abl's
phases.  Firstly, three different instances of the alignment
\index{phase}%
\index{instantiation}%
learning phase will be described, followed by three different
selection learning methods. Finally, two different grammar
extraction methods will be given.

%%%%%%%%%%%%%%%%%%%%%%%%%%%%%%%%%%%%%%%%%%%%%%%%%%%%%%%%%%%%%%%%%%%%%%%%%%%%%%%%
%%%%%%%%%%%%%%%%%%%%%%%%%%%%%%%%%%%%%%%%%%%%%%%%%%%%%%%%%%%%%%%%%%%%%%%%%%%%%%%%
\thesissection{Alignment learning instantiations}
%%%%%%%%%%%%%%%%%%%%%%%%%%%%%%%%%%%%%%%%%%%%%%%%%%%%%%%%%%%%%%%%%%%%%%%%%%%%%%%%
%%%%%%%%%%%%%%%%%%%%%%%%%%%%%%%%%%%%%%%%%%%%%%%%%%%%%%%%%%%%%%%%%%%%%%%%%%%%%%%%
\label{s:alignmentinst}
\index{alignment learning!instances}%

The first phase of \abl aligns pairs of sentences against each
other, where unequal parts of the sentences are considered
hypotheses. In algorithm~\vref{alg:al}, the function
\texttt{FindSubstitutableSubsentences}, which finds unequal parts
in a pair of sentences, is still undefined. In the previous
chapter it was assumed that such a function exists, but no
further details were given.

In this section, three different implementations of this function
are discussed. The first two are based on the edit distance
\index{edit distance algorithm}%
algorithm by~\citet{Wagner:74-21-168}.\footnote{Apparently, this
algorithm has been independently discovered by several
researchers at roughly the same time \citep{Sankoff:TWS-99}.}
\index{Wagner, R.A.}%
\index{Fischer, M.J.}%
This algorithm finds ways to convert one sentence into another,
\index{convert!sentences}%
from which equal and unequal parts of the sentences can be found.
The third method finds the substitutable subsentences by
considering all possible conversions if there exists more than
one.

%%%%%%%%%%%%%%%%%%%%%%%%%%%%%%%%%%%%%%%%%%%%%%%%%%%%%%%%%%%%%%%%%%%%%%%%%%%%%%%%
\thesissubsection{Alignment learning with edit distance}
%%%%%%%%%%%%%%%%%%%%%%%%%%%%%%%%%%%%%%%%%%%%%%%%%%%%%%%%%%%%%%%%%%%%%%%%%%%%%%%%

The edit distance algorithm consists of two distinct phases (as
will be described below). The first phase finds the edit or
Levenshtein distance between two sentences
\index{Levenshtein distance}%
\citep{Levenshtein:65-163-845}.
\index{Levenshtein, V.I.}%

\begin{definition}[Edit distance]
\index{edit distance}%
The \emph{edit distance} between two sentences is the minimum
edit cost\footnote{\citet{Gusfield:AOSTAS-97} calls this version
\index{Gusfield, D.}%
of edit distance, \emph{operation-weight edit distance}.
\index{edit distance!operation-weight}%
} needed to transform one sentence into the other.
\end{definition}

When transforming one sentence into the other, words unequal in
both sentences need to be converted. Normally, three edit
operations are distinguished: \emph{insertion}, \emph{deletion},
\index{edit operation}%
\index{insertion}%
\index{deletion}%
\index{substitution}%
\index{match}%
and \emph{substitution}, even though other operations may be
defined. Matching of words is not considered an edit operation,
although it works exactly the same.

Each of the edit operations have an accompanying cost, described
by the predefined cost function $\gamma$. When one sentence is
\index{cost function}%
\index{$\gamma$}%
transformed into another, the cost of this transformation is the
sum of the costs of the separate edit operations. The edit
distance is now the cost of the ``cheapest'' way to transform one
sentence into the other.

The edit distance between two sentences, however, does not yield
substitutable subsentences. The second phase of the algorithm
gives an \emph{edit transcript}, which is a step closer to the
\index{edit transcript}%
wanted information.

\begin{definition}[Edit transcript]
An \emph{edit transcript} is a list of labels denoting the
possible edit operations, which describes a transformation of one
sentence into the other.
\end{definition}

An example of an edit transcript can be found in
figure~\ref{fig:edittranscript}. In this figure \emph{INS}
denotes an insertion, \emph{DEL} means deletion, \emph{SUB}
stands for substitution and \emph{MAT} is a match.

\begin{fig}{Example edit transcript and alignment}
\label{fig:edittranscript}
\vspace{1em}
\begin{tabular}{l||l|l|l|l|l|l|}
Edit transcript: & INS & MAT      & MAT  & SUB  & MAT  & DEL        \\\hline
Sentence 1:      &     & Monsters & like & tuna & fish & sandwiches \\
Sentence 2:      & All & monsters & like & to   & fish & 
\end{tabular}
\end{fig}

Another way of looking at the edit transcript is an \emph{alignment}.
\index{alignment}%

\begin{definition}[Alignment]
An \emph{alignment} of two sentences is obtained by first
inserting chosen spaces, either into or at the ends of the
sentences, and then placing the two resulting sentences one above
the other so that every word or space in either sentence is
opposite a unique character or a unique space in the other
string.
\end{definition}

Edit transcripts and alignments are simply alternative ways of writing
the same notion. From an edit transcript it is possible to find the
corresponding alignment and vice versa (as can be seen in
figure~\ref{fig:edittranscript}).

Finding indices of words that are equal in two aligned sentences
is easy. Words that are located above each other and that are
equal in the alignment are called \emph{links}.

\begin{definition}[Link]
\index{link}%
A \emph{link} is a pair of indices \tuple{$i^S$, $j^T$} in two
sentences $S$ and $T$, such that $S[i^S]=T[j^T]$ and $S[i^S]$ is
above $T[j^T]$ in the alignment of the two sentences.
\end{definition}

In the example in figure~\ref{fig:edittranscript}, \tuple{1, 2},
\tuple{2, 3}, and \tuple{4, 5} are the links (when indices of the
words in the sentences start counting with 1). The first link
describes the word \sent{monsters}, the second \sent{like}, and
the third describes \sent{fish}.

Links describe which words are equal in both sentences. Combining
\index{word!equal in two sentences}%
the adjacent links results in equal subsentences. Two links
\tuple{$i_1$, $j_1$} and \tuple{$i_2$, $j_2$} are adjacent when
$i_1-i_2=j_1-j_2=\pm1$. For example, the pairs of indices
\tuple{1, 2} and \tuple{2, 3} are adjacent, since $1-2=-1$ and so
is $3-2$. Links \tuple{2, 3} and \tuple{4, 5} are not adjacent,
since $2-4=-2$ as is $3-5$.

From the maximal combination of adjacent links it is straightforwards to
construct a \emph{word cluster}.
\index{word cluster}%

\begin{definition}[Word cluster]
A \emph{word cluster} is a pair of subsentences $a^S_{i\ldots j}$
and $b^T_{k\ldots l}$ of the same length where $a^S_{i\ldots
j}=b^T_{k\ldots l}$ and $S[i-1]\not=T[k-1]$ and
$S[j+1]\not=T[l+1]$.
\end{definition}

Note that each maximal combination of adjacent links is
automatically a word cluster, but since sentences can sometimes
be aligned in different ways, a word cluster need not consists of
links.\footnote{Equal words in two sentences are only called
links when one is above the other \emph{in a certain alignment.}}

The subsentences in the form of word clusters describe parts of
the sentences that are equal. However, the unequal subsentences
are needed as hypotheses. Taking the complement of the word
clusters yields exactly the set of unequal subsentences.

\begin{definition}[Complement (of subsentences)]
\index{subsentence!complement of}%
The \emph{complement} of a list of subsentences $[a^S_{i_1\ldots
i_2}, a^S_{i_3\ldots i_4}, \ldots, a^S_{i_{n-1}\ldots i_n}]$ is
the list of non-empty subsentences $[b^S_{0\ldots i_1},
b^S_{i_2\ldots i_3}, \ldots b^S_{i_n\ldots |S|}]$.
\end{definition}

The definition of the complement of subsentences implies that
$b^S_{0\ldots i_1}+a^S_{i_1\ldots i_2}+b^S_{i_2\ldots
i_3}+a^S_{i_3\ldots i_4}+\cdots+a^S_{i_{n-1}\ldots
i_n}+b^S_{i_n\ldots |S|}=S$ and that $b^S_{0\ldots i_1}$ is not
present when $i_1=0$ and $b^S_{i_n\ldots |S|}$ is not present
when $i_n=|S|$.

To summarise, the function \texttt{FindSubstitutableSubsentences}
is implemented using the edit distance algorithm. This algorithm
finds a list of pairs of indices where words are equal in both
sentences. From this list, word clusters are constructed, which
describe subsentences that are equal in both sentences. The
complement of these subsentences is then returned as the result
of the function.

%%%%%%%%%%%%%%%%%%%%%%%%%%%%%%%%%%%%%%%%%%%%%%%%%%%%%%%%%%%%%%%%%%%%%%%%%%%%%%%%
\thesissubsubsection{The edit distance algorithm}
%%%%%%%%%%%%%%%%%%%%%%%%%%%%%%%%%%%%%%%%%%%%%%%%%%%%%%%%%%%%%%%%%%%%%%%%%%%%%%%%
\label{s:editdistancealgorithm}

The edit distance algorithm makes use of a technique called
\emph{dynamic programming} \citep{Bellman:DP-57}. ``For a problem
\index{Bellman, R.E.}%
\index{dynamic programming}%
to be solved by [the dynamic programming technique], it must be
capable of being divided repeatedly into subproblems in such a
way that identical subproblems arise again and again''
\citep{Russell:AI-95}.
\index{Russell, S.}%
\index{Norvig, P.}%

The dynamic programming approach consists of three components.
\begin{enumerate}
\item recurrence relation
\index{recurrence relation}%
\item tabular computation
\index{tabular computation}%
\item traceback
\index{traceback}%
\end{enumerate}

A recurrence relation describes a recursive relationship between
a solution and the solutions of its subproblems. In the case of
the edit distance this comes down to the following. When
computing the edit cost between sentences $A$ and $B$, $D(i, j)$
\index{edit cost}%
\index{cost function}%
\index{$\gamma$}%
denotes the edit cost of subsentences $u^A_{0\ldots i}$ and
$v^B_{0\ldots j}$. The recurrence relation is then defined as
\[
D(i, j) = \min\left(
\begin{array}{l}
D(i-1, j)+\gamma(A[i]\rightarrow\ew)\\
D(i, j-1)+\gamma(\ew\rightarrow B[j])\\
D(i-1, j-1)+\gamma(A[i]\rightarrow B[j])
\end{array}\right)
\]

In this relation, $\gamma(X\rightarrow Y)$ returns the edit cost
of \emph{substituting} $X$ into $Y$, where substituting $X$ into
the empty word ($\ew$) is the same as \emph{deleting} $X$ and
substituting the empty word into $Y$ means \emph{inserting} $Y$.
$\gamma(X\rightarrow X)$ does not normally count as a
substitution; the two words \emph{match}.

Next to the recurrence relation, base conditions are needed when
\index{base condition}%
no smaller indices exist. Here, the base conditions for $D(i, j)$
are $D(i, 0)=i*\gamma(A[i]\rightarrow\ew)$ and $D(0,
j)=j*\gamma(\ew\rightarrow B[j])$.  These base conditions mean
that it takes $i$ deletions to get from the subsentence
$u^A_{0\ldots i}$ to the empty sentence and $j$ insertions to
construct the subsentence $v^B_{0\ldots j}$ from the empty
sentence. Note that this implies that $D(0, 0)=0$.

Using the base conditions and the recurrence relation, a table is
filled with the edit costs $D(i, j)$. In each entry $(i, j)$ in
the matrix, the value $D(i, j)$ is stored.  Computing an entry in
the matrix (apart from the entries $D(i, 0) \mbox{ with } 0\leq
i\leq |A|$ and $D(0, j) \mbox{ with } 0\leq j\leq |B|$ which are
covered by the base conditions), is done using the recurrence
relation. The recurrence relation only uses information from the
direct left, upper, and upper-left entries in the matrix. An
overview of the algorithm that fills the matrix can be found in
algorithm~\ref{alg:edx}.

As an important side note, one would expect that if the order of
\index{order of sentences}%
\index{sentence!order of}%
the two sentences that are to be aligned is reversed, all INSs
will be DELs and vice versa. Matching words still match and
substituted words still need to be substituted, which would lead
to the same alignment and thus to the same substitutable
subsentences. However, reversing the sentences may in some
specific cases find different hypotheses. To see how this works,
consider the sentences in~\ref{cross}. If the algorithm chooses
to link \sent{Bert} (when the sentences are in this order) since
it occurs as the first word that can be linked in the first
sentence, it will choose to link \sent{Ernie} when the sentences
are given in the other order (second sentence first and first
sentence second), since \sent{Ernie} is the first word in the
first sentence that can be linked then. The problem lies in that
the computation of the minimum cost of the three edit operations
is done in a fixed order, where multiple edit operations can
return the same cost.

\eenumsentence{
\toplabel{cross}
\item \align{Bert} sees \dalign{Ernie}
\item \dalign{Ernie} kisses \align{Bert}
}

\begin{alg}{Edit distance: building the matrix}
\label{alg:edx}
\begin{programbox}
\keyword{func} \texttt{EditDistanceX}%
($A$, $B$: sentence): matrix\\
\mbox{\# $A$ and $B$ are the two sentences for which the edit cost will be %
         computed}\\
\keyword{var} $i$, $j$, $m_{sub}$, $m_{del}$, $m_{ins}$: integer\\
\>$D$: matrix\\
\keyword{begin}\\
\>$D[0, 0]:=0$\\
\>\keyword{for} $i:=1$ \keyword{to} $|A|$ \keyword{do}\\
\>\>$D[i, 0]:=D[i-1, 0]+\gamma(A[i]\rightarrow \ew)$\\
\>\keyword{od}\\
\>\keyword{for} $j:=1$ \keyword{to} $|B|$ \keyword{do}\\
\>\>$D[0, j]:=D[0, j-1]+\gamma(\ew\rightarrow B[j])$\\
\>\keyword{od}\\
\>\keyword{for} $i:=1$ \keyword{to} $|A|$ \keyword{do}\\
\>\>\keyword{for} $j:=1$ \keyword{to} $|B|$ \keyword{do}\\
\>\>\>$m_{sub}:=D[i-1, j-1]+\gamma(A[i]\rightarrow B[j])$\\
\>\>\>$m_{del}:=D[i-1, j]+\gamma(A[i]\rightarrow \ew)$\\
\>\>\>$m_{ins}:=D[i, j-1]+\gamma(\ew\rightarrow B[j])$\\
\>\>\>$D[i, j]:=\min(m_{sub}, m_{del}, m_{ins})$\\
\>\>\keyword{od}\\
\>\keyword{od}\\
\>\keyword{return} $D$\\
\keyword{end.}
\end{programbox}
\end{alg}

When the matrix is built, the entry $D(|A|, |B|)$ gives the
minimal edit cost to convert sentence $A$ into sentence $B$. This
gives a metric that indicates how different the two sentences
are. However, the matrix also contains information on the edit
transcript of $A$ and $B$.

The third component in dynamic programming, the traceback, finds
\index{traceback}%
\index{trace}%
an edit transcript which results in the minimum edit cost.
Algorithm~\ref{alg:edy} finds such a trace. A trace normally
describes which words should be deleted, inserted or substituted.
In algorithm~\ref{alg:edy}, only the links in the two sentences
are returned. Note that this is a slightly edited version of
\citet{Wagner:74-21-168}. The original algorithm incorrectly
\index{Wagner, R.A.}%
\index{Fischer, M.J.}%
printed words that are equal \emph{and} words that needed the
substitution operation.

\begin{alg}{Edit distance: finding a trace}
\label{alg:edy}
\begin{programbox}
\keyword{func} \texttt{EditDistanceY}%
($A$, $B$: sentence, $D$: matrix): set of pairs of indices\\
\mbox{\# $A$ and $B$ are the two sentences for which the edit cost will be %
         computed}\\
\mbox{\# $D$ is the matrix with edit cost information (build by %
         \texttt{EditDistanceX})}\\
\keyword{var} $i$, $j$: integer\\
\>$P$: set of pairs of indices\\
\keyword{begin}\\
\>$P:=\{\}$\\
\>$i:=|A|$\\
\>$j:=|B|$\\
\>\keyword{while} ($i\not=0$ \keyword{and} $j\not=0$) \keyword{do}\\
\>\>\keyword{if} ($D[i, j]=D[i-1, j]+\gamma(A[i]\rightarrow\ew)$) %
    \keyword{then}\\
\>\>\>$i:=i-1$\\
\>\>\keyword{elsif} ($D[i, j]=D[i, j-1]+\gamma(\ew\rightarrow B[j])$) %
    \keyword{then}\\
\>\>\>$j:=j-1$\\
\>\>\keyword{else}\\
\>\>\>\keyword{if} ($A[i]=B[j]$) \keyword{then}\\
\>\>\>\>$P:=P+\tuple{$i$, $j$}$\\
\>\>\>\keyword{fi}\\
\>\>\>$i:=i-1$\\
\>\>\>$j:=j-1$\\
\>\>\keyword{fi}\\
\>\keyword{od}\\
\>\keyword{return} $P$\\
\keyword{end.}
\end{programbox}
\end{alg}

Remember that from the links, the equal subsentences can be
found. Taking the complement of the equal subsentences yields the
substitutable subsentences.

%%%%%%%%%%%%%%%%%%%%%%%%%%%%%%%%%%%%%%%%%%%%%%%%%%%%%%%%%%%%%%%%%%%%%%%%%%%%%%%%
\thesissubsubsection{Default alignment learning}
%%%%%%%%%%%%%%%%%%%%%%%%%%%%%%%%%%%%%%%%%%%%%%%%%%%%%%%%%%%%%%%%%%%%%%%%%%%%%%%%

The general idea of using the edit distance algorithm to find substitutable
subsentences is described in the previous section. However, nothing has been
said about the cost function $\gamma$. 

$\gamma$ is defined to return the edit cost of its argument. For
\index{cost function}%
\index{$\gamma$}%
example, $\gamma(A[i]\rightarrow\ew)$ returns the cost of
deleting $A[i]$ (i.e.\ changing $A[i]$ into the empty word) or
$\gamma(A[i]\rightarrow B[j])$ is the cost of replacing $A[i]$ by
$B[j]$.

Following \citet{Wagner:74-21-168}, setting $\gamma$ to return
\index{Wagner, R.A.}%
\index{Fischer, M.J.}%
$1$ for the insertion and deletion operations and $2$ for the
substitution operation yields an algorithm that finds the
\emph{longest common subsequence} of two sentences. This common
\index{longest common subsequence}%
\index{subsequence!longest common}%
subsequence coincides with the notion of word cluster as
\index{word cluster}%
described in the previous section.

Implementing the alignment learning phase using algorithm~\vref{alg:al}
and algorithms~\ref{alg:edx} and~\ref{alg:edy} results in an
alignment instantiation which is called \emph{\ablone} when
$\gamma$ is defined as follows:
\index{default}%
\index{$\gamma$!default}%
\index{alignment learning!instances!default}%
\begin{itemize}
\addtolength{\itemsep}{-2.5pt}
\item $\gamma(X\rightarrow X)=0$
\item $\gamma(X\rightarrow\ew)=1$
\item $\gamma(\ew\rightarrow X)=1$
\item $\gamma(X\rightarrow Y)=2$
\end{itemize}

\newcommand{\hl}[1]{\multicolumn{1}{|r|}{#1}}
\newcommand{\hb}[1]{\multicolumn{1}{|r|}{\textbf{#1}}}
\newcommand{\hm}[1]{\multicolumn{1}{|r|}{#1}}
\begin{fig}{Example of a filled edit distance table}
\label{fig:fillededt}
\vspace{1em}
\begin{tabular}{r||lr|lr|lr|lr|lr|lr|}
& \multicolumn{2}{r|}{} &
  \multicolumn{2}{r|}{monsters} &
  \multicolumn{2}{r|}{like} &
  \multicolumn{2}{r|}{tuna} &
  \multicolumn{2}{r|}{fish} &
  \multicolumn{2}{r|}{sandwiches} \\\hline\hline
& &      & &      & &      & &      & &      & &      \\
\cline{3-3}\cline{5-5}\cline{7-7}\cline{9-9}\cline{11-11}\cline{13-13}
& &\hb{0}&1&\hl{1}&2&\hl{2}&3&\hl{3}&4&\hl{4}&5&\hm{5}\\\hline
& &    1 &2&    2 &3&    3 &4&    4 &5&    5 &6&    6 \\
\cline{3-3}\cline{5-5}\cline{7-7}\cline{9-9}\cline{11-11}\cline{13-13}
all
& &\hb{1}&2&\hl{2}&3&\hl{3}&4&\hl{4}&5&\hl{5}&6&\hm{6}\\\hline
& &    2 &1&    3 &4&    4 &5&    5 &6&    6 &7&    7 \\
\cline{3-3}\cline{5-5}\cline{7-7}\cline{9-9}\cline{11-11}\cline{13-13}
monsters
& &\hl{2}&3&\hb{1}&2&\hl{2}&3&\hl{3}&4&\hl{4}&5&\hm{5}\\\hline
& &    3 &4&    2 &1&    3 &4&    4 &5&    5 &6&    6 \\
\cline{3-3}\cline{5-5}\cline{7-7}\cline{9-9}\cline{11-11}\cline{13-13}
like
& &\hl{3}&4&\hl{2}&3&\hb{1}&2&\hl{2}&3&\hl{3}&4&\hm{4}\\\hline
& &    4 &5&    3 &4&    2 &3&    3 &4&    4 &5&    5 \\
\cline{3-3}\cline{5-5}\cline{7-7}\cline{9-9}\cline{11-11}\cline{13-13}
to
& &\hl{4}&5&\hl{3}&4&\hl{2}&3&\hb{3}&4&\hl{4}&5&\hm{5}\\\hline
& &    5 &6&    4 &5&    3 &4&    4 &3&    5 &6&    6 \\
\cline{3-3}\cline{5-5}\cline{7-7}\cline{9-9}\cline{11-11}\cline{13-13}
fish
& &\hl{5}&6&\hl{4}&5&\hl{3}&4&\hl{4}&5&\hb{3}&4&\hb{4}\\\hline
\end{tabular}
\end{fig}

Figure~\ref{fig:fillededt} is an example of the edit distance
algorithm with the $\gamma$ function as described above.  The
sentence \sent{Monsters like tuna fish sandwiches} is transformed
into \sent{All monsters like to fish}. The values in the upper
row and left column correspond to the base conditions.  The other
entries in the table have four values. The upper left value
describes the cost when substituting (or matching) the words in
that row and column plus the previous edit cost (found in the
entry to the north-west). The upper right entry describes the
edit cost of insertion plus the edit cost to the north. The lower
left value is the edit cost of deletion plus the edit cost to
the west. The lower right value is the minimum of the other three
values, which is also the value stored in the actual matrix.

The bold values describe an alignment. The transcript of this
\index{alignment}%
alignment is found when starting from the lower right entry in
the matrix and going back (following the bold value, which are
constantly the minimum values on that point in the matrix) to the
upper left entry. The transcript is then a DEL, MAT, SUB, MAT,
MAT, INS, which is the (reversed) transcript shown in
figure~\ref{fig:edittranscript}.

%%%%%%%%%%%%%%%%%%%%%%%%%%%%%%%%%%%%%%%%%%%%%%%%%%%%%%%%%%%%%%%%%%%%%%%%%%%%%%%%
\thesissubsubsection{Biased alignment learning}
%%%%%%%%%%%%%%%%%%%%%%%%%%%%%%%%%%%%%%%%%%%%%%%%%%%%%%%%%%%%%%%%%%%%%%%%%%%%%%%%
\label{s:biasedalignment}

Using the algorithm (and cost function) described above to find
the dissimilar parts of the sentences does not always result in
the preferred hypotheses.  As can be seen when aligning the parts
of the sentences in~\ref{ambiguous}, the \ablone algorithm
generates the alignment in~\ref{ambsanfrancisco}, because
\sent{Sesame Street} is the longest common subsequence. Linking
\index{longest common subsequence}%
\index{subsequence!longest common}%
\textit{Sesame Street} costs 4, while linking \textit{England}
(shown in the sentences in~\ref{ambdallas}) or \textit{to} (shown
in the sentences in~\ref{ambto}) costs 6.

\eenumsentence{
  \toplabel{ambiguous}
  \item \ldots from Sesame Street to England
  \item \ldots from England to Sesame Street
}
\eenumsentence{
  \toplabel{ambsanfrancisco}
  \item \ldots \align{from} [ ]\snt{2} \align{Sesame Street} [to
  England]\snt{3}
  \item \ldots \align{from} [England to]\snt{2} \align{Sesame
  Street}
[ ]\snt{3}
}
\eenumsentence{
  \toplabel{ambdallas}
  \item \ldots \align{from} [Sesame Street to]\snt{4}
  \align{England} [ ]\snt{5}
  \item \ldots \align{from} [ ]\snt{4} \align{England} [to Sesame
  Street]\snt{5}
}
\eenumsentence{
  \toplabel{ambto}
  \item \ldots \align{from} [Sesame Street]\snt{6} \align{to}
  [England]\snt{7}
  \item \ldots \align{from} [England]\snt{6} \align{to} [Sesame
  Street]\snt{7}
}

However, aligning \sent{Sesame Street} results in unwanted syntactic
\index{syntactic structure!unwanted}%
structures. Both \sent{to England} and \sent{England to} are considered
hypotheses. A more preferred alignment can be found when linking the word
\sent{to}.

This problem occurs every time the algorithm links words that are
``too far apart''. The relative distance between the two
\sent{Sesame Street}s in the two sentences is much larger than
the relative distance between the word \sent{to} in both
sentences.

This can be solved by biasing the $\gamma$ cost function towards
\index{$\gamma$!biased}%
\index{cost function!biased}%
linking words that have similar offsets in the sentence. The
\sent{Sesame Street}s reside in the beginning and end of the
sentences (the same applies to \sent{England} in the alignment of
sentences~\ref{ambdallas}), so the difference in offset is large.
This is not the case for \sent{to}; both reside roughly in the
middle.

An alternative cost function may be biased towards linking words
\index{cost function}%
\index{cost function!biased}%
that have a small relative distance. This can be accomplished by
letting the cost function return a high cost when the difference
of the relative offsets of the words is large. The relative
distance between the two \textit{Sesame Street}s in
sentences~\ref{ambsanfrancisco} is larger compared to the
relative distance between the two \textit{to}s in
sentences~\ref{ambto}.  Therefore the total edit cost of
sentences~\ref{ambto} will be less than the edit cost of
sentences~\ref{ambsanfrancisco} or sentences~\ref{ambdallas}.

In the system called \emph{\abldis}, the $\gamma$ function will
\index{biased}%
\index{alignment learning!instances!biased}%
be changed as follows:
\begin{itemize}
\item $\displaystyle\gamma(X\rightarrow X)=
\left|\frac{i^S_X}{|S|}-
\frac{i^T_X}{|T|}\right|*\mbox{mean}(|S|, |T|)$ where $i^U_W$ is
the index of word $W$ in sentence $U$ and $S$ and $T$ are the two
sentences.
\item $\gamma(X\rightarrow\ew)=1$
\item $\gamma(\ew\rightarrow X)=1$
\item $\gamma(X\rightarrow Y)=2$
\end{itemize}

If the parts of the sentences shown in~\ref{ambiguous} are the
complete sentences, the edit transcription with minimum edit cost
of the alignment in~\ref{ambsanfrancisco} is [MAT, INS, INS, MAT,
MAT, DEL, DEL] with costs: $0+1+1+2+2+1+1=8$, for the sentences
in~\ref{ambdallas} this is [MAT, DEL, DEL, DEL, MAT, INS, INS,
INS] with costs: $0+1+1+1+3+1+1+1=9$ and the sentences
in~\ref{ambto} become [MAT, SUB, DEL, MAT, SUB, INS] with costs:
$0+2+1+1+2+1=7$.  Therefore, the alignment with \sent{to} is
chosen.

The \abldis system does not entirely solve the problem, since in
sentences similar to those in~\ref{ambproblem} the words
\sent{from} and \sent{Sesame Street} are linked (instead of the
words \sent{from} and \sent{to} as preferred).
\eenumsentence{
\toplabel{ambproblem}
\item \sent{\ldots \align{from} England to \align{Sesame Street}}
\item \sent{\ldots \align{from Sesame Street} where Big Bird
lives to England}
}

Additionally, the \abldis system will find less hypotheses, since
less matches will be found compared to the \ablone system. Where
the \ablone system still matched words that are relatively far
apart, the \abldis system will not match them. Less links and
thus less word groups and hypotheses will be found.

%%%%%%%%%%%%%%%%%%%%%%%%%%%%%%%%%%%%%%%%%%%%%%%%%%%%%%%%%%%%%%%%%%%%%%%%%%%%%%%%
\thesissubsection{Alignment learning with all alignments}
%%%%%%%%%%%%%%%%%%%%%%%%%%%%%%%%%%%%%%%%%%%%%%%%%%%%%%%%%%%%%%%%%%%%%%%%%%%%%%%%

Another solution to the problem of introducing incorrect
hypotheses is to simply generate all possible alignments (using
an alignment algorithm that is not based on the edit distance
algorithm).\footnote{It is also possible to implement this using
an adapted version of the edit distance algorithm that keeps
track of all possible alignments using a matrix with pointers.}
This method is called \emph{\ablall.}
\index{all}%
\index{alignment learning!instances!all}%

When all possible alignments are considered, the selection
learning phase of \abl has a harder job selecting the best
hypotheses. Since the alignment learning finds more alignments,
more hypotheses are inserted into the hypothesis space. And thus
the selection learning phase has more to choose from. However,
because the alignment learning phase does not know which are the
correct hypotheses to insert, inserting all of them might be the
best option.

Algorithm~\ref{alg:allalign} finds all possible alignments
between two sentences. The function \texttt{AllAlignments} takes
two sentences as arguments. The first step in the algorithm finds
a list of all matching terminals. This list is generated in the
function \texttt{FindAllMatchingTerminals}. This list contains
pairs of indices of words in the two sentences that can be linked
(i.e.\ words that might be a link in an alignment). Note that this
list \emph{can} contain crossing links (in contrast to the links
found by the edit distance algorithm).

Next, the algorithm incrementally adds each of these links into
all possible alignments. If inserting a link in an alignment
results in overlapping links within that alignment (which is not
allowed), a new alignment is introduced. In
algorithm~\ref{alg:allalign} this is the case when
$O\not=\emptyset$. The first alignment is unchanged (i.e.\ the
current link is not inserted): $P:=P+(j)$ and the current link is
added to the new alignment, which contains all links from the
first alignment which do not overlap with the current link:
$P:=P+(j-O+i)$. This might insert alignments that are proper
subsets of other alignments in $P$, so before returning, these
subsets need to be filtered away.

Since variable $P$ is a \emph{set}, no duplicates are introduced, so when all
links are appended to the possible alignments, $P$ contains all possible
alignments in the two sentences.

\begin{alg}{Finding all possible alignments}
\label{alg:allalign}
\begin{programbox}
\keyword{func} \texttt{AllAlignments}%
($A$, $B$: sentence): set of pairs of indices\\
\mbox{\# $A$ and $B$ are the two sentences for which all alignments %
         will be computed}\\
\keyword{var} $M$, $O$, $j$: list of pairs of integers,\\
\>$P$, $P_{old}$: set of lists of pairs of integers,\\
\>$i$, $e$: pair of integers\\
\keyword{begin}\\
\>$M:={}$\texttt{FindAllMatchingTerminals}($A$, $B$)\\
\>$P:=\{[ ]\}$\`\mbox{\# $P$ is a singleton set with an empty list}\\
\>\keyword{foreach} $i\in M$ \keyword{do}\\
\>\>$P_{old}:=P$\\
\>\>$P:=\{\}$\\
\>\>\keyword{foreach} $j\in P_{old}$ \keyword{do}\\
\>\>\>$O:=\{e\in j:(e[0]\leq i[0]$ \keyword{and} $e[1]\geq i[1])$ \keyword{or}\\
\>\>\>\>\>\>\>\>$(e[0]\geq i[0]$ \keyword{and} $e[1]\leq i[1])\}$\\
\>\>\>\keyword{if} ($O=\emptyset$) \keyword{then}\`\mbox{\# O is the set of
links in $j$ overlapping $i$}\\
\>\>\>\>$P:=P+(j+i)$\`\mbox{\# Add the list $j$ with $i$ inserted to $P$}\\
\>\>\>\keyword{else}\\
\>\>\>\>$P:=P+(j)$\`\mbox{\# Add the list $j$ to $P$}\\
\>\>\>\>$P:=P+(j-O+i)$\`\mbox{\# Add the list $(j-O+i)$ to $P$}\\
\>\>\>\keyword{fi}\\
\>\>\keyword{od}\\
\>\keyword{od}\\
\>\keyword{foreach} $k\in P$ \keyword{do}\`\mbox{\# Filter subsets
from $P$}\\
\>\>\keyword{if} ($k \subset l \in P$) \keyword{then}\\
\>\>\>$P:=P-k$\\
\>\>\keyword{fi}\\
\>\keyword{od}\\
\>\keyword{return} $P$\\
\keyword{end.}
\end{programbox}
\end{alg}

%%%%%%%%%%%%%%%%%%%%%%%%%%%%%%%%%%%%%%%%%%%%%%%%%%%%%%%%%%%%%%%%%%%%%%%%%%%%%%%%
%%%%%%%%%%%%%%%%%%%%%%%%%%%%%%%%%%%%%%%%%%%%%%%%%%%%%%%%%%%%%%%%%%%%%%%%%%%%%%%%
\thesissection{Selection learning instantiations}
%%%%%%%%%%%%%%%%%%%%%%%%%%%%%%%%%%%%%%%%%%%%%%%%%%%%%%%%%%%%%%%%%%%%%%%%%%%%%%%%
%%%%%%%%%%%%%%%%%%%%%%%%%%%%%%%%%%%%%%%%%%%%%%%%%%%%%%%%%%%%%%%%%%%%%%%%%%%%%%%%
\label{s:selectioninst}

Not only the alignment learning phase can have different
instantiations.  Selection learning can also be done in different
\index{selection learning!instances}%
ways. Several different methods will be described here. First,
there is the simple, non-probabilistic method as described in
chapter~\ref{ch:learningalignment}. Next, two probabilistic
methods will be described. These probabilistic selection learning
methods differ in the way probabilities of hypotheses are
computed.

%%%%%%%%%%%%%%%%%%%%%%%%%%%%%%%%%%%%%%%%%%%%%%%%%%%%%%%%%%%%%%%%%%%%%%%%%%%%%%%%
\thesissubsection{Non-probabilistic selection learning}
%%%%%%%%%%%%%%%%%%%%%%%%%%%%%%%%%%%%%%%%%%%%%%%%%%%%%%%%%%%%%%%%%%%%%%%%%%%%%%%%

Remember that the selection learning phase should take care that no
overlapping hypotheses remain in the structure generated by the alignment
learning phase. The easy solution to this problem is to make sure that
overlapping hypotheses are never even introduced. In other words, if at some
point, the system tries to insert a hypothesis that overlaps with a hypothesis
that was already present, it is rejected.

The underlying assumption in this non-probabilistic method is that a
\index{selection learning!instances!non-probabilistic}%
hypothesis that is learned earlier is always correct. This means
that newly learned hypotheses that overlap with older ones are
incorrect, and thus should be removed. This method is called
\emph{\first.}
\index{incr}%
\index{selection learning!instances!incr}%

The advantage of this method is that it is very easy to
incorporate in the system as described in
section~\vref{s:removeoverlapping}. The main disadvantage is that
once an incorrect hypothesis has been learned, it can never be
corrected. The incorrect hypothesis will always remain in the
hypothesis space and will in the end be converted into an
incorrect constituent.\footnote{Since hypotheses learned earlier
may block certain (overlapping) hypotheses from being stored,
changing the order of the sentences in the corpus may change the
resulting treebank.}

The assumption that hypotheses learned earlier are correct may
perhaps be only likely for human language learning. It so happens
that the type of sentences in a corpus are generally not
comparable to the type of sentences human hear when they are
learning a language.  Furthermore, the implication that once an
incorrect hypothesis has been learned, it can never be corrected,
is (cognitively) highly implausible.

%%%%%%%%%%%%%%%%%%%%%%%%%%%%%%%%%%%%%%%%%%%%%%%%%%%%%%%%%%%%%%%%%%%%%%%%%%%%%%%%
\thesissubsection{Probabilistic selection learning}
%%%%%%%%%%%%%%%%%%%%%%%%%%%%%%%%%%%%%%%%%%%%%%%%%%%%%%%%%%%%%%%%%%%%%%%%%%%%%%%%
\label{s:probsellearn}

To solve the disadvantage of the first method, probabilistic
selection learning methods have been implemented. The
probabilistic selection learning methods select the combination
\index{selection learning!instances!probabilistic}%
of hypotheses with the highest combined probability. These
methods are accomplished after the alignment learning phase,
since more specific information (in the form of better counts)
can be found at that time.

First of all, the probability of each hypothesis has to be
computed. This can be accomplished in several ways. Here, two
different methods will be considered. By combining the
probabilities of the single hypotheses, the combined probability
of a set of hypotheses can be found.

Different ways of computing the probability of a hypothesis will
be discussed first, followed by a description on how the
probability of a combination of hypotheses can be computed.

%%%%%%%%%%%%%%%%%%%%%%%%%%%%%%%%%%%%%%%%%%%%%%%%%%%%%%%%%%%%%%%%%%%%%%%%%%%%%%%%
\thesissubsubsection{The probability of a hypothesis}
%%%%%%%%%%%%%%%%%%%%%%%%%%%%%%%%%%%%%%%%%%%%%%%%%%%%%%%%%%%%%%%%%%%%%%%%%%%%%%%%

The probability of a hypothesis is the chance that a hypothesis
\index{probability!of a hypothesis}%
\index{hypothesis!probability of}%
is drawn from the entire space of possible hypotheses. If it is
assumed that the alignment learning phase generates the space of
hypotheses, the probability of a hypothesis can be computed by
counting the number of times that the specific hypothesis occurs
in hypotheses generated by alignment learning. The hypotheses
generated by alignment learning make up the hypothesis
universe.\footnote{Remember
(definition~\ref{def:hypothesisspace}) that a hypothesis
\emph{space} is a set of fuzzy trees.  The hypothesis
\emph{universe} is the combination of the hypotheses of all fuzzy
trees in a certain hypothesis space.}

\begin{definition}[Hypothesis universe]
\index{hypothesis!universe}%
The \emph{hypothesis universe} is the union of all sets of hypotheses of the
fuzzy trees in a hypothesis space. In other words, the hypothesis universe
$U^D$ of hypothesis space $D$ is $U^D=\bigcup_{F=\tuple{$S_F$, $H_F$}\in
D}H_F$.
\end{definition}

The probability that a certain hypothesis $h$ occurs in a hypothesis universe
$D$ is its relative frequency under the assumption of the uniform
distribution:
\[
P^U(h)\stackrel{\mathrm{def}}{=}\frac{|h|}{|U|}
\]

All hypotheses in a hypothesis universe are unique. Within the
\index{hypothesis!unique}%
set of hypotheses of a fuzzy tree they are unique, and all are
indexed with their fuzzy tree, so hypotheses of different fuzzy
trees can never be the same. This means that the previous formula
can be rewritten as:
\[
P^U(h)=\frac{1}{|U|}
\]
The numerator becomes 1, since each hypothesis only occurs once and the
denominator is exactly the total number of hypotheses in the hypothesis
universe. The probability of a hypothesis is the same for all hypotheses.

The fact that each hypothesis has an equal probability of
occurring in the hypothesis universe, does not help in selecting the
better hypotheses.  Hypotheses receive the same probabilities,
because all hypotheses are unique.

Even though hypotheses \emph{are} unique, it can be said that
certain hypotheses are equal when concentrating on only certain
aspects of the hypotheses. Instead of obliging all properties of
two hypotheses to be the same, taking only certain properties of
a hypothesis into account when deciding which hypotheses are the
same relaxes the equality relation.

$P^U(h)$ describes the probability of hypothesis $h$ in the
hypothesis universe $U$. By grouping hypotheses with a certain
property, the probability of a hypothesis is calculated relative
to the subset of hypotheses that all have that same property. One
way of grouping hypotheses is by their yield. 

\begin{definition}[Yield of a hypothesis]
\index{hypothesis!yield of}%
\index{yield of hypothesis}%
The \emph{yield of a hypothesis} $yield(h^S)$ is the list of
words (in the form of a subsentence) grouped together by the
hypothesis. If $h^S=\tuple{$b$, $e$, $n$}$ then
$yield(h^S)=v^S_{b\ldots e}$.
\end{definition}

The first probabilistic method computes the probability of a
hypothesis by counting the number of times the \emph{subsentence}
described by the hypothesis has occurred as a hypothesis in the
hypothesis universe, normalised by the total number of
hypotheses. Thus the probability of a hypothesis $h$ in
hypothesis universe $U$ can be computed using the following
formula.
\[
P^U_{\mathit{\terms}}(h)=
\frac{|h'\in U:yield(h')=yield(h)|}{|U|}
\]

This method is called \emph{\terms{}} since we count the number of
\index{selection learning!instances!leaf}%
\index{leaf}%
times the leaves (i.e.\ the words) of the hypothesis co-occur in
the hypothesis universe as hypotheses.

The second method relaxes the equality relation in a different
way. In addition to comparing the words of the sentence delimited
by the hypothesis (as in the \terms method), this model computes
the probability of a hypothesis based on the words of the
hypothesis \emph{and} its type label (the function $root$ returns
the type label of a hypothesis). This model is effectively a
normalised version of $P_{\mathit{\terms}}$. This probabilistic
method of computing the probability of a hypothesis is called
\index{selection learning!instances!branch}%
\index{branch}%
\emph{\const}.
\[
P^U_{\mathit{\const}}(h)=
\frac{|h'\in U:yield(h')=yield(h)\wedge root(h')=root(h)|}
{|h''\in U: root(h'')=root(h)|}
\]

First, a partition of the hypothesis space is made by considering
\index{hypothesis!space!partition of}%
only hypotheses with a certain type label. In other words, the
hypothesis universe $U$ as used in $P_{\mathit{\terms}}$ is
partitioned into parts bounded by all possible type labels. For a
certain hypothesis $h$ this becomes: $U'_h=\{h'\in
U:root(h')=root(h)\}$, the set of hypotheses where the root is
the same as the root of $h$. Substituting $U'_h$ (where $h$ is
the hypothesis for which the probability is computed) for $U$ in
$P_{\mathit{\terms}}$ yields $P_{\mathit{\const}}$.

The two methods just described are not the only possible
approaches. Another interesting method, for example, could take
into account the inner structure of a hypothesis. In other words,
\index{selection learning!instances!inner structure}%
the probability of a hypothesis not only depends on the words in
its yield, but also on other hypotheses that occur within the
part of the sentence encompassed by the hypothesis. Such an
approach will be described in
section~\vref{s:parsingselectionlearning}.\footnote{This yields a
(stochastically) more precise model (taking also the information
of non-terminals into account). However, the information
contained in the hypothesis space is unreliable, since the
alignment learning phase also inserts incorrect hypotheses into
the hypothesis universe. In other words, the model tries to give
more precise results based on more imprecise information from the
hypothesis universe.}

%%%%%%%%%%%%%%%%%%%%%%%%%%%%%%%%%%%%%%%%%%%%%%%%%%%%%%%%%%%%%%%%%%%%%%%%%%%%%%%%
\thesissubsubsection{The probability of a combination of hypotheses}
%%%%%%%%%%%%%%%%%%%%%%%%%%%%%%%%%%%%%%%%%%%%%%%%%%%%%%%%%%%%%%%%%%%%%%%%%%%%%%%%

The previous section described two ways to compute the
probability of a hypothesis. Using these probabilities, it is
\index{probability!of a combination of hypotheses}
possible to calculate the probability of a combination of
hypotheses. The combination of hypotheses with the highest
probability is then chosen to be the ``correct'' combination.
This section describes how the probability of a combination of
hypotheses can be computed using the probabilities of the
separate hypotheses.

Since each selection of a hypothesis from the hypothesis universe
is independent of the previous selections, the probability of a
combination of hypotheses is the product of the probabilities of
the constituents as in \scfg{}s (cf.\ \citep{Booth:ASSA69-74}).
\index{Booth, T.}%
\index{SCFG}%
This means that if the separate probabilities of the hypotheses
are known, the combined probability of hypotheses $h_1, h_2,
\ldots, h_n$ can be computed as follows:
\[
P_{\mathit{\scfg}}(h_1, h_2, \ldots, h_n)=
\prod_{i=1}^n P(h_i)
\]

However, using the product of the probabilities of hypotheses
results in a \emph{trashing} effect, since the product of
\index{trashing effect}%
hypotheses is always smaller than or equal to the separate
probabilities. Since probabilities are all between 0 and 1,
multiplying many probabilities tends to reduce the combined
probability towards 0. Consider comparing the probability of the
singleton set of the hypothesis $\{h_1\}$ with probability
$P_{h_1}=P(h_1)$ to the set of hypotheses $\{h_1, h_2\}$ with
probability $P_{h_1, h_2}=P(h_1)P(h_2)$. It will always hold that
$P_{h_1, h_2}\leq P_{h_1}$. Thus in general, taking the product
of probabilities prefers smaller sets of hypotheses.

To eliminate this effect, a normalised method of computing the
combined probability is used. According to
\citet{Caraballo:98-24-275}, the geometric mean reduces the
\index{Caraballo, S.A.}%
\index{Charniak, E.}%
\index{geometric mean}%
trashing effect. Therefore, the probability of a set of
constituents $h_1, \ldots, h_n$ is computed using:
\[
P_{\mathit{GM}}(h_1, \ldots, h_n)=
\sqrt[n]{\prod_{i=1}^n P(h_i)}
\]
The probability of a certain hypothesis $h$ can be computed using
one of the methods described in the previous section
($P_{\mathit{\terms}}$ or $P_{\mathit{\const}}$).

In practice, many probabilities may be multiplied, which can
result in a numerical underflow. To solve this, the logprob
\index{logprob}%
\index{$-\log$}%
(i.e.\ the $-\log$) of the probabilities is used. The geometric
mean can be rewritten (where LP denotes the logprob) as:
\[
LP_{\mathit{GM}}(h_1, \ldots, h_n)=
\frac{\sum_{i=1}^n LP(h_i)}{n}
\]
which shows that the geometric mean actually computes the mean of
the logprob of the hypotheses.

Using this formula effectively selects only those hypotheses that
have the highest probability (in the set). Assume the mean of a
\index{hypothesis!with highest probability}%
set of values is computed. If a value higher than that mean is
added to the set, the newly computed mean will be higher, while
adding a value lower than the mean will lower the resulting mean.
This means that the set with (only) the lowest values will be
chosen when looking for the minimum mean. If there are more
elements that have the same value, any combination of these
elements result in the same mean value (the mean of for example
[3, 3] is the same as the mean of [3]).

Although using the $P_{\mathit{GM}}$ (or the equivalent
$LP_{\mathit{GM}}$) method eliminates the trashing effect, it
does not have a preference for richer structures when there are
two (or more) combinations of hypotheses that have the same
probability.

To let the system have a preference for more constituents in the
final tree structure when there are more possibilities with the
same probability, the \emph{extended geometric mean} is
\index{geometric mean!extended}%
\index{extended geometric mean}%
implemented. The only difference with the (standard) geometric
mean is that when there are more possibilities (single hypothesis
or combinations of hypotheses) with the same probability, this
system selects the combination with the most hypotheses. To
indicate that the systems use the extended geometric mean, a
${}^+$ is added to the name of the methods that use the extended
geometric mean. For example, using the \terms method to compute
the probabilities of hypotheses and the extended geometric mean
to compute the combined probability is called \termsl (and
\index{selection learning!instances!leaf\ensuremath{{}^+}}%
\index{leaf\ensuremath{{}^+}}%
similarly \constl when using \const). If there are sets of
\index{selection learning!instances!branch\ensuremath{{}^+}}%
\index{branch\ensuremath{{}^+}}%
hypotheses that have the same probability and the same number of
hypotheses, one of them is chosen at random.

The properties of the geometric mean also explain why only the
set of \emph{overlapping} hypotheses is considered when computing
\index{hypothesis!overlapping}%
the most probable structure of the fuzzy tree. In other words,
\index{fuzzy tree}%
\index{tree!fuzzy}%
hypotheses that do not overlap any other hypothesis in the fuzzy
tree are always considered correct. If all hypotheses were
considered, only the hypotheses with the highest probability will
be selected. This means that many correct hypotheses will be
thrown away. (It is very probable that the hypothesis with the
start symbol will not be selected, since the probability of that
hypothesis is very small; it only occurs as often as the entire
sentence occurs in the corpus.)

Previous publications also mentioned methods that used the \terms
and \const methods of computing the probabilities of hypotheses
and the (standard) geometric mean to compute the combined
probabilities. These systems however did not prefer more richly
structured trees. They chose a random solution if multiple
solutions were found. Since these systems will always learn the
same amount of or less structure, they will not be considered here. 

The probability of each combination of mutually non-overlapping
hypotheses is computed using a Viterbi style optimisation
algorithm \citep{Viterbi:67-13-260}. The combination of
\index{Viterbi, A.}%
hypotheses with the highest probability (and the lowest logprob)
is selected and the overlapping hypotheses not present in this
combination are removed from the fuzzy tree.

%%%%%%%%%%%%%%%%%%%%%%%%%%%%%%%%%%%%%%%%%%%%%%%%%%%%%%%%%%%%%%%%%%%%%%%%%%%%%%%%
%%%%%%%%%%%%%%%%%%%%%%%%%%%%%%%%%%%%%%%%%%%%%%%%%%%%%%%%%%%%%%%%%%%%%%%%%%%%%%%%
\thesissection{Grammar extraction instantiations}
%%%%%%%%%%%%%%%%%%%%%%%%%%%%%%%%%%%%%%%%%%%%%%%%%%%%%%%%%%%%%%%%%%%%%%%%%%%%%%%%
%%%%%%%%%%%%%%%%%%%%%%%%%%%%%%%%%%%%%%%%%%%%%%%%%%%%%%%%%%%%%%%%%%%%%%%%%%%%%%%%
\label{s:grammarextract}%

When the selection learning phase has disambiguated the fuzzy
trees, regular tree structures remain. From these tree
structures, it is possible to extract a grammar. First, the focus
\index{grammar!extraction}%
is on extracting a stochastic context-free grammar (\scfg), since
\index{SCFG}%
\index{stochastic context-free grammar}%
\index{context-free grammar!stochastic}%
the underlying grammar of the final treebank is considered
context-free. Next, extracting a stochastically stronger grammar
type, stochastic tree substitution grammar (\stsg), is described. 
\index{STSG}%
\index{stochastic tree substitution grammar}%
\index{tree substitution grammar!stochastic}%

%%%%%%%%%%%%%%%%%%%%%%%%%%%%%%%%%%%%%%%%%%%%%%%%%%%%%%%%%%%%%%%%%%%%%%%%%%%%%%%%
\thesissubsection{Extracting a stochastic context-free grammar}
%%%%%%%%%%%%%%%%%%%%%%%%%%%%%%%%%%%%%%%%%%%%%%%%%%%%%%%%%%%%%%%%%%%%%%%%%%%%%%%%
\label{s:extractscfg}%

In general, it is possible to extract grammar rules from tree
structures (i.e.\ parsed sentences). These grammar rules can
generate the tree structures it was extracted from. In other
words, parsing the sentence with the extracted grammar can return
\index{extract!stochastic context-free grammar}%
the same structure.\footnote{Since an ambiguous grammar may be
extracted from a set of tree structures, it is not necessarily
the case that a structure equal to the original is assigned to
the sentence.}

Imagine the structured sentences as displayed in
figure~\ref{fig:cfgtree}.  These two sentences can also be
written as tree structures. Extracting a context-free grammar
from the tree structures is rather straightforward. For each node
(non-terminal in the tree structure), take the label as the
left-hand side of a grammar rule. The list of direct daughters of
the node are the right-hand side of the grammar rule.

When the extracted context-free grammars are stored in a bag, the
probabilities of the grammar rules can easily be computed, which
converts the context-free grammar in a stochastic version. The
probability of a context-free grammar rule is the number of times
the specific grammar rule occurs in the bag, divided by the total
number of grammar rules with the same non-terminal on its
left-hand side. Extracting a stochastic context-free grammar from
the two trees, results in the \scfg in figure~\ref{fig:cfgtree}.

\begin{fig}{Extracting an \scfg from a tree structure}
\label{fig:cfgtree}
[[Bert]\snt{NP}[[sees]\snt{V}[Ernie]\snt{NP}]\snt{VP}]\snt{S}\\{}
[[Ernie]\snt{NP}[[walks]\snt{V}]\snt{VP}]\snt{S}
\begin{multicols}{2}
\psset{nodesep=2pt,levelsep=35pt}
\pstree{\TR{\nt{S}}}
{\pstree{\TR{\nt{NP}}}{\TR{Bert}}
 \pstree{\TR{\nt{VP}}}{\pstree{\TR{\nt{V}}}{\TR{sees}}
                       \pstree{\TR{\nt{NP}}}{\TR{Ernie}}
                      }
}

\pstree{\TR{\nt{S}}}
{\pstree{\TR{\nt{NP}}}{\TR{Ernie}}
 \pstree{\TR{\nt{VP}}}{\pstree{\TR{\nt{V}}}{\TR{walks}}}
}
\end{multicols}
\begin{tabular}{l@{ \der\ }lr}
\nt{S}   & \nt{NP} \nt{VP}   & 2/2=1 \\
\nt{VP}  & \nt{V} \nt{NP}    & 1/2=0.5 \\
\nt{VP}  & \nt{V}            & 1/2=0.5 \\
\nt{NP}  & Bert              & 1/3=0.33 \\
\nt{NP}  & Ernie             & 2/3=0.67 \\
\nt{V}   & sees              & 1/2=0.5 \\
\nt{V}   & walks             & 1/2=0.5
\end{tabular}
\end{fig}

When all grammar rules are extracted from the tree structure, the
probabilities of the grammar rules in the grammar are then computed using the
following formula:
\[
P^G(s)=\frac{|s'\in G:LHS(s')=LHS(s)\wedge RHS(s')=RHS(s)|}
{|s''\in G:LHS(s'')=LHS(s)|}
\]
In this formula, $LHS(x)$ denotes the left-hand side of grammar rule $x$,
while $RHS(x)$ denotes the right-hand side of $x$.

%%%%%%%%%%%%%%%%%%%%%%%%%%%%%%%%%%%%%%%%%%%%%%%%%%%%%%%%%%%%%%%%%%%%%%%%%%%%%%%%
\thesissubsection{Extracting a stochastic tree substitution grammar}
%%%%%%%%%%%%%%%%%%%%%%%%%%%%%%%%%%%%%%%%%%%%%%%%%%%%%%%%%%%%%%%%%%%%%%%%%%%%%%%%
\label{s:extractstsg}

The previous section showed how a stochastic context-free grammar
can be extracted from a tree structure. This section will
concentrate on extracting a stochastic tree substitution grammar
\index{extract!stochastic tree substitution grammar}%
(\stsg). This type of grammar can generate the same tree
structures as a context-free grammar, but is stochastically
stronger \citep{Bod:BG-98}.
\index{Bod, R.}%

When grammar rules, called \emph{elementary trees} in an \stsg, are
\index{elementary trees}%
extracted from a tree $T$, each grammar rule $t$ is in the form
such that:
\begin{itemize}
\item $t$ consists of more than one node
\item $t$ is connected
\item except for frontier nodes of $t$, each node in $t$ has the same daughter
nodes as the corresponding node in $T$
\end{itemize}

An example of the elementary trees in a tree structure can be
found in figure~\ref{fig:elemtrees}. This figure contains all
elementary trees that can be extracted from the first tree in the
figure. Note that the first tree is also an elementary tree (of
itself).

\newcommand{\lpstree}{\pstree[xbbl=.4cm,xbbr=.4cm,xbbd=.2cm,xbbh=.2cm]}
\begin{fig}{Example elementary trees}
\label{fig:elemtrees}
{\small
\psset{nodesep=2pt,levelsep=35pt}
\lpstree{\TR{\nt{S}}}
{\pstree{\TR{\nt{NP}}}{\TR{Bert}}
 \pstree{\TR{\nt{VP}}}{\pstree{\TR{\nt{V}}}{\TR{sees}}
                       \pstree{\TR{\nt{NP}}}{\TR{Ernie}}
                      }
}
\lpstree{\TR{\nt{S}}}
{\pstree{\TR{\nt{NP}}}{\TR{Bert}}
 \pstree{\TR{\nt{VP}}}{\TR{\nt{V}}
                       \pstree{\TR{\nt{NP}}}{\TR{Ernie}}
                      }
}
\lpstree{\TR{\nt{S}}}
{\pstree{\TR{\nt{NP}}}{\TR{Bert}}
 \pstree{\TR{\nt{VP}}}{\pstree{\TR{\nt{V}}}{\TR{sees}}
                       \TR{\nt{NP}}
                      }
}\\
\lpstree{\TR{\nt{S}}}
{\pstree{\TR{\nt{NP}}}{\TR{Bert}}
 \pstree{\TR{\nt{VP}}}{\TR{\nt{V}}
                       \TR{\nt{NP}}
                      }
}
\lpstree{\TR{\nt{S}}}
{\pstree{\TR{\nt{NP}}}{\TR{Bert}}
 \TR{\nt{VP}}
}
\lpstree{\TR{\nt{S}}}
{\TR{\nt{NP}}
 \pstree{\TR{\nt{VP}}}{\pstree{\TR{\nt{V}}}{\TR{sees}}
                       \pstree{\TR{\nt{NP}}}{\TR{Ernie}}
                      }
}
\lpstree{\TR{\nt{S}}}
{\TR{\nt{NP}}
 \pstree{\TR{\nt{VP}}}{\TR{\nt{V}}
                       \pstree{\TR{\nt{NP}}}{\TR{Ernie}}
                      }
}\\
\lpstree{\TR{\nt{S}}}
{\TR{\nt{NP}}
 \pstree{\TR{\nt{VP}}}{\pstree{\TR{\nt{V}}}{\TR{sees}}
                       \TR{\nt{NP}}
                      }
}
\lpstree{\TR{\nt{S}}}
{\TR{\nt{NP}}
 \pstree{\TR{\nt{VP}}}{\TR{\nt{V}}
                       \TR{\nt{NP}}
                      }
}
\lpstree{\TR{\nt{S}}}
{\TR{\nt{NP}}
 \TR{\nt{VP}}
}
\lpstree{\TR{\nt{VP}}}
{\pstree{\TR{\nt{V}}}{\TR{sees}}
 \pstree{\TR{\nt{NP}}}{\TR{Ernie}}
}\\
\lpstree{\TR{\nt{VP}}}
{\TR{\nt{V}}
 \pstree{\TR{\nt{NP}}}{\TR{Ernie}}
}
\lpstree{\TR{\nt{VP}}}
{\pstree{\TR{\nt{V}}}{\TR{sees}}
 \TR{\nt{NP}}
}
\lpstree{\TR{\nt{VP}}}
{\TR{\nt{V}}
 \TR{\nt{NP}}
}
\lpstree{\TR{\nt{NP}}}{\TR{Bert}}
\lpstree{\TR{\nt{V}}}{\TR{sees}}
\lpstree{\TR{\nt{NP}}}{\TR{Ernie}}
}
\end{fig}

The description of what an elementary trees is, does not give a
procedure to actually find these. \citet[p.\ 49]{vanZaanen:ECD-97}
\index{van Zaanen, M.M.}%
gives a (slightly informal) description of an algorithm that
finds all elementary trees in a given tree.

As can be seen in figure~\ref{fig:elemtrees}, extracting
elementary trees from one tree structure can result in many
elementary trees. When the number of trees from which elementary
trees are extracted becomes larger, the number of elementary
trees becomes huge. To keep the number of elementary trees within
practical limits, it is possible to set a condition on the
maximum depth of the elementary trees \cite[p.\
51]{vanZaanen:ECD-97}.
\index{van Zaanen, M.M.}%

The probabilities of the elementary trees are computed similarly
to the probabilities of the context-free grammar rules in an
\scfg:
\[
P^G(s)=\frac{|s'\in G:s'=s|}
{|s''\in G:root(s'')=root(s)|}
\]
Again, $root(x)$ denotes the root type label of the elementary tree $x$.

%% end of file: instantiations.tex

%% file: results.tex
%%%%%%%%%%%%%%%%%%%%%%%%%%%%%%%%%%%%%%%%%%%%%%%%%%%%%%%%%%%%%%%%%%%%%%%%%%%%%%%%
%% Menno van Zaanen                                                           %%
%% menno@comp.leeds.ac.uk                                                     %%
%%%%%%%%%%%%%%%%%%%%%%%%%%%%%%%%%%%%%%%%%%%%%%%%%%%%%%%%%%%%%%%%%%%%%%%%%%%%%%%%
%% Filename: results.tex                                                      %%
%%%%%%%%%%%%%%%%%%%%%%%%%%%%%%%%%%%%%%%%%%%%%%%%%%%%%%%%%%%%%%%%%%%%%%%%%%%%%%%%

%%%%%%%%%%%%%%%%%%%%%%%%%%%%%%%%%%%%%%%%%%%%%%%%%%%%%%%%%%%%%%%%%%%%%%%%%%%%%%%%
%%%%%%%%%%%%%%%%%%%%%%%%%%%%%%%%%%%%%%%%%%%%%%%%%%%%%%%%%%%%%%%%%%%%%%%%%%%%%%%%
%%%%%%%%%%%%%%%%%%%%%%%%%%%%%%%%%%%%%%%%%%%%%%%%%%%%%%%%%%%%%%%%%%%%%%%%%%%%%%%%
\thesischapter{Empirical Results}
{After all the purpose of computing is insight, not numbers,\\
and one good way to demonstrate this is to take a situation\\
where the numbers are clearly less important than the insights gained.}
{\citet{Knuth:SPCS-96-??}}
%%%%%%%%%%%%%%%%%%%%%%%%%%%%%%%%%%%%%%%%%%%%%%%%%%%%%%%%%%%%%%%%%%%%%%%%%%%%%%%%
%%%%%%%%%%%%%%%%%%%%%%%%%%%%%%%%%%%%%%%%%%%%%%%%%%%%%%%%%%%%%%%%%%%%%%%%%%%%%%%%
%%%%%%%%%%%%%%%%%%%%%%%%%%%%%%%%%%%%%%%%%%%%%%%%%%%%%%%%%%%%%%%%%%%%%%%%%%%%%%%%
\label{ch:results}
\index{result}%
\index{Knuth, D.E.}%

This chapter will put the theory of the previous chapters into
practice. Each of the \abl systems (which are combinations of an
alignment and a selection learning instance) is evaluated.
Firstly, the focus is on evaluating the alignment learning phase,
followed by a comparison of the results of both phases. Using the
data generated by these systems, the grammar extraction and
parsing phase of the \parseabl systems are also evaluated.

Apart from a numerical analysis, which will be described first,
the learned treebanks generated by \abl systems are looked at in
a qualitative way.  The learned treebanks contain for example
constituents that closely resemble nouns and ``from-to'' phrases,
but also many words are roughly tagged accoring to
parts-of-speech.  Furthermore, each of the generated treebanks
contain recursive structures.

%%%%%%%%%%%%%%%%%%%%%%%%%%%%%%%%%%%%%%%%%%%%%%%%%%%%%%%%%%%%%%%%%%%%%%%%%%%%%%%%
%%%%%%%%%%%%%%%%%%%%%%%%%%%%%%%%%%%%%%%%%%%%%%%%%%%%%%%%%%%%%%%%%%%%%%%%%%%%%%%%
\thesissection{Quantitative results}
%%%%%%%%%%%%%%%%%%%%%%%%%%%%%%%%%%%%%%%%%%%%%%%%%%%%%%%%%%%%%%%%%%%%%%%%%%%%%%%%
%%%%%%%%%%%%%%%%%%%%%%%%%%%%%%%%%%%%%%%%%%%%%%%%%%%%%%%%%%%%%%%%%%%%%%%%%%%%%%%%
\index{result!quantitative}%

This section first discusses the advantages and disadvantages of
the method used here to evaluate the \abl and \parseabl
instances. Next, the test environment, consisting of the
treebanks, metrics and learning instances used, is described.
\index{treebank}%
\index{structured corpus}%
Finally, the actual results of the different phases of the
framework and their evaluation are given.

%%%%%%%%%%%%%%%%%%%%%%%%%%%%%%%%%%%%%%%%%%%%%%%%%%%%%%%%%%%%%%%%%%%%%%%%%%%%%%%%
\thesissubsection{Different evaluation approaches}
%%%%%%%%%%%%%%%%%%%%%%%%%%%%%%%%%%%%%%%%%%%%%%%%%%%%%%%%%%%%%%%%%%%%%%%%%%%%%%%%
\label{s:approach}

Evaluating language learning systems is difficult.  Usually, one
of three different evaluation methods is chosen.  Here, the three
methods will be described briefly concentrating on their advantages
and disadvantages.
\begin{description}

\item [Looks-good-to-me approach]
\index{evaluation!looks-good-to-me approach}%
When a language learning system is evaluated using the
looks-good-to-me approach, the system is applied to an
unstructured piece of text and the resulting grammar rules or
structured sentences are qualitatively evaluated. If
(intuitively) correct grammatical structures are found in the
grammar or structured sentences, the system is said to be good.
Since this is such a simple approach, it has been used to
evaluate many systems, for example those by
\citet{Cook:76-10-59,Cook:94-1-231,%
Finch:SHOE92-230,Grunwald:CSS-94-203,Huckle:ACL95-??,%
Losee:95-??-??,Scholtes:IJCNN92-??,Stolcke:ICGI94-106,%
Vervoort:GWG-00}.
\index{Cook, C.M.}%
\index{Rosenfeld, A.}%
\index{Aronson, A.R.}%
\index{Cook, D.J.}%
\index{Holder, L.B.}%
\index{Finch, S.}%
\index{Chater, N.}%
\index{Gr\"unwald, P.}%
\index{Huckle, C.C.}%
\index{Losee, R.M.}%
\index{Scholtes, J.C.}%
\index{Bloembergen, S.}%
\index{Stolcke, A.}%
\index{Omohundro, S.}%
\index{Vervoort, M.R.}%

\begin{description}
\item [advantages]
The main advantage of this approach is that only unstructured
data is needed. This means that the system can easily be
evaluated on different languages without the need of
structured corpora.

Another advantage is that the evaluation can focus on certain
specific syntactic constructions the system should be able to
learn.  Since the evaluation is done in a qualitative way, the
input data can be specialised to show that the system can learn
such constructions and the structured output can be searched
for the wanted syntax.

\item [disadvantages]
It may be clear that this approach has a few disadvantages.
First of all, for the evaluation of language learning sytems
using this approach, an expert who has specific knowledge of
the syntax of the language in the test corpus is needed.  The
expert can tell which syntactic structures should be present in
the learned treebank and also if certain generated structures are
correct or incorrect.

This leads to the second disadvantage, that of biased evaluation.
In this approach, it is possible to pick the correctly learned
grammatical features only and leave out the incorrect ones. It
will seem that the system works well, since all grammatical
structures that are shown are correct. However, the learned
treebank also contains incorrect structures which are not shown.

Furthermore, it is difficult to compare two systems based on this
approach. Imagine two language learning systems that each find some
other correct and incorrect structures. Which of the two systems
is better?

The final disadvantage is that the unstructured input corpus may
be biased towards a specific system. In other words, it may be
possible to (unknowingly) feed the system only sentences that
will generate the wanted syntactic structures.
\end{description}

\item [Rebuilding known grammars]
\index{evaluation!rebuilding known grammar}%
Another approach in evaluating language learning systems is to
let the system rebuild a known grammar. This evaluation method
starts out with a (simple) grammar, from which a set of
example sentences is generated. This set of sentences must
at least represent each of the features of the grammar once. The
sentences are then fed to the learning system and the resulting
grammar is compared manually to the original grammar. If the
learned grammar is similar or equal to the original grammar then
the learning system is considered good.  This evaluation approach
has been used by, a.o.\
\citet{Cook:76-10-59,Nakamura:ICGI00-186,Pereira:ACL92-128,%
Sakakibara:ICGI00-229,Stolcke:BLP-94,Wolff:CLP-96-67}.
\index{Cook, C.M.}%
\index{Rosenfeld, A.}%
\index{Aronson, A.R.}%
\index{Nakamura, K.}%
\index{Ishiwata, T.}%
\index{Pereira, F.}%
\index{Schabes, Y.}%
\index{Sakakibara, Y.}%
\index{Muramatsu, H.}%
\index{Stolcke, A.}%
\index{Wolff, J.G.}%

\begin{description}
\item [advantages]
This method has similar advantages as the looks-good-to-me
approach. It does not need structured sentences to evaluate,
because plain sentences, generated by the grammar, are used to
learn structure.  These grammars can be tailored to specific
grammatical constructions, which allows for a specific evaluation
of certain aspects of the system.

Additionally, the looks-good-to-me approach needs an expert to
indicate whether a grammatical construction is correct or
incorrect, but this is unnecessary for this approach. If the
learned grammar is similar to the original grammar, the learning
system works well.

Another advantage is that this method of evaluation yields a more
objective way of comparing different language learning systems,
since the entire grammars are compared (and not only the
intuitively correct parts of the grammar). However, even then it
is still difficult to say which of two (slightly) incorrect
grammars is closer to the original grammar.

\item [disadvantages]
One of the disadvantages of this approach is that the evaluation
of the system depends heavily on the chosen grammar.  Some
learning systems can rebuild certain types of grammars more
easily than others; by choosing simple grammars, the system can
be shown to be better than it really is.

The idea of a language learning system is that it finds correct
structure in real natural language data. This means that the
original grammar should be as close to the underlying grammar of
natural language sentences as possible.  However, this grammar is
in general not fully known.

A related problem is that the language generation model, which
generates the example sentences from the grammar, is not known
either. The language generation model should create a set of
sentences that describes each of the grammatical features
contained in the grammar, but if the language learning system is
designed to work on real natural language texts (as is \abl), it
should also create sentences that resemble real natural language
sentences.
\end{description}

\item [Compare against a treebank]
\index{evaluation!compare against treebank}%
The third method of evaluation a language learning system, which
will be used in this thesis, is to apply the system to plain
natural language sentences which are extracted from a treebank.
The structured sentences generated by the language learning
system are then compared against the original structured
sentences from the treebank.

There are several metrics that can be used to compare the learned
tree against the original tree structure. Most often, the
\emph{recall}, which gives a measure of the completeness of the
learned grammar, and the \emph{precision}, which shows how
correct the learned structure is, are computed. These metrics
give a gradual indication of the performance of a learning system
which allows the comparison of two systems that find correct, but
also some incorrect, structure.

This method of evaluating learning systems is, in our view, the
most objective. Lately, almost all systems are evaluated using
this approach. Examples can be found
in~\citep{Brent:99-34-71,Brill:ACL93-259,Clark:ACL/CNLL01-105,%
Dejean:CONLL00-95,Nevado:ICGI00-196}.
\index{Brent, M.R.}%
\index{Brill, E.}%
\index{Clark, A.}%
\index{D\'ejean, H.}%
\index{Nevado, F.}%
\index{S\'anchez, J-A.}%
\index{Bened\'\i, J-M.}%

\begin{description}
\item [advantages]
This approach does not need an expert to indicate if some
construction is correct or incorrect. All linguistic knowledge is
already contained in the original treebank. This also means that
real natural language data can be used. Of the three evaluation
methods, this approach comes closest to the evaluation of the
system in the context of a real world application.

Furthermore, this approach allows for a relatively objective
comparison of learning systems. Since several metrics can be
used, gradual distinctions between systems can be made.
Precise information indicating how well the systems perform on
the treebank can be found, since the systems are given a score
even when only parts of the learned structure are correct.

\item [disadvantages]
This method also has its disadvantages. The first disadvantage is
that a collection of structured sentences is needed. At the
moment however, structured data is only available for a limited
number of languages.  This severely restricts the evaluation of
language independency of the systems.

A second disadvantage is that the annotation scheme and imperfect
annotation influence the results.  If, for example, the structure
of the sentences in the original treebank is relatively flat, but
the language learning system finds rich, deep structures, it will
be penalised for that, since it learns structure that is not
present in the ``correct'' treebank. Suppose that for example
sentence~\ref{flatdeep:a} is the structured sentences in a
treebank and sentence~\ref{flatdeep:b} is the learned tree, the
constituent [\sent{Big}]\snt{AP} will be counted as incorrect,
since it is not present in the original treebank.  However,
depending on the annotation scheme, this constituent could have
been correct.  Similar things happen when the annotation of the
sentences in the original treebank contains errors.
\end{description}
\end{description}

\eenumsentence{
\item \empty[[Oscar]\snt{NP} [sees [Big Bird]\snt{NP}]\snt{VP}]\snt{S}
\label{flatdeep:a}
\item \empty[[Oscar]\snt{NP} [sees [[Big]\snt{AP}
Bird]\snt{NP}]\snt{VP}]\snt{S}
\label{flatdeep:b}
}

Evaluating language learning systems by comparing them against a
treebank is done as shown in figure~\ref{fig:evaluating}.
Evaluation starts off with a base treebank.  The trees in this
treebank are taken to be perfectly correct, a gold standard.
From each tree in this treebank, the plain sentence (i.e.\ the
yield of the tree) is extracted. These sentences (contained in
the plain corpus in the figure) are fed to the language learning
system and the results are stored in the learned treebank. The
trees in the learned treebank are then compared to the trees from
the base treebank.

\begin{fig}{Evaluating a structure induction system}
\label{fig:evaluating}
\vspace{1em}
\psset{framearc=.2,arrowscale=2}
%\begin{psmatrix}[rowsep=.8cm,colsep=1.2cm]
\begin{psmatrix}[colsep=25pt]
&&[name=base]\psframebox[linestyle=dashed]{\parbox{2cm}{Base\\treebank}}\\
\empty[name=extract]\psframebox{\parbox{2cm}{Extract\\sentences}}&
&&[name=eval]\psframebox{\parbox{2cm}{Compare\\treebanks}}&
[name=res]\psframebox[doubleline=true]{\parbox{2cm}{Results\\}}\\
\empty[name=plain]\psframebox[linestyle=dashed]{\parbox{2cm}{Plain\\corpus}}&
[name=abl]\psframebox{\parbox{2cm}{Learning\\system}}&
[name=learned]\psframebox[linestyle=dashed]{\parbox{2cm}{Learned\\treebank}}
\psset{arrows=->}
\ncangle[angleA=180,angleB=90]{base}{extract}
\ncangle[angleB=90]{base}{eval}
\ncline{extract}{plain}
\ncline{plain}{abl}
\ncline{abl}{learned}
\ncangle[angleB=270]{learned}{eval}
\ncangle[angleB=180]{eval}{res}
\end{psmatrix}
\end{fig}

Applying the evaluation method to a language learning system
indicates how much of the structure in the original treebank is
found and how much correct structure is generated by the system.
To get an idea how good the system really is, it needs to be
compared against another system.  Comparing the results of two
language learning systems against each other shows which of the
two works better (on this specific corpus).  Normally, a system
is compared against a baseline system. This is usually a simple
system that generates for example trees with random structure.

%%%%%%%%%%%%%%%%%%%%%%%%%%%%%%%%%%%%%%%%%%%%%%%%%%%%%%%%%%%%%%%%%%%%%%%%%%%%%%%%
\thesissubsection{Test environment}
%%%%%%%%%%%%%%%%%%%%%%%%%%%%%%%%%%%%%%%%%%%%%%%%%%%%%%%%%%%%%%%%%%%%%%%%%%%%%%%%
\index{test environment}%

This section describes the settings used for testing the \abl and
\parseabl frameworks. Starting with a description of the
different treebanks that will be used in this thesis, the section
continues by explaining the evaluation metrics that are computed
to compare the learned tree structures against the original tree
structures.  Finally, a brief overview of the tested systems will
be given.

%%%%%%%%%%%%%%%%%%%%%%%%%%%%%%%%%%%%%%%%%%%%%%%%%%%%%%%%%%%%%%%%%%%%%%%%%%%%%%%%
\thesissubsubsection{Treebanks}
%%%%%%%%%%%%%%%%%%%%%%%%%%%%%%%%%%%%%%%%%%%%%%%%%%%%%%%%%%%%%%%%%%%%%%%%%%%%%%%%
\label{s:treebanks}
\index{treebank}%

As mentioned in the previous section, the grammar induction
systems will be applied to the plain sentences of a structured
corpus. This section will describe the treebanks that have been
used to test the \abl and \parseabl systems in this thesis.

First of all, it needs to be mentioned that not many structured
treebanks are available. There is only a limited number of
languages for which structured treebanks are available.
\index{treebank!non-English}%
Unfortunately, to test if a grammar induction system is really
language independent, it needs to be applied to treebanks of
several (different) languages.

Here, two treebanks have been used to extensively test the \abl
and \parseabl systems. Results on a third treebank, the Wall
Street Journal treebank, will be discussed in a separate section.
\index{Wall Street Journal}%
\index{WSJ}%
\index{treebank!Wall Street Journal}%
The first treebank is the Air Traffic Information System (ATIS)
\index{Air Traffic Information System}%
\index{ATIS}%
\index{treebank!Air Traffic Information System}%
treebank \citep{Marcus:93-19-313}, which is taken from the Penn
\index{Marcus, M.}%
\index{Santorini, B.}%
\index{Marcinkiewicz, M.}%
treebank 2. It is an English treebank containing mostly questions
and imperatives on air traffic. The sentences in~\ref{exatis} are
some (random) samples that are found in the treebank. The
treebank consists of 577 sentences with a mean sentence length of
7.5 words per sentence.  All empty constituents (and
traces) that can be found in this treebank have been removed
beforehand.

\eenumsentence{
\toplabel{exatis}
\item What airline is this
\item The return flight should leave at around seven pm
\item Show me the flights from Baltimore to Oakland please
}

The second treebank is the Openbaar Vervoer Informatie
Systeem\footnote{Openbaar Vervoer Informatie Systeem translates
to Public Transport Information System.} (OVIS) treebank
\index{Openbaar Vervoer Informatie Systeem}%
\index{OVIS}%
\index{treebank!Openbaar Vervoer Informatie Systeem}%
\citep{Bonnema:ACL97-159}. This Dutch treebank, with its 10,000
\index{Bonnema, R.}%
\index{Bod, R.}%
\index{Scha, R.}%
trees, is larger than the ATIS corpus. The sentences
in~\ref{exovis} are example sentences taken from this corpus. The
mean sentence length in this treebank, which is 3.5 words per
sentence, is much shorter than in the ATIS corpus. Apart from a
few (simple) questions, the sentences in this corpus are all
imperatives or answers to questions.

\eenumsentence{
\toplabel{exovis}
\item van bergambacht naar lisse
\item naar vlissingen zei ik toch
\item ja dat heb ik nou de hele tijd gezegd ik wil niet naar
alkmaar
}

The third treebank, section~23 of the Wall Street Journal (WSJ)
treebank, will be discussed in section~\ref{s:wsj}.  Where the
ATIS and OVIS can be seen as development corpora, the WSJ
treebank is chosen to show that the system also works on a new
and more complex set of sentences.  Additionally, this is the
first time that an unsupervised language learning system is
applied to the plain sentences of this corpus.

%%%%%%%%%%%%%%%%%%%%%%%%%%%%%%%%%%%%%%%%%%%%%%%%%%%%%%%%%%%%%%%%%%%%%%%%%%%%%%%%
\thesissubsubsection{Metrics}
%%%%%%%%%%%%%%%%%%%%%%%%%%%%%%%%%%%%%%%%%%%%%%%%%%%%%%%%%%%%%%%%%%%%%%%%%%%%%%%%
\label{s:metrics}
\index{metric}%

To compare the trees in the learned treebank against the trees in
the original treebank, metrics have to be defined. The metrics
indicate how similar tree structures are, and thus can be used to
show how well the learning systems perform. This section will
describe the metrics that are used in this thesis.

The results have been computed with the commonly used
EVALB\footnote{Available at
\texttt{http://www.cs.nyu.edu/cs/projects/proteus/evalb/}.}
\index{EVALB}%
\index{evaluation}%
program \citep{Collins:ACL97-16}. The only difference from the
\index{Collins, M.}%
parameter file that is supplied with the program is that
\emph{unlabelled} metrics, which do not take into account the
\index{metric!unlabelled}%
type labels of the constituents, are used instead of their
\emph{labelled} counterparts.
\index{metric!labelled}%

The metrics that are used in this thesis are described here
briefly.\footnote{The EVALB program additionally computes other
metrics, but these metrics do not yield much more insight into
the systems described here; the changes in recall and precision
directly match the changes in the other metrics.} To be able to
describe the metrics formally, some notions have to be introduced
first.  \textsl{Sentences} is the list of sentences (without
structure) that are contained in the structured corpus. The
function \textsl{gold(s)} returns the tree in the original
treebank, or the ``gold standard'', that belongs to sentence
\textsl{s}. The function \textsl{learned(s)} is similar to
\textsl{gold(s)}, however, it returns the tree in the learned
treebank. The function \textsl{correct(t, u)} returns the set of
constituents that can be found in both trees \textsl{t} and
\textsl{u}. It finds the constituents that have the same beginning
and end in both trees (i.e.\ it finds all constituents that have
an equivalent constituent in the other tree structure).  Note
that since the metrics are non-labelled, two constituents are
considered equal if their begin and end indices are equal; their
non-terminal type may be different.

\newcommand{\msl}[1]{\mbox{\textsl{#1}}}
\newcommand{\msls}[1]{\mbox{\scriptsize{\textsl{#1}}}}
\begin{description}
\item [(Bracketing) Recall] This metric shows how many of the
\index{recall}%
\index{metric!recall}%
correct constituents have been learned, which gives an idea about
the completeness of the learned grammar. It is the percentage of
correctly learned constituents that are also found in the
original treebank.
\[
\mbox{Recall}=\frac{\sum_{\msls{s}\in\msls{sentences}}
|\msl{correct(gold(s), learned(s))}|}
{\sum_{\msls{s}\in\msls{sentences}}|\msl{gold(s)}|}
\]

\item [(Bracketing) Precision] The precision metric indicates how
\index{precision}%
\index{metric!precision}%
many learned constituents are also correct. It describes the
correctness of the learned grammar. This metric returns the
percentage of correctly learned constituents with respect to all
learned constituents.
\[
\mbox{Precision}=\frac{\sum_{\msls{s}\in\msls{sentences}}
|\msl{correct(gold(s), learned(s))}|}
{\sum_{\msls{s}\in\msls{sentences}}|\msl{learned(s)}|}
\]

\item [F-score] The f-score (which is not computed by the EVALB
\index{f-score}%
\index{metric!f-score}%
program) combines the recall and precision measures into one
score. Increasing the $\beta$ value makes the precision metric
more important. Here, it is assumed that recall and precision are
equally important, so $\beta$ is set to 1.
\[
\mbox{F}_\beta=\frac{(\beta^2+1)*\mbox{Precision}*\mbox{Recall}}
{(\beta^2*\mbox{Precision})+\mbox{Recall}}
\]
\end{description}

Of course, one can think of many other metrics. In the rest of
this chapter some simple, additional metrics are used, for
example, to indicate how many constituents are learned, or what
the mean sentence length is. These metrics are considered
self-explaining.

As discussed in section~\ref{s:approach}, evaluating language
learning system has some disadvantages. The main problem here is
that the metrics rely heavily on the annotation scheme used.
Constituents that are correct when compared to a tree annotated
using one scheme might be incorrect if the tree had been
annotated using another scheme. However, since the systems are
all extensively tested on two different corpora (with different
annotation schemes), the combination of metrics that have just
\index{annotation scheme}%
been described will hopefully allow us to compare the different
language learning methods. Furthermore, the metrics are only used
to \emph{compare} systems; the evaluation only depends on the
relative values of the metrics. The absolute values of the
metrics are not important in this case.

One important final note is that the metrics described above were
originally designed to measure and compare supervised systems
\index{evaluation!supervised system}%
(especially parsers).  In this thesis, however, the metrics will
be used to compare unsupervised systems. This means that the
actual values for each metric cannot under any circumstance be
compared against the values of supervised systems. It may be
expected that the unsupervised systems evaluated in this thesis
yield much lower values for each of the metrics in contrast to
supervised systems, since the unsupervised systems have no prior
knowledge of the structure present in the treebank it is compared
against.

\nocite{Sampson:00-5-53}

%%%%%%%%%%%%%%%%%%%%%%%%%%%%%%%%%%%%%%%%%%%%%%%%%%%%%%%%%%%%%%%%%%%%%%%%%%%%%%%%
\thesissubsubsection{Tested systems}
%%%%%%%%%%%%%%%%%%%%%%%%%%%%%%%%%%%%%%%%%%%%%%%%%%%%%%%%%%%%%%%%%%%%%%%%%%%%%%%%

The \abl framework consists of two distinct phases and both
phases have several instances. Selecting an instance for both
\index{instance!tested}%
phases yields a specific system.  Since three alignment learning
and three selection learning instances are discussed, this
results in nine different systems. The three alignment learning
instances are evaluated separately in section~\ref{s:aligneval}
and each of the combined systems is evaluated in
section~\ref{s:selecteval}.

When considering the \parseabl framework, a grammar is extracted
from the output of one of the different \abl systems. Systems
using both types of grammar (\scfg and \stsg) are tested.  For
the evaluation of the \stsg framework, two instances with the
maximum depth of subtrees set to two or three are tested.  The
evaluation of these results will be discussed in
section~\ref{s:parseeval}.

Since there are a lot of possible combinations of instances, a
simple naming scheme is introduced here. All instances have their
own name, so a system is named by the combination of names of its
instances. For example, \ablone:\termsl is an \abl system that
uses the \ablone alignment learning instance and the \termsl
selection learning phase. \parseabl systems have an extra phase,
so these names will be similar to \ablall:\first:\scfg. Here the
result of the \ablall alignment learning instance combined with
the \first selection learning phase is used to extract an \scfg.
An overview of all possible combinations is depicted in
figure~\ref{fig:testedsystems}.

\begin{fig}{Tested \abl and \parseabl systems}
\label{fig:testedsystems}
\psset{framearc=.2,arrowscale=2}
\begin{psmatrix}[rowsep=.5cm]
Alignment learning &
  Selection learning &
    Grammar extraction \\\empty
[name=ao]\psframebox{\parbox{2cm}{\ablone}} &
  [name=sf]\psframebox{\parbox{2cm}{\first}} \\
&
  &
    [name=ec]\psframebox{\parbox{2cm}{\scfg}} \\\empty
[name=ad]\psframebox{\parbox{2cm}{\abldis}} &
  [name=st]\psframebox{\parbox{2cm}{\termsl}} \\
& 
  &
    [name=et]\psframebox{\parbox{2cm}{\stsg}} \\\empty
[name=aa]\psframebox{\parbox{2cm}{\ablall}} &
  [name=sc]\psframebox{\parbox{2cm}{\constl}}
\end{psmatrix}
\psset{arrows=->}
\ncline{ao}{sf}\ncline{ao}{st}\ncline{ao}{sc}
\ncline{ad}{sf}\ncline{ad}{st}\ncline{ad}{sc}
\ncline{aa}{sf}\ncline{aa}{st}\ncline{aa}{sc}
\ncline{sf}{ec}\ncline{st}{ec}\ncline{sc}{ec}
\ncline{sf}{et}\ncline{st}{et}\ncline{sc}{et}
\end{fig}

To be able to compare the results, a baseline system called
\emph{random}, is applied to the three corpora in addition to the
\abl systems. Like the \abl systems, the resulting treebank is
compared against the original treebank and the values for the
three metrics are computed. The baseline system randomly chooses
\index{baseline system}%
\index{random system}%
for each sentence in the corpus a left or right branching
structure (as displayed in figure~\ref{fig:leftright}).  This
system was chosen as a baseline, since it is a simple, language
independent system (like the \abl systems). A right branching
system (which only assigns right branching structures to
\index{right branching system}%
\index{left branching system}%
sentences) would perform better on for example an English or
\index{English}%
Dutch corpus, but it would not perform as well on a corpus of a
\index{Dutch}%
left branching language (like Japanese) and hence it is language
\index{Japanese}%
dependent. The random system is expected to be much more robust
and does not assume anything about the sentences it is assigning
structure to.

\begin{fig}{Left and right branching trees}
\label{fig:leftright}
\begin{multicols}{2}
\psset{nodesep=2pt,levelsep=35pt}
Left branching tree\\
{}[[[[Oscar]\snt{4} sees]\snt{3} Big]\snt{2} Bird]\snt{1}\\
\pstree{\TR{\nt{1}}}
{\pstree{\TR{\nt{2}}}{\pstree{\TR{\nt{3}}}{\pstree{\TR{\nt{4}}}{\TR{Oscar}}
                                           \TR{sees}
                                          }
                      \TR{Big}
                     }
 \TR{Bird}
}

Right branching tree\\
{}[Oscar [sees [Big [Bird]\snt{4}]\snt{3}]\snt{2}]\snt{1}\\
\pstree{\TR{\nt{1}}}
{\TR{Oscar}
 \pstree{\TR{\nt{2}}}{\TR{sees}
                      \pstree{\TR{\nt{3}}}{\TR{Big}
                                           \pstree{\TR{\nt{4}}}{\TR{Bird}}
                                          }
                     }
}
\end{multicols}
\end{fig}

Since each of the alignment learning instances (apart from the
\ablall instance) depends on the order of the sentences in the
\index{sentence!order of}%
plain corpus, all systems have been applied to the plain corpus
ten times. The results that will be shown in the rest of the
thesis are the mean values of the metrics, followed by the
\index{mean}%
standard deviation between brackets.
\index{standard deviation}%

%%%%%%%%%%%%%%%%%%%%%%%%%%%%%%%%%%%%%%%%%%%%%%%%%%%%%%%%%%%%%%%%%%%%%%%%%%%%%%%%
\thesissubsection{Test results and evaluation}
%%%%%%%%%%%%%%%%%%%%%%%%%%%%%%%%%%%%%%%%%%%%%%%%%%%%%%%%%%%%%%%%%%%%%%%%%%%%%%%%

This section will give the numerical results of the baseline,
\index{result!numerical}%
\abl and \parseabl systems when applied to the ATIS, OVIS and WSJ
corpora.\footnote{These results differ from results in previous
publications. There are several reasons for this.  First,
slightly different corpora and metrics are used. Secondly, a new
implementation has been used here, which finds hypotheses in a
different way from the previous implementation (as described in
this thesis) and some minor implementation errors have been
corrected.} First, the alignment learning phase will be
evaluated, followed by the combinations of alignment learning and
selection learning.  Finally, after the evaluation on the WSJ
corpus, the \parseabl framework will be tested.

%%%%%%%%%%%%%%%%%%%%%%%%%%%%%%%%%%%%%%%%%%%%%%%%%%%%%%%%%%%%%%%%%%%%%%%%%%%%%%%%
\thesissubsubsection{Alignment learning systems}
%%%%%%%%%%%%%%%%%%%%%%%%%%%%%%%%%%%%%%%%%%%%%%%%%%%%%%%%%%%%%%%%%%%%%%%%%%%%%%%%
\label{s:aligneval}

The alignment learning inserts all hypotheses that will be
present in the final treebank (after selection learning).  The
selection learning phase only removes hypotheses.  This means
that if constituents in the gold standard cannot be found in the
learned hypothesis space after alignment learning, they will not
be present in the final treebank.  The alignment learning phase
works best when it inserts as many correct hypotheses and as few
incorrect hypotheses as possible.

To evaluate the alignment learning phase separately from the
\index{evaluation!alignment learning}%
selection learning phase, the three alignment learning systems
have been applied to the ATIS and OVIS corpora. This results in
ambiguous hypothesis spaces. These hypothesis spaces cannot be
compared to the gold standard directly, since they contain fuzzy
trees instead of proper tree structures.

The hypothesis spaces are evaluated by assuming the perfect
selection learning system. The hypothesis space is disambiguated
by selecting only those hypotheses that are also present in the
gold standard. This will give the upper bound on all metrics. The
\index{evaluation!upper bound}%
\index{alignment learning!upper bound}%
real selection learning methods (evaluated in the next section)
can never improve on these values.

The results of applying the alignment learning phases to the ATIS
and OVIS corpora and selecting only the hypotheses that are
present in the original treebanks can be found in
table~\ref{tab:align}.  The precision in this table is 100\%,
since only correct constituents are selected from the hypothesis
space. The recall indicates how many of the correct constituents
can be found in the learned treebank.

\newcommand{\mc}[1]{\multicolumn{2}{c|}{#1}}
\newcommand{\md}[1]{\multicolumn{2}{c||}{#1}}
\begin{tab}{Results alignment learning on the ATIS and OVIS corpus}
\label{tab:align}
\vspace{1em}
\begin{tabular}{|l|l||*{3}{rr|}}
\hline
&        & \mc{Recall}    & \mc{Precision} & \mc{F-score} \\
\hline\hline
ATIS
& random  & 28.90 & (0.58) & 100.00 & (0.00) & 44.83 & (0.70) \\
& \ablone & 48.08 & (0.09) & 100.00 & (0.00) & 64.94 & (0.08) \\
& \abldis & 19.52 & (2.67) & 100.00 & (0.00) & 32.60 & (3.64) \\
& \ablall & 50.11 & (0.00) & 100.00 & (0.00) & 66.76 & (0.00) \\
\hline\hline
OVIS
& random  & 52.73 & (0.09) & 100.00 & (0.00) & 69.05 & (0.40) \\
& \ablone & 94.22 & (0.04) & 100.00 & (0.00) & 97.02 & (0.02) \\
& \abldis & 53.65 & (2.27) & 100.00 & (0.00) & 69.81 & (1.93) \\
& \ablall & 96.47 & (0.00) & 100.00 & (0.00) & 97.68 & (0.00) \\
\hline
\end{tabular}
\end{tab}

Table~\ref{tab:nhalign} gives an overview of the number of
hypotheses contained in the hypothesis spaces generated by the
alignment learning phases. To get an idea of how these amount
compare to the original treebank, the gold standard ATIS and OVIS
treebanks contains respectively 7,197 and 57,661 constituents.
Additionally, it shows how many hypotheses were removed from the
hypothesis spaces to build the upper bound treebanks. The number
of constituents present in those treebanks are also given.

\begin{tab}{Number of hypotheses after alignment learning}
\label{tab:nhalign}
\vspace{1em}
\begin{tabular}{|l|l||*{3}{rr|}}
\hline
&         & \mc{Learned}    & \mc{Best}      & \mc{Removed} \\
\hline\hline
ATIS
& random  &  4,353 &   (0.0) & 1,851 &  (25.6) & 2,502 &  (25.6) \\
& \ablone & 12,692 &   (8.8) & 4,457 &   (4.4) & 8,235 &   (9.8) \\
& \abldis &  2,189 & (796.8) & 1,175 & (331.2) & 1,013 & (460.3) \\
& \ablall & 14,048 &   (0.0) & 4,619 &   (0.0) & 9,429 &   (0.0) \\
\hline\hline
OVIS
& random  &  34,221 &    (0.0) & 22,301 &  (108.1) & 11,920 &  (108.1) \\
& \ablone & 123,699 &   (62.6) & 50,365 &   (28.7) & 73,334 &   (44.0) \\
& \abldis &  40,399 & (1,506.6) & 21,488 & (1,049.8) & 18,911 & (1,250.7) \\
& \ablall & 129,646 &   (0.00) & 51,158 &   (0.00) & 78,488 &   (0.00) \\
\hline
\end{tabular}
\end{tab}

It is interesting to see that the \abldis system does not perform
very well at all. The recall is low (even compared to the
baseline) and the results vary widely as indicated by the
standard deviation. This is the case on both the ATIS and OVIS
corpora, although the system performs slightly better on the
latter. Table~\ref{tab:nhalign} shows us why this is the case.
The \abldis alignment learning method, like the random base line
system, does not introduce many hypotheses.  It even inserts less
hypotheses than there are constituents in the gold treebanks.

The \ablone and \ablall systems might seem to yield roughly
similar results, however, the \ablall system is significantly
better. Both systems insert almost the same amount of hypotheses
in their hypothesis spaces, but since the \ablall system
processes all possible alignments, it finds more (and thus more
correct) hypotheses than the other alignment learning instances.

Note that the \ablall system does not depend on the order of the
sentences in the corpus (hence the zero standard deviation),
since all possible alignments are computed. The other systems do
depend on the order of the sentences. Especially the \abldis
system introduces a largely varying number of hypotheses.

Since the \ablone and \ablall systems add more hypotheses, the
hypothesis spaces will contain more correct hypotheses (as
indicated by the higher recall for both systems), but the
selection learning phase also has a harder task, since it has
more hypotheses to choose from. This phase will be investigated
next.

%%%%%%%%%%%%%%%%%%%%%%%%%%%%%%%%%%%%%%%%%%%%%%%%%%%%%%%%%%%%%%%%%%%%%%%%%%%%%%%%
\thesissubsubsection{Selection learning systems}
%%%%%%%%%%%%%%%%%%%%%%%%%%%%%%%%%%%%%%%%%%%%%%%%%%%%%%%%%%%%%%%%%%%%%%%%%%%%%%%%
\label{s:selecteval}

For the evaluation of the complete \abl systems, all instances
have been applied to the two corpora. The recall, precision and
f-scores of the \abl systems and the baseline can be found in
\index{selection learning!evaluation}%
\index{evaluation!selection learning}%
table~\ref{tab:selectatis} for the ATIS corpus and in
table~\ref{tab:selectovis} for the OVIS corpus.

Remember that the selection learning phase works best when it
removes as many incorrect and as few correct hypotheses from the
hypothesis space. If there are many correct and few incorrect
hypotheses in the hypothesis universe, then the selection
learning phase has an easy task selecting the correct hypotheses.
From this, it can be expected that the results of the selection
learning phases on the hypothesis space generated by the \abldis
system will be close to the upper bound, whereas the selection
learning on the hypothesis space of the \ablall system will
perform less than perfect.

\begin{tab}{Results selection learning on the ATIS corpus}
\label{tab:selectatis}
\vspace{1em}
\begin{tabular}{|l|l||*{3}{rr|}}
\hline
&       & \mc{Recall}    & \mc{Precision} & \mc{F-score}   \\
\hline\hline
random
&       & 28.90 & (0.58) & 33.73 & (0.68) & 31.13 & (0.63) \\
\hline
\ablone
& upper & 48.08 & (0.09) & 100.00 & (0.00) & 64.94 & (0.08) \\
&\first & 31.64 & (0.94) &  38.94 & (1.32) & 34.91 & (1.10) \\
&\termsl& 25.82 & (0.19) &  54.73 & (0.42) & 35.09 & (0.25) \\
&\constl& 20.81 & (0.20) &  46.57 & (0.39) & 28.76 & (0.26) \\
\hline
\abldis
& upper & 19.52 & (2.67) & 100.00 & (0.00) & 32.60 & (3.64) \\
&\first & 18.20 & (2.06) &  55.32 & (4.82) & 27.21 & (1.59) \\
&\termsl& 18.01 & (1.39) &  56.56 & (3.97) & 27.23 & (1.13) \\
&\constl& 17.82 & (1.24) &  56.62 & (3.67) & 27.02 & (1.03) \\
\hline
\ablall
& upper & 50.11 & (0.00) & 100.00 & (0.00) & 66.76 & (0.00) \\
&\first & 32.42 & (1.02) &  39.34 & (1.34) & 35.54 & (1.16) \\
&\termsl& 25.19 & (0.11) &  53.31 & (0.24) & 34.21 & (0.15) \\
&\constl& 20.68 & (0.02) &  45.25 & (0.05) & 28.39 & (0.03) \\
\hline
\end{tabular}
\end{tab}

Table~\ref{tab:selectatis} and~\ref{tab:selectovis} give the
results of applying the alignment learning and selection
learning phases to the ATIS and OVIS corpora, respectively.
The ``upper'' selection learning method corresponds to the
results of the previous section, denoting the upper bound of the
results after the selection learning phase.

The results show that almost all \abl systems are better than the
baseline.  Only the \abldis systems on the ATIS corpus and most
of the \constl systems perform slightly worse.  Even though the
f-score of the \abldis systems is lower, these systems do have a
much higher precision than the baseline.

The disappointing results of the \abldis systems can be explained
from the fact that the \abldis alignment learning phase does not
introduce many hypotheses (as shown in the previous section).
Only about 18\% of the number of constituents present in the
ATIS and almost 50\% in the OVIS treebank are correct. However,
almost all correct hypotheses inserted by the alignment learning
phase are still contained in the resulting treebank. This means
that the selection learning phases work relatively well for this
alignment learning instance.  A final remark about the \abldis
systems is that the results can vary wildly, which can already be
expected from the results of the alignment learning phase alone.

\begin{tab}{Results selection learning on the OVIS corpus}
\label{tab:selectovis}
\vspace{1em}
\begin{tabular}{|l|l||*{3}{rr|}}
\hline
&       & \mc{Recall}    & \mc{Precision} & \mc{F-score}   \\
\hline\hline
random
&       & 52.73 & (0.46) & 50.91 & (0.45) & 51.80 & (0.45) \\
\hline
\ablone
& upper & 94.22 & (0.04) & 100.00 & (0.00) & 97.02 & (0.02) \\
&\first & 56.01 & (3.45) &  54.38 & (3.35) & 55.18 & (3.40) \\
&\termsl& 53.63 & (0.11) &  63.78 & (0.10) & 58.27 & (0.10) \\
&\constl& 42.24 & (0.14) &  51.04 & (0.11) & 46.23 & (0.13) \\
\hline
\abldis
& upper & 53.65 & (2.27) & 100.00 & (0.00) & 69.81 & (1.93) \\
&\first & 48.03 & (3.52) &  74.84 & (5.62) & 58.50 & (4.28) \\
&\termsl& 47.63 & (3.08) &  76.30 & (4.60) & 58.64 & (3.66) \\
&\constl& 46.64 & (2.94) &  74.62 & (4.37) & 57.40 & (3.49) \\
\hline
\ablall
& upper & 96.47 & (0.00) & 100.00 & (0.00) & 97.68 & (0.00) \\
&\first & 56.49 & (3.22) &  54.74 & (3.13) & 55.60 & (3.22) \\
&\termsl& 53.95 & (0.07) &  62.15 & (0.08) & 57.76 & (0.07) \\
&\constl& 41.83 & (0.01) &  48.91 & (0.01) & 45.09 & (0.01) \\
\hline
\end{tabular}
\end{tab}

The \constl system uses more precise statistics to select the
best hypotheses. However, it does not perform well. The
hypothesis universe contains many correct, but also incorrect
hypotheses which are used in the computation of the
probabilities. It may be the case that when using more precise
statistics, the incorrect hypotheses have a larger impact on the
final probability, yielding worse results. Apart from this, the
\constl system relies on the non-terminal types of the
hypotheses. However, the types are clustered in an imperfect way
(as described in section~\ref{s:clustering}) which introduces an
extra margin of error.

The \first systems all perform relatively well. The
\ablall:\first system even outperforms all other systems on the
ATIS corpus. From the relatively large standard deviation of
these systems, it can be concluded that the order of the
sentences in the corpora is important.

The \ablone and \ablall systems seem to perform relatively
similar, but the \termsl and \constl systems yield significantly
better results when combined with the \ablone system on both
corpora. Overall, the \ablone systems perform best. They have
high scores and small standard deviations. From the systems
within \ablone, the \termsl system clearly performs best. For the
rest of this chapter, i.e.\ the results on the WSJ corpus, the learning
curve and \parseabl, this system will be used.

%%%%%%%%%%%%%%%%%%%%%%%%%%%%%%%%%%%%%%%%%%%%%%%%%%%%%%%%%%%%%%%%%%%%%%%%%%%%%%%%
\thesissubsubsection{Results on the Wall Street Journal corpus}
%%%%%%%%%%%%%%%%%%%%%%%%%%%%%%%%%%%%%%%%%%%%%%%%%%%%%%%%%%%%%%%%%%%%%%%%%%%%%%%%
\label{s:wsj}
\index{result!Wall Street Journal}%

To test how the system performs on a completely new corpus, it
has been applied to the Wall Street Journal (WSJ) corpus. This
corpus is, like the ATIS corpus, part of the Penn treebank 2. The
\ablone:\termsl system has been applied to section~23 of this
treebank\footnote{Section~23 has informally developed into the
test section of the WSJ corpus (see e.g.\
\citep{Collins:ACL97-16,Charniak:NCAI97-598}).}, which contains
\index{Collins, M.}%
\index{Charniak, E.}%
1,094 sentences.  The WSJ corpus consists of newspaper articles,
which means that the sentences are more complex than the ATIS or
OVIS corpora. The main difference between the corpora is that the
WSJ corpus has a much larger vocabulary size. Where the other two
corpora are samples of a small domain, the WSJ corpus is from a
much larger domain. Apart from that, the mean sentence length is
over 35 words per sentence. Some example sentences can be found
in~\ref{sentwsj}.

\eenumsentence{
\toplabel{sentwsj}
\item At about 3:30 pm EDT S\&P futures resumed
trading and for a brief time the futures and stock markets
started to come back in line
\item In the year quarter the designer and operator of
cogeneration and waste heat recovery plants had net income of
\$ 326,000 or four cents a share on revenue
of about \$ 414 million
\item Under terms of the plan independent generators would be
able to compete for 15 \% of customers until 1994 and
for another 10 \% between 1994 and 1998
}

\begin{tab}{Results \abl on the WSJ corpus}
\label{tab:selectwsj}
\vspace{1em}
\begin{tabular}{|l|*{3}{rr|}}
\hline
                & \mc{Recall}    & \mc{Precision} & \mc{F-score}     \\
\hline\hline
random          & 23.94 & (0.29) &  22.62 & (0.27) & 23.27 & (0.28) \\
upper           & 52.86 & (0.03) & 100.00 & (0.00) & 69.16 & (0.03) \\
\ablone:\termsl & 12.46 & (0.54) &  42.56 & (1.73) & 19.26 & (0.52) \\
\hline
\end{tabular}
\end{tab}

Before looking at the results, it must be mentioned that to our
knowledge, this is the first time an unsupervised language
learning system has been applied to the plain sentences of the
Wall Street Journal corpus.

The results of applying the random baseline system, the upper
bound of the alignment learning phase and the \ablone:\termsl
system to section~23 of the WSJ corpus are shown in
table~\ref{tab:selectwsj}. The baseline system outperforms the
\abl system. However, \ablone:\termsl has a much higher
precision. Note that applying the system to several sections
indicate that the recall decreases slightly, but the precision
improves even more. 

The upper bound shows that many of the correct hypotheses are
being learned. The selection learning phase, however, is unable
to select them. It may be the case that the \termsl selection
learning system does not perform very well when confronted with
more hypotheses compared to the ATIS and OVIS corpora.  Future
work should concentrate on better selection learning methods.
Since there are many hypotheses, the probabilities used in the
\termsl system may not be precise enough. The \constl system or
the systems described in section~\ref{s:parsingselectionlearning}
may perhaps perform better (even though they are based on
imprecise non-terminal type data).

%%%%%%%%%%%%%%%%%%%%%%%%%%%%%%%%%%%%%%%%%%%%%%%%%%%%%%%%%%%%%%%%%%%%%%%%%%%%%%%%
\thesissubsubsection{\parseabl systems}
%%%%%%%%%%%%%%%%%%%%%%%%%%%%%%%%%%%%%%%%%%%%%%%%%%%%%%%%%%%%%%%%%%%%%%%%%%%%%%%%
\label{s:parseeval}

For the evaluation of the \parseabl system, a grammar has been
\index{evaluation!parseABL}%
\index{parseABL!evaluation}%
extracted from each of the treebanks generated by the
\ablone:\termsl system. Each of the grammars have been used to
parse the plain sentences of the ATIS corpus.\footnote{The corpus
has been parsed using the efficient DOPDIS parser by Khalil
\index{DOPDIS parser}%
\index{parser!DOPDIS}%
\citet{Simaan:LED-99}. Only one minor problem
\index{Sima'an, K.}%
was that the \abl systems sometimes learn too ``flat'' structure
(i.e.\ constituents containing too many elements). The parser has
not been optimised for this type of structure.} The results of
the parsed treebanks are shown in table~\ref{tab:parseatis}.  The
first entry is computed by extracting an SCFG and the final entry
\index{SCFG}%
\index{stochastic context-free grammar}%
contains the results of the unparsed treebank (as shown in
table~\ref{tab:selectatis}. The other entries show the results of
parsing the sentences using STSGs with the designated maximum
\index{STSG}%
\index{stochastic tree substitution grammar}%
tree depth.

\begin{tab}{Results \parseabl on the ATIS corpus}
\label{tab:parseatis}
\vspace{1em}
\begin{tabular}{|l||*{3}{rr|}}
\hline
Treedepth & \mc{Recall}  & \mc{Precision}  & \mc{F-score}     \\
\hline\hline
1 & 24.87 & (0.54) & 56.79 & (1.00) & 34.59 & (0.69) \\
2 & 25.62 & (0.17) & 55.38 & (0.49) & 35.03 & (0.25) \\
3 & 25.79 & (0.19) & 54.74 & (0.43) & 35.06 & (0.26) \\
- & 25.82 & (0.19) & 54.73 & (0.42) & 35.09 & (0.25) \\
\hline
\end{tabular}
\end{tab}

Each of the reparsed corpora have a lower recall, but a higher
precision. When the maximum tree depth is increased, the results
grow closer to the unparsed treebank. Since the \dop system has a
preference for shorter derivations and thus has a preference for
the use of larger subtrees \citep{Bod:COLING00-69}, the \stsg
\index{Bod, R.}%
instances that have a higher maximum tree depth will prefer the
larger parts of the structures. This corresponds to the
structures that are present in the unparsed treebank generated by
the \ablone:\termsl system. Increasing the treedepth even more
will probably yield results similar to those of depth 3 and to
the unparsed treebank.

%%%%%%%%%%%%%%%%%%%%%%%%%%%%%%%%%%%%%%%%%%%%%%%%%%%%%%%%%%%%%%%%%%%%%%%%%%%%%%%%
\thesissubsubsection{Learning curve}
%%%%%%%%%%%%%%%%%%%%%%%%%%%%%%%%%%%%%%%%%%%%%%%%%%%%%%%%%%%%%%%%%%%%%%%%%%%%%%%%
\label{s:learningcurve}
\index{learning curve}%

To investigate how the \abl system responds to differences in
amount of input data, the \ablone:\terms system is applied to
corpora of increasing size. The results on the ATIS corpus can be
found in figure~\ref{fig:curveatis} and those on the OVIS corpus
in figure~\ref{fig:curveovis}.

\begin{fig}{Learning curve on the ATIS corpus}
\label{fig:curveatis}
\vspace{-1.3cm}
\includegraphics[width=\textwidth]{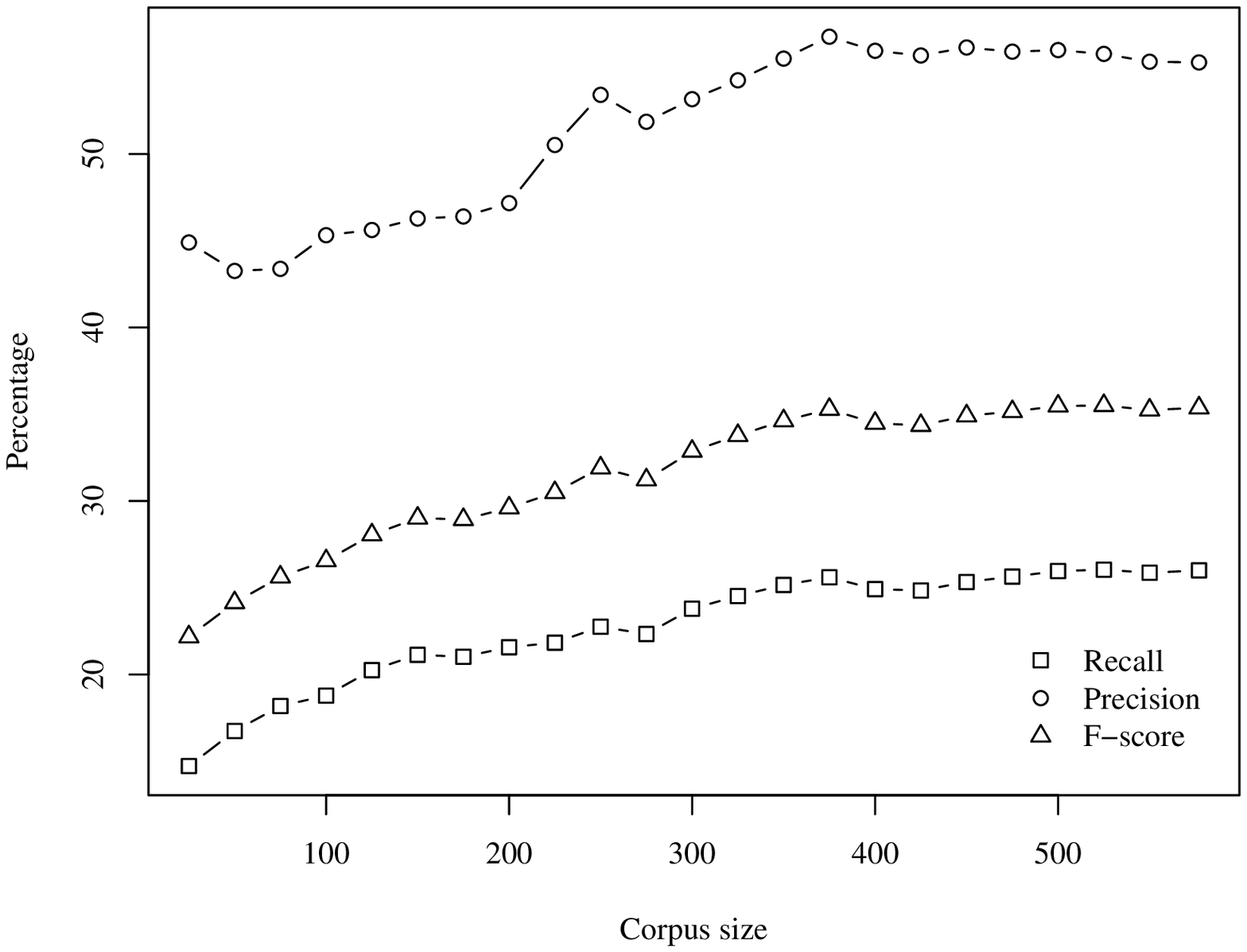}
\vspace{-1cm}
\end{fig}

\begin{fig}{Learning curve on the OVIS corpus}
\label{fig:curveovis}
\vspace{-1.3cm}
\includegraphics[width=\textwidth]{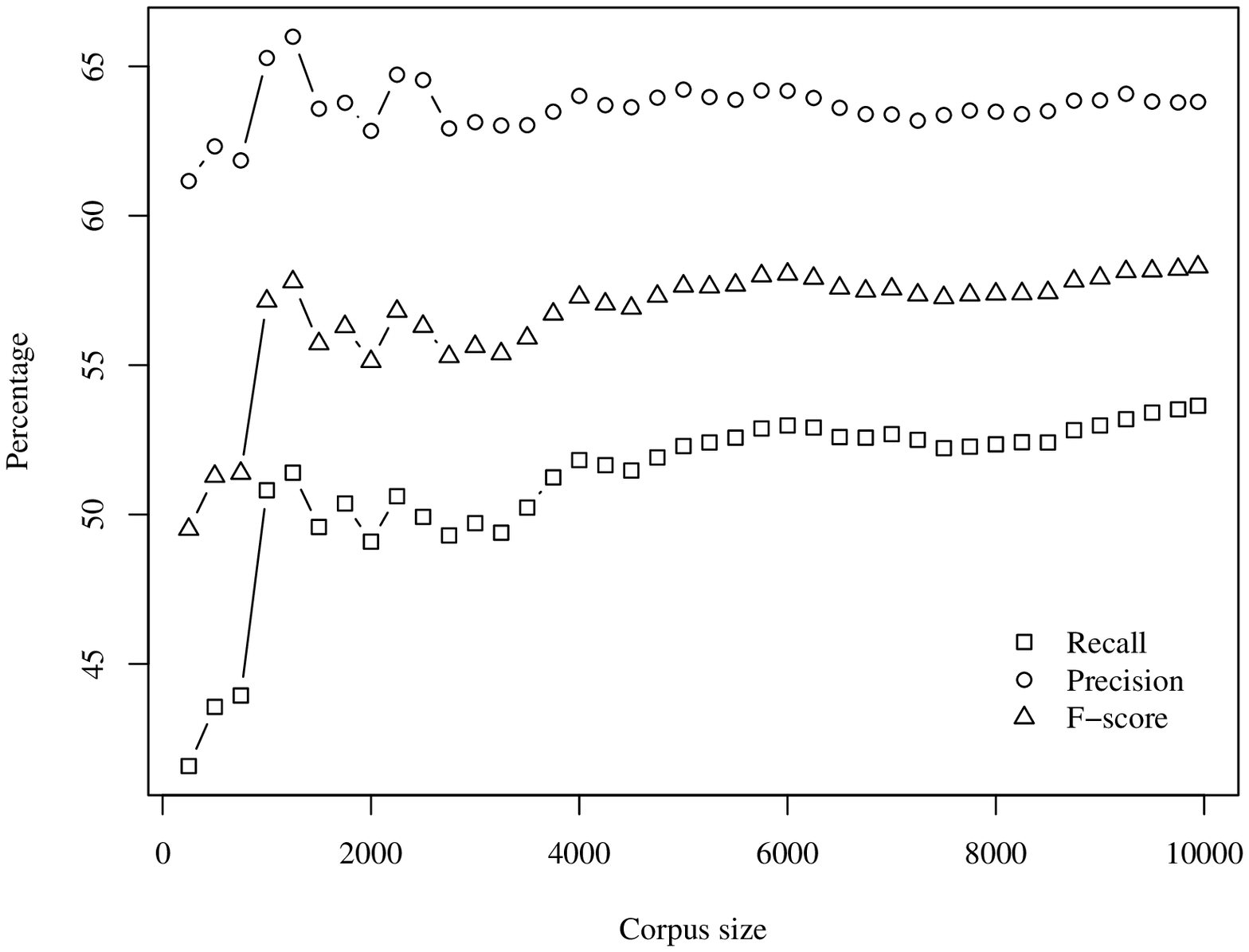}
\vspace{-1cm}
\end{fig}

The measures on both corpora seem to have been stabilised when
the entire corpus is used to learn. It might still be the case
that if the corpora were larger, the results would increase
slightly, but no drastic improvements are to be expected.

It is interesting to see that the recall and precision metrics
both respond similar to changes. The jump in performance that
occurs between sentences 500 and 750 on the OVIS corpus can be
found in all metrics. This shows that the \abl systems is very
balanced. Note that the \ablone:\termsl system has been applied
to the corpora of different size only once in this section. The
large increase in performance is explained by a number of
``easy'' sentences in that range.  The standard deviation of the
results become smaller when more data is available. When the
system is applied ten times (and using the mean as values), this
jump in performance is flattened out.

The system takes a longer time to stabilise on the OVIS corpus
than on the ATIS corpus. The OVIS corpus contains mainly short
sentences. These sentences are easy to structure, since there are
not many possible hypotheses to insert. The ATIS corpus, on the
other hand, has longer sentences, which are all about as
difficult to structure. When longer sentences occur early in the
corpus, the results on the OVIS corpus will fluctuate (as happens
between sentences 500 and 750). If the longer sentences occur
later in the corpus (for example when the order of the sentences
is different), then there is already enough data in the
hypothesis space to absorb the fluctuation.

%%%%%%%%%%%%%%%%%%%%%%%%%%%%%%%%%%%%%%%%%%%%%%%%%%%%%%%%%%%%%%%%%%%%%%%%%%%%%%%%
%%%%%%%%%%%%%%%%%%%%%%%%%%%%%%%%%%%%%%%%%%%%%%%%%%%%%%%%%%%%%%%%%%%%%%%%%%%%%%%%
\thesissection{Qualitative results}
%%%%%%%%%%%%%%%%%%%%%%%%%%%%%%%%%%%%%%%%%%%%%%%%%%%%%%%%%%%%%%%%%%%%%%%%%%%%%%%%
%%%%%%%%%%%%%%%%%%%%%%%%%%%%%%%%%%%%%%%%%%%%%%%%%%%%%%%%%%%%%%%%%%%%%%%%%%%%%%%%
\index{result!qualitative}%
\index{qualitative result}%

Apart from the numerical analysis, the treebanks resulting from
applying the \abl systems are analysed and they exhibit some nice
properties.  This section takes a closer look at the properties
of the generated treebanks. First, a rough ``looks-good-to-me''
\index{evaluation!looks-good-to-me approach}%
approach is taken to evaluate the learned treebanks. Following
this, it will be shown that the generated treebanks contain
recursive structures.

%%%%%%%%%%%%%%%%%%%%%%%%%%%%%%%%%%%%%%%%%%%%%%%%%%%%%%%%%%%%%%%%%%%%%%%%%%%%%%%%
\thesissubsection{Syntactic constructions}
%%%%%%%%%%%%%%%%%%%%%%%%%%%%%%%%%%%%%%%%%%%%%%%%%%%%%%%%%%%%%%%%%%%%%%%%%%%%%%%%
\index{result!syntactic constructions}%

The learned treebanks all contain interesting syntactic
structure. In this section, three different constructions worth
mentioning are discussed. Of course, the structured corpora
contain more, interesting, syntactic constructions, but these three
are most remarkable.

The examples given in this section are taken from a version of
the ATIS corpus, which is structured by the \ablone:\termsl
system. The exact non-terminal types change for each
instantiation, but similar constructions can be found in
each of the structured corpora.

\begin{description}
\item [Noun phrases] The system is able
\index{result!noun phrase}%
\index{noun phrase}%
to consistently learn constituents that are similar to noun
phrases. The sentences in~\ref{exnp} show some examples.

\eenumsentence{
\toplabel{exnp}
\item \empty[How much would the [coach fare cost]\snt{1}]\snt{0}
\item \empty[I need a [flight]\snt{213} from Indianapolis to Houston on
T W A]\snt{0}
\item \empty[List all [flights]\snt{1026} from Burbank to Denver]\snt{0}
}

The \abl system finds almost all noun phrases in the corpus.
However, it inserts constituents that contain the entire noun
phrase except for the determiner. This happens because the
determiners occur frequently, which means that they are often
linked. When the determiners are linked, the parts of the
sentences following (and preceding) the determiner are stored in
a hypothesis.

Note that since the determiners are not inside the noun phrase
constituent, almost all noun phrases are learned incorrectly.
Only noun phrases that do not have a determiner at all or noun
phrases containing a noun only are learned correctly.

\item [From-to phrases] A case that is related to the noun phrase
\index{result!from-to phrase}%
\index{from-to phrase}%
constituents is that of \sent{from} and \sent{to} phrases. Since
the corpus contains air traffic information, \sent{from} and
\sent{to} are words that occur frequently. Like in the previous
case, where determiners are used as a hypothesis boundary,
\sent{from} and \sent{to} are also linked regularly. This results
in constituents (which are mostly names of places) as shown in
the sentences in~\ref{exft}. It implies that all names of places
are found correctly.

\eenumsentence{
\toplabel{exft}
\item \empty[What are the flights from [Milwaukee]\snt{54} to
[Tampa]\snt{55}]\snt{0}
\item \empty[Show me the flights from [Newark]\snt{54} to [Los
Angeles]\snt{55}]\snt{0}
\item \empty[I would like to travel from [Indianapolis]\snt{54} to
[Houston]\snt{55}]\snt{0}
}

It is interesting to see that even the non-terminal types of the
two constituents are consistent. A \sent{from}-phrase has type
\nt{54} and a \sent{to}-phrase has type \nt{55}.

\item [Part-of-speech tags] Apart from the larger constituents
described above, \abl finds many part-of-speech tags of for
example verbs, nouns.  Many verbs are clustered into the same
non-terminal type. All forms of the verb ``to be'' occuring in
the corpus are grouped together, but also verbs that occur on the
first position in the sentence have mostly the same type.
Additionally, frequently occuring nouns, like \sent{flight},
\sent{flights}, \sent{number} and \sent{dinner} have the same
non-terminal types consistently.
\end{description}

%%%%%%%%%%%%%%%%%%%%%%%%%%%%%%%%%%%%%%%%%%%%%%%%%%%%%%%%%%%%%%%%%%%%%%%%%%%%%%%%
\thesissubsection{Recursion}
%%%%%%%%%%%%%%%%%%%%%%%%%%%%%%%%%%%%%%%%%%%%%%%%%%%%%%%%%%%%%%%%%%%%%%%%%%%%%%%%
\index{recursion}%
\index{recursive structure}%
\index{structure!recursive}%

All tested treebanks structured by the \abl systems contain
recursive structure. This section will concentrate on examples
taken from the structured ATIS corpus. To be completely clear
about what recursion is, it is defined here as follows.

\begin{definition}[Recursion in a tree]
A (fuzzy) tree $T=\tuple{$S$, $C$}$ contains \emph{recursive
structure} iff there are constituents $c_1=\tuple{$b_1$, $e_1$,
$n_1$}\mbox{ and } c_2=\tuple{$b_2$, $e_2$, $n_2$}\in C$ for
which it holds that $n_1=n_2$ and either $(b_1\leq b_2 \wedge e_1\geq
e_2)$ or $(b_1\geq b_2 \wedge e_1\leq e_2)$.
\end{definition}

\begin{definition}[Recursion in a treebank or structured corpus]
A treebank or structured corpus is said to contain
\emph{recursive structure} iff at least one tree or fuzzy tree
contains recursive structure.
\end{definition}

The sentences~\ref{rec1},~\ref{rec2}, and \ref{rec3} are examples
of recursion in trees found in the ATIS corpus. The first
sentence of the pair is the learned tree, while the second is the
original tree structure. These particular structures can be found
in the corpus generated by the \ablone:\termsl system, but the
other systems learn equivalent recursive structures.

\eenumsentence{
\toplabel{rec1}
\item \empty[Fares less than one [hundred fifty one
[dollars]\snt{32}]\snt{32}]\snt{0}
\item \empty[Fares less than [[one hundred fifty one]\snt{QP}
dollars]\snt{NP}]\snt{FRAG}
}

In the sentences in~\ref{rec1}, the original tree structure did
not show any recursion. The recursive structure, however, can be
easily explained. If the sentence is aligned with for example the
sentence \sent{Cheapest fare one way}, the word \sent{one} can
aligned against the first or second occurrence. This introduces
the structure as shown, although the non-terminal types may still
be different. At some later point, the different non-terminal
types are merged and the recursive structure is a fact.

\eenumsentence{
\toplabel{rec2}
\item \empty[Is there a flight tomorrow morning from Columbus to [to
[Nashville]\snt{55}]\snt{55}]\snt{0}
\item \empty[Is there a flight tomorrow morning from Columbus [to
]\snt{X} [to Nashville]\snt{PP}]\snt{SQ}
}

The sentences in~\ref{rec2} are a bit strange. The input corpus
contained an error, and this allows different links of the word
\sent{to}, the system learned recursion. Again, the original
corpus did not show any recursion.

\eenumsentence{
\toplabel{rec3}
\item \empty[Is dinner served on the [first leg or the [second
leg]\snt{1}]\snt{1}]\snt{0}
\item \empty[Is dinner served on [[the first leg]\snt{NP} or [the
second leg]\snt{NP}]\snt{NP}]\snt{SQ}
}

The learned tree structure in the final example, which can be
found in~\ref{rec3}, is much more like the original structure. In
fact, the constituents would have been completely correct if only
the determiner was put \emph{inside} the constituents. The fact
why this is not the case has been discussed in the previous
section. 

Intuitively, a recursive structure is formed first by building
the constituents that form the structure of the recursion as
hypotheses with different root non-terminals.  Now the
non-terminals need to be merged. This happens when two partially
structured sentences are compared to each other yielding
hypotheses that already existed in both sentences with the
non-terminals present in the ``recursive'' structure (as
described in sections~\ref{s:inserthypothesis}
and~\ref{s:clustering}). The non-terminals are then merged,
resulting in a recursive structure.

%% end of file: results.tex

%% file: compare.tex
%%%%%%%%%%%%%%%%%%%%%%%%%%%%%%%%%%%%%%%%%%%%%%%%%%%%%%%%%%%%%%%%%%%%%%%%%%%%%%%%
%% Menno van Zaanen                                                           %%
%% menno@comp.leeds.ac.uk                                                     %%
%%%%%%%%%%%%%%%%%%%%%%%%%%%%%%%%%%%%%%%%%%%%%%%%%%%%%%%%%%%%%%%%%%%%%%%%%%%%%%%%
%% Filename: compare.tex                                                      %%
%%%%%%%%%%%%%%%%%%%%%%%%%%%%%%%%%%%%%%%%%%%%%%%%%%%%%%%%%%%%%%%%%%%%%%%%%%%%%%%%

%%%%%%%%%%%%%%%%%%%%%%%%%%%%%%%%%%%%%%%%%%%%%%%%%%%%%%%%%%%%%%%%%%%%%%%%%%%%%%%%
%%%%%%%%%%%%%%%%%%%%%%%%%%%%%%%%%%%%%%%%%%%%%%%%%%%%%%%%%%%%%%%%%%%%%%%%%%%%%%%%
%%%%%%%%%%%%%%%%%%%%%%%%%%%%%%%%%%%%%%%%%%%%%%%%%%%%%%%%%%%%%%%%%%%%%%%%%%%%%%%%
\thesischapter{\abl versus the World}
{This world is spinning around me.}
{Dream Theater (Images and Words)}
%%%%%%%%%%%%%%%%%%%%%%%%%%%%%%%%%%%%%%%%%%%%%%%%%%%%%%%%%%%%%%%%%%%%%%%%%%%%%%%%
%%%%%%%%%%%%%%%%%%%%%%%%%%%%%%%%%%%%%%%%%%%%%%%%%%%%%%%%%%%%%%%%%%%%%%%%%%%%%%%%
%%%%%%%%%%%%%%%%%%%%%%%%%%%%%%%%%%%%%%%%%%%%%%%%%%%%%%%%%%%%%%%%%%%%%%%%%%%%%%%%
\label{ch:compare}

The general framework has been described and tested in the
previous chapters, so now it can be compared against other
systems. This chapter will relate \abl to other learning systems.
In addition, similarities between the \abl and Data-Oriented
Parsing frameworks will be discussed. 

This chapter first describes the previous work, giving \abl a
niche in the world of language learning systems. Following this,
\abl is compared against other language learning systems. Next,
\abl is extensively compared to the EMILE system, since EMILE is
in many ways similar to \abl and finally, the relationships
between \abl and the \dop framework will be discussed. Even
though \dop is not a language learning system, they have many
similarities.

%%%%%%%%%%%%%%%%%%%%%%%%%%%%%%%%%%%%%%%%%%%%%%%%%%%%%%%%%%%%%%%%%%%%%%%%%%%%%%%%
%%%%%%%%%%%%%%%%%%%%%%%%%%%%%%%%%%%%%%%%%%%%%%%%%%%%%%%%%%%%%%%%%%%%%%%%%%%%%%%%
\thesissection{Background}
%%%%%%%%%%%%%%%%%%%%%%%%%%%%%%%%%%%%%%%%%%%%%%%%%%%%%%%%%%%%%%%%%%%%%%%%%%%%%%%%
%%%%%%%%%%%%%%%%%%%%%%%%%%%%%%%%%%%%%%%%%%%%%%%%%%%%%%%%%%%%%%%%%%%%%%%%%%%%%%%%

Before going into a discussion of language learning systems, it
must be mentioned that there is a whole research area dedicated
to a formal description of the learnability of languages.
\index{learnability}%
\citet{Lee:LCFL-96} gives a nice overview of this field.
\index{Lee, L.}%

One of the negative results in the theoretical field of language
learning is that by \citet{Gold:67-10-447} which proved that
\index{Gold, E.M.}%
learning context-free grammars from text (only) is in general
\index{learnability!of context-free grammars}%
impossible.\footnote{Although the learning of context-free
grammars in general is not possible,
\citeauthor{Gold:67-10-447}'s theorems do not cover all cases,
such as for example finite grammars \citep[pp.\
43--44]{Adriaans:LLC-92}.} Another result, which is even more
\index{Adriaans, P.W.}%
important in our case, is by \citet{Horning:SGI-69}, who showed
\index{Horning, J.J.}%
that \emph{stochastic} context-free grammars are indeed
\index{learnability!of stochastic context-free grammars}%
learnable. On the other hand, \citeauthor{Gold:67-10-447}'s
concept of identification in the limit has been amended with the
notion of PAC (Probably Approximately Correct) and PACS (PAC
\index{PAC learning}%
learning under Simple distributions) learning
\index{PACS learning}%
\citep{Valiant:84-27-1134,Li:91-5-911}. The idea with PAC
\index{Valiant, L.G.}%
learning is to minimise the chance of learning something wrong,
without being completely sure to be right.

Existing (language) learning algorithms can be roughly divided
into two groups, \emph{supervised} and \emph{unsupervised}
\index{supervised}%
\index{unsupervised}%
\index{learning!supervised}%
\index{learning!unsupervised}%
learning algorithms, based on the type of information they use.
All learning algorithms use a \emph{teacher} that gives examples
\index{teacher}%
of (unstructured) sentences in the language. In addition, some
algorithms use a \emph{critic} (also called an \emph{oracle}). A
\index{critic}%
\index{oracle}%
critic may be asked if a certain sentence (possibly including
structure) is a valid sentence in the language.  The algorithm
can use a critic to validate hypotheses about the
language.\footnote{When an algorithm uses a treebank or
structured corpus to initialise, it is said to be supervised. The
structure of the sentences in the corpus can be seen as the
critic.} Supervised language learning methods use a teacher and a
critic, whereas the unsupervised methods only use a teacher
\citep{MLNL97}.
\index{Powers, D.M.P.}%

Apart from the division of learning systems based on their
\emph{information}, systems can also be separated based on the
types of \emph{data} they use. Some systems learn using
\emph{positive data} only, whereas other methods use
\index{positive data}%
\emph{complete information} (positive as well as negative).
\index{negative data}%
\index{complete data}%

Figure~\ref{fig:ontologylls} shows the different types of
language learning systems based on the type of information and
type of data used. The \abl framework falls in the type 4
class, since it is unsupervised (it does not use a critic) and it
uses only positive information (\abl is fed with sentences
that can be generated by the language only).

\definecolor{gray}{rgb}{.8,.8,.8}
\newcommand{\panel}[1]{\multicolumn{1}{|>{\columncolor{gray}}c|}{#1}}
\begin{fig}{Ontology of learning systems}
\label{fig:ontologylls}
\vspace{1em}
\begin{tabular}{|c|r|c|c|}\hline
\multicolumn{2}{|c|}{} & \multicolumn{2}{c|}{Type of information}\\\cline{3-4}
\multicolumn{2}{|c|}{} & Supervised & Unsupervised               \\\hline
Type of & Complete     & Type 1     & Type 2                     \\\cline{2-4}
data    & Positive     & Type 3     & \panel{Type 4}             \\\hline
\end{tabular}
\end{fig}

Supervised language learning methods typically generate better
results. These methods can tune their output, since they receive
knowledge of the structure of the language (by initialisation or
querying a critic). In contrast, unsupervised language learning
methods do not receive these structured sentences, so they do not
know at all what the output should look like and therefore cannot
adjust the output towards the ``expected'' output.

Although unsupervised methods perform worse than supervised
methods, unsupervised methods are necessary for the (otherwise)
time-consuming and costly creation of treebanks of languages for which
no
\index{treebank!build}%
\index{structured corpus!build}%
initial treebank nor grammar yet exists. This indicates that there is
a strong practical motivation to develop unsupervised grammar
induction algorithms that work on plain text.

%%%%%%%%%%%%%%%%%%%%%%%%%%%%%%%%%%%%%%%%%%%%%%%%%%%%%%%%%%%%%%%%%%%%%%%%%%%%%%%%
%%%%%%%%%%%%%%%%%%%%%%%%%%%%%%%%%%%%%%%%%%%%%%%%%%%%%%%%%%%%%%%%%%%%%%%%%%%%%%%%
\thesissection{Bird's-eye view over the world}
%%%%%%%%%%%%%%%%%%%%%%%%%%%%%%%%%%%%%%%%%%%%%%%%%%%%%%%%%%%%%%%%%%%%%%%%%%%%%%%%
%%%%%%%%%%%%%%%%%%%%%%%%%%%%%%%%%%%%%%%%%%%%%%%%%%%%%%%%%%%%%%%%%%%%%%%%%%%%%%%%

This section will briefly describe some of the existing language
learning systems. First, systems using complete information
(types~1 and~2) will be described.  Next, the systems which use
positive information only will be considered, subdividing these
into supervised (type~3) and unsupervised (type~4).

When progressing the next sections, more detail will be given
towards types and systems that are more closely related to \abl,
but even the section on unsupervised systems using only positive
information is not meant to be a complete overview of the
(sub-)field. These short descriptions are merely given to
illustrate the ideas of the established work in this area.

%%%%%%%%%%%%%%%%%%%%%%%%%%%%%%%%%%%%%%%%%%%%%%%%%%%%%%%%%%%%%%%%%%%%%%%%%%%%%%%%
\thesissubsection{Systems using complete information}
%%%%%%%%%%%%%%%%%%%%%%%%%%%%%%%%%%%%%%%%%%%%%%%%%%%%%%%%%%%%%%%%%%%%%%%%%%%%%%%%
\index{system!using complete information}%

For completeness sake, this section will describe two systems
that use positive \emph{as well as} negative examples to learn a
grammar. First, a supervised system which uses partially
structured sentences will be described, followed by an
unsupervised system.

\citet{Sakakibara:ICGI00-229} describe a system that induces a
\index{Sakakibara, Y.}%
grammar using partially structured sentences. The partially
structured sentences are stored in a tabular representation
(similar to the one used in the CYK algorithm
\index{CYK}%
\index{Cocke-Younger-Kasami}%
\citep{Younger:67-10-189}). A genetic algorithm partitions the
\index{Younger, D.H.}%
\index{genetic algorithm}
set of non-terminals, effectively merging certain non-terminals.
The different possible partitions are then tested against the
negative examples.

The system by \citet{Nakamura:ICGI00-186} learns a context-free
\index{Nakamura, K.}%
\index{Ishiwata, T.}%
grammar from positive and negative examples by adapting the CYK
algorithm, which introduces a grammar rule if the sentence is
otherwise not parsable. If the introduction of a grammar rule
allows for parsing a negative example, the system returns
failure.

This approach is in a way similar to the alignment learning phase
of \abl. Parts of sentences that cannot be parsed (which are
parts that are unequal to the right-hand side of the known
grammar rules) are used to introduce hypotheses. However, no
disambiguation takes place, the system returns failure when
overlapping hypotheses are introduced (in contrast to \abl which
has selection learning as a disambiguation phase).

Both systems are evaluated by letting the system rebuild a known
context-free grammar. The emphasis of the first system, however,
is on how many iterations of the genetic algorithm are needed to
find a grammar similar to the original one, whereas the second
system concentrates on processing time needed to find the
grammar.  This approach unfortunately makes it impossible to
compare the two systems directly. Furthermore, it is unclear how
the two methods would perform on real natural language data.

%%%%%%%%%%%%%%%%%%%%%%%%%%%%%%%%%%%%%%%%%%%%%%%%%%%%%%%%%%%%%%%%%%%%%%%%%%%%%%%%
\thesissubsection{Systems using positive information}
%%%%%%%%%%%%%%%%%%%%%%%%%%%%%%%%%%%%%%%%%%%%%%%%%%%%%%%%%%%%%%%%%%%%%%%%%%%%%%%%
\index{system!using positive information}%

The systems described in this section (which use positive
examples only) are more closely related to the \abl system than
the systems in the previous section. First, some supervised
systems will be described, followed by a section on unsupervised
systems.

The section on unsupervised systems also contains some systems
that learn word categories or segment sentences only. Even though
these systems are different from \abl in that \abl learns
context-free grammars, they are treated here because they start
with similar information and are in some ways similar to \abl.

%%%%%%%%%%%%%%%%%%%%%%%%%%%%%%%%%%%%%%%%%%%%%%%%%%%%%%%%%%%%%%%%%%%%%%%%%%%%%%%%
\thesissubsubsection{Supervised systems}
%%%%%%%%%%%%%%%%%%%%%%%%%%%%%%%%%%%%%%%%%%%%%%%%%%%%%%%%%%%%%%%%%%%%%%%%%%%%%%%%
\index{supervised system!using positive information}%
\index{system!using positive information!supervised}%

The supervised Transformation-Based Learning system described in
\index{TBL}%
\index{transformation-based learning}%
\citep{Brill:ACL93-259}, is a non-probabilistic system. It starts
\index{Brill, E.}%
with naive knowledge on structured sentences (for example right
branching structures). The system then compares these structured
sentences against the correctly structured examples. From the
differences between the two, the system learns
``transformations'', which can transform the naive structure into
\index{transformation}%
the correct structure. The learned transformations can then be
used to correctly structure new (unstructured) text, by first
assuming the naive structure and then applying the
transformations. The system is evaluated by applying it to the
Wall Street Journal treebank, which yields very good results.
However, the initial structure of the sentences is taken to be
right branching, which is already quite similar to the structure
of English sentences. This means that not many transformations
are needed to convert it into the correct structure. It is
unclear how well this method would work on a corpus of a mainly
left branching language.

The ALLiS system \citep{Dejean:CONLL00-95} is based on theory
\index{D\'ejean, H.}%
\index{ALLiS}%
refinement. It starts by building a roughly correct grammar based
on background knowledge.  This initial grammar is extracted from
structured examples. When the grammar is confronted to the
bracketed corpus, revision points in the grammar are found. For
these revision points, possible revisions are created. The best
of these is chosen to revise the grammar. This is repeated until
no more revision points are found. Like the previous system, this
method is evaluated on a natural language treebank (although it
is not mentioned which treebank is used). The evaluation
concentrates on the structure of noun phrases and verb phrases
only, therefore, it is unclear how well this method can generate
structure for complete sentences.

In his thesis, \citet{Osborne:LUB-94} builds a grammar learning
\index{Osborne, M.}%
system by combining a model-driven with a data-driven approach.
\index{model-driven system}%
\index{data-driven system}%
The model-driven system starts with a grammar and meta-rules
which describe how to introduce new grammar rules. The
data-driven system extracts counts from a structured corpus and
uses these counts to prune the new rules induced by the
meta-rules. The system is extensively tested on a treebank,
measuring several different metrics. These tests show that the
combination of data-driven and model-driven approach performs
best. However, it is unclear how the quality of the initial
grammar, needed for the model-driven part of the system,
influences the results of the entire system.

\citet{Pereira:ACL92-128} describe a system that uses a partially
\index{Pereira, F.}%
\index{Schabes, Y.}%
bracketed corpus to infer parameters of a Stochastic Context-Free
Grammar (SCFG) using inside-outside reestimation. It is tested
\index{inside-outside reestimation}%
by letting the system rebuild a grammar (one that generates
palindrome sentences) and by applying it to the ATIS corpus
(which is a different version of the one that is used in this
thesis). It is possible to use this system on a raw (unbracketed)
corpus, but the results decrease drastically.
\citet{Hwa:ACL99-73} uses a variant of
\index{Hwa, R.}%
\citeauthor{Pereira:ACL92-128} system in which the inferred
grammars are represented in a different formalism, which is
slightly more efficient. This version is again tested on the ATIS
(and, additionally, on the Wall Street Journal) corpus, however,
the system is pre-trained on 3600 fully structured sentences from
the Wall Street Journal treebank.

The algorithm described by \citet{Sakakibara:92-97-23} is in many
\index{Sakakibara, Y.}%
ways similar to the algorithm by \citet{Sakakibara:ICGI00-229},
\index{Sakakibara, Y.}%
\index{Muramatsu, H.}%
which uses complete data. Again, structured examples are used to
initialise the grammar, but instead of using negative information
to decide which partitions can be used to merge non-terminals,
the algorithm merges non-terminals to make the grammar
reversible.\footnote{A grammar is called reversible if:
\index{grammar!reversible}%
\begin{enumerate}
\item it is invertible, that is, $A\rightarrow\alpha$ and $B\rightarrow\alpha$
implies $A=B$, and
\item it is reset-free, that is, $A\rightarrow\alpha B\beta$ and
$A\rightarrow\alpha C\beta$ implies $B=C$.
\end{enumerate}} \citet{Nevado:ICGI00-196} describe a version of the
\index{Nevado, F.}%
\index{S\'anchez, J-A.}%
\index{Bened\'\i, J-M.}%
\citeauthor{Sakakibara:92-97-23} algorithm generating a stochastic
context-free grammar. This method has been tested on a subset of the Wall
Street Journal treebank, but the only metrics mentioned are the
number of iterations the algorithm needed, the number of learned grammar
rules and the perplexity of the learned grammar.

%%%%%%%%%%%%%%%%%%%%%%%%%%%%%%%%%%%%%%%%%%%%%%%%%%%%%%%%%%%%%%%%%%%%%%%%%%%%%%%%
\thesissubsubsection{Unsupervised systems}
%%%%%%%%%%%%%%%%%%%%%%%%%%%%%%%%%%%%%%%%%%%%%%%%%%%%%%%%%%%%%%%%%%%%%%%%%%%%%%%%
\index{system!using positive information!unsupervised}%

Before discussing unsupervised grammar induction systems, a brief
overview of systems that learn syntactic categories will be
given. Next, an article which describes systems that find word
boundaries is mentioned. Finally, unsupervised grammar induction
systems using positive data only will be treated.

\citet{Huckle:ACL95-??} gives a brief overview of systems that
\index{Huckle, C.C.}%
cluster (semantically) similar words based on the distribution of
the contexts of the words and their psychological relevance. His
system uses a Naive Bayes method \citep{Duda:PCSA-73}, which for
\index{Duda, R.O.}%
\index{Hart, P.E.}%
\index{naive Bayes}%
each word, counts occurrences of words in the contexts of the
considered word.  The distance between two words is computed by
taking into account the counts of the words in the different
contexts. However, the evaluation of the systems, using the
looks-good-to-me approach, is meager. 

The system described in \citep{Finch:SHOE92-230}
\index{Finch, S.}%
\index{Chater, N.}%
bootstraps a set of categories. Words in the input text are
classified in the same category when they can be replaced in the
same contexts (i.e.\ according to a similarity measure). It is
\index{similarity measure}%
based on bigram statistics describing the contexts of the words.
\index{bigram statistics}%
This system can also be used to classify short sequences of
words. The article by \citet{Redington:98-22-425} contains the
\index{Redington, M.}%
\index{Chater, N.}%
\index{Finch, S.}%
results of several experiments of a similar system that
classifies words using distributional information. A system based
on neural networks can be found in \citep{Honkela:IANN95-3}.
\index{Honkela, T.}%
\index{Pullki, V.}%
\index{Kohonen, T.}%
Using a Self-Organising Map (SOM) the words of the input text are
\index{SOM}%
\index{self-organising map}%
roughly clustered according to their semantic type. All articles
evaluate their system using the looks-good-to-me approach, which
makes it impossible to compare them directly. Additionally, the
article by~\citeauthor{Redington:98-22-425} makes a more formal
evaluation by computing accuracy, completeness and
informativeness.

The \abl framework is in some ways a generalisation of the
systems described above. Where these systems take a fixed window
size for the context of a word or word sequence, \abl considers
the entire sentence as context. If there is some context that can
be found in at least two sentences, \abl will introduce the
hypotheses of the words within that context, i.e.\ the unequal
parts. This allows \abl to learn constituents of any size.

\citet{Brent:99-34-71} describes the comparison of a variety of
\index{Brent, M.R.}%
systems (by other people) that segment sentences finding word
boundaries (which were not present in the input data). The
systems do not generate grammars, but some structure (in the form
of word boundaries) is found. The system is evaluated on the
CHILDES corpus, which contains phonemic transcriptions of
child-directed English, by computing recall and precision
metrics.  This system does not learn any further syntactic
structure, it is only evaluated on how well it finds word
boundaries.

Algorithms that use the minimum description length (MDL)
\index{MDL}%
\index{minimum description length}%
principle build grammars that describe the input sentences using
the minimal number of bits.  The MDL principle results in
grouping re-occurring parts of sentences yielding a reduction in
the amount of information needed to describe the corpus.  The
system by \citet{Grunwald:CSS-94-203} makes use of the MDL
\index{Gr\"unwald, P.}%
principle.  Similarly,
\citet{deMarcken:ACL/Student95-??,deMarcken:ULA-96,deMarcken:UAL-99}
\index{de Marcken, C.}%
uses the MDL principle to find structure in (unsegmented) text.
This system finds word boundaries and inner-word structure.
Most of these articles only perform a looks-good-to-me
evaluation. Only \citet{deMarcken:ULA-96} does a more formal
\index{de Marcken, C.}%
evaluation. The recall and crossing-brackets rate is computed.
Unfortunately, it uses other corpora than the
\citet{Brent:99-34-71} article, which again makes it impossible
\index{Brent, M.R.}%
to compare the systems.

The \abl system does not make use of the MDL principle, but by
introducing hypotheses, it indicates how the plain sentences can
be compressed (as shown in figure~\vref{fig:grammarsize}). By
taking the unequal parts of sentences as hypotheses it compresses
the input sentences better than when using the equal parts of
sentences as hypotheses.

\citet{Stolcke:BLP-94} and \citet{Stolcke:ICGI94-106} describe a
\index{Stolcke, A.}%
\index{Omohundro, S.}%
grammar induction method that merges elements of models using a
Bayesian framework. At first, a simple model is generated
\index{Bayesian framework}%
\index{Bayesian model merging}%
(typically, just the set of examples).  These examples are then
chunked (i.e.\ examples are split into sub-elements). By merging
the elements of this set, more complex models arise. Elements are
merged guided by the Bayesian posterior probability (which
indicates when the resulting grammar is ``simpler''). Evaluation
is again done by rebuilding a grammar. In the \abl framework,
non-terminals are merged in the clustering step as described in
section~\vref{s:clustering}. Future extensions may take an
approach similar to the one described in these articles (see
section~\vref{s:duallevel}).

\citet{Chen:ACL95-228} presents a Bayesian grammar induction
\index{Chen, S.F.}%
method, which is followed by a post-pass using the inside-outside
algorithm \citep{Lari:90-4-35}. The system starts with a grammar
\index{Lari, K.}%
\index{Young, S.J.}%
that generates left-branching tree structures. The grammar rules
are then changed and the probability of the resulting grammar
based on the observations (example sentences) is computed,
where smaller grammars are favoured over larger ones.
Afterwards, the grammar is rebuilt using the inside-outside
algorithm. This method is tested on the part-of-speech tags of
the WSJ corpus, where the entropy of the resulting
grammar is used as the evaluation metric.

Similarly, \citet{Cook:76-10-59} describe a hill-climbing
\index{Cook, C.M.}%
\index{Rosenfeld, A.}%
\index{Aronson, A.R.}%
\index{hill-climbing}%
algorithm (which chooses another grammar if the cost of that
grammar is better than the cost of the current grammar). The cost
function ``measures the complexity of a grammar, as well as the
discrepancy between its language and a given stochastic language
sample.'' The method is evaluated by letting the system rebuild
some simple context-free grammars and examining the result using
the looks-good-to-me approach.

The system by \citeauthor{Wolff:75-66-79} has a long history in
the field of language learning. His earlier work describes a
sentence segmentation system called MK10
\index{MK10}%
\citep{Wolff:75-66-79,Wolff:77-68-97}. It computes the joint
\index{Wolff, J.G.}%
frequencies of contiguous elements in the sentence. When a pair
of elements occurs regularly, it is taken to be an element
itself. Later work
\citep{Wolff:80-23-255,Wolff:82-2-57,Wolff:CPL-88-179} describes
\index{Wolff, J.G.}%
SNPR, which is more directed towards finding context-free
\index{SNPR}%
grammars describing the example sentences. The system is
explained from the viewpoint of compression
\citep{Wolff:CLP-96-67,Wolff:PICE-98,Wolff:PICS-98}. Again, the
\index{Wolff, J.G.}%
systems are evaluated using a looks-good-to-me approach.

Sequitur is a system developed by \citet{Nevill-Manning:97-7-67}
\index{Nevill-Manning, C.G.}%
\index{Witten, I.H.}%
\index{Sequitur}%
and is in many ways related to the SNPR system. It generates a
grammar by incrementally inserting the words of the sample
sentences in the grammar. Sequitur then makes sure that the
following constraints are always satisfied: \emph{digram
\index{digram uniqueness}%
uniqueness}, which means that no pair of adjacent symbols (words)
appears more than once in the grammar and \emph{rule utility},
\index{rule utility}%
which makes sure that every rule is used more than once. An
interesting feature of this system is that it runs in linear time
and space. The system is mostly evaluated on a formal basis (in
the form of time and space complexity). Other than that, it has
been applied to several corpora, but only some features of the
learned structure are discussed (very briefly).

\citet{Magerman:AAAI90-984} describe a method that finds
\index{Magerman, D.M.}%
\index{Marcus, M.}%
constituent boundaries (called \emph{distituents}) using mutual
\index{distituents}%
information values of the part of speech \mbox{n-grams} within a
sentence. The mutual information describes which words cannot be
\index{mutual information}%
adjacent within a constituent. Between these words there should
be a constituent boundary. The evaluation of this system is
vague. It has been applied to a corpus, but the results are only
given as rough mean error rates.

The system by \citet{Clark:ACL/CNLL01-105} combines several
\index{Clark, A.}%
techniques. First of all, it uses distributional clustering to
find grammar rules. A mutual information criterion is then used
to remove incorrect non-terminals. These ideas are incorporated
into a MDL algorithm. The system has been evaluated similarly to
the evaluation of the \abl system in this thesis, even using the
same corpus and evaluation metrics. The main difference with
\abl, however, is that Clark's system has been trained on the
large British National Corpus before applying it to the ATIS
corpus.  Furthermore, the system is tested on part-of-speech tags
instead of the plain words.

\citet{Klein:ACL/CNLL01-113} describe two systems: the
\index{Klein, D.}%
\index{Manning, C.D.}%
\emph{Greedy-Merge} system clusters part-of-speech tags according
\index{greedy-merge}%
to a cost (divergence) function, whereas the
\emph{Constituency-Parser} ``learns distributions over sequences
\index{constituency-parser}%
representing the probability that a constituent is realized as
that sequence.'' The exact details of the systems are hard to
understand from the article, but both systems are evaluated on
the part-of-speech tags of the Wall Street Journal corpus. This
allows it to be roughly compared against the system by
\citeauthor{Clark:ACL/CNLL01-105}.

From the description of the systems it may be clear that it is
nearly impossible to compare two systems. The systems are divided
over the three ways of evaluation (as described in
section~\ref{s:approach}), but even the systems that evaluate
using the same approach as in this thesis are nearly impossible
to compare, since other (subsets of) corpora, grammars or metrics
are used.

The evaluation of methods described by
\citet{Clark:ACL/CNLL01-105}, \citet{Klein:ACL/CNLL01-113} and
\index{Clark, A.}%
\index{Klein, D.}%
\index{Manning, C.D.}%
\citet{Pereira:ACL92-128} comes reasonably close to ours.
\index{Pereira, F.}%
\index{Schabes, Y.}%
\citet{Pereira:ACL92-128} use a slightly different version of the
\index{Pereira, F.}%
\index{Schabes, Y.}%
ATIS corpus.  \citet{Clark:ACL/CNLL01-105} trains his system on
\index{Clark, A.}%
the BNC corpus. Unlike the \abl system, all systems (including
the one by \citet{Klein:ACL/CNLL01-113}) use sequences of
\index{Klein, D.}%
\index{Manning, C.D.}%
part-of-speech tags as sentences.

Now several grammar induction systems have been discussed
briefly, we will take a more detailed look at the EMILE system,
\index{EMILE}%
since it is the system most similar to \abl. Most of the next
section has been previously published in
\citet{vanZaanen:BNAIC01-??,vanZaanen:CTU-01}.
\index{van Zaanen, M.M.}%

%%%%%%%%%%%%%%%%%%%%%%%%%%%%%%%%%%%%%%%%%%%%%%%%%%%%%%%%%%%%%%%%%%%%%%%%%%%%%%%%
%%%%%%%%%%%%%%%%%%%%%%%%%%%%%%%%%%%%%%%%%%%%%%%%%%%%%%%%%%%%%%%%%%%%%%%%%%%%%%%%
\thesissection{Zooming in on EMILE}
%%%%%%%%%%%%%%%%%%%%%%%%%%%%%%%%%%%%%%%%%%%%%%%%%%%%%%%%%%%%%%%%%%%%%%%%%%%%%%%%
%%%%%%%%%%%%%%%%%%%%%%%%%%%%%%%%%%%%%%%%%%%%%%%%%%%%%%%%%%%%%%%%%%%%%%%%%%%%%%%%

The EMILE 4.1\footnote{EMILE 4.1 is a successor to EMILE 3.0,
\index{EMILE}%
conceived by \citeauthor{Adriaans:LLC-92}. The original acronym
stands for Entity Modelling Intelligent Learning Engine. It
refers to earlier versions of EMILE that also had semantic
capacities. The name EMILE is also motivated by the book on
education by J.-J. Rousseau.} algorithm is motivated by the
concepts behind categorial grammar and it falls into the PACS
\index{categorial grammar}%
\index{grammar!categorial}%
paradigm \citep{Li:91-5-911}. The theoretical concepts used in
\index{Li, M.}%
\index{Vit\'anyi, P.M.B.}%
EMILE 4.1 are elaborated on in articles on EMILE 1.0/2.0
\citep{Adriaans:LLC-92} and EMILE 3.0 \citep{Adriaans:LSC-99}.
\index{Adriaans, P.W.}%
More information on the precursors of EMILE 4.1 may be found in
the above articles, as well as in D\"ornenburg's
\citeyearpar{Dornenburg:EEA-97} Master's thesis. The EMILE 4.1
\index{D\"ornenburg, E.}%
algorithm was designed and implemented by
\citet{Vervoort:GWG-00}. \citet{Adriaans:SOFSEM00-173} report
\index{Vervoort, M.R.}%
\index{Adriaans, P.W.}%
some experiments using the EMILE system on large corpora.

The general idea behind EMILE is the notion of identification of
substitution classes by means of clustering. If a language has a
\index{substitution class}%
context-free grammar, then expressions that are generated from the
same non-terminal can be substituted for each other in each
context where that non-terminal is a valid constituent.
Conversely, if there is a sufficiently rich sample from this
language available, then one expects to find classes of
expressions that cluster together in comparable contexts.
Figure~\ref{fig:emile:clustering} illustrates how EMILE finds
clusters and contexts. The context \sent{Oscar likes} occurs with
the expressions \sent{all dustbins} and \textit{biscuits}.
Actually, this type of clustering can be seen as a form of text
compression~\citep{Grunwald:CSS-94-203}.
\index{Gr\"unwald, P.}%

EMILE's notion of substitution classes exactly coincides with the
notion depicted in figure~\vref{fig:derivation}, which shows that
unequal parts of sentences can easily be generated from the same
non-terminal.

\begin{fig}{Example clustering expressions in EMILE}
\label{fig:emile:clustering}
\vspace{1em}
\begin{tabular}{|l|l|}
\hline
Sentences & Structure \\
\hline\hline
\sent{\align{Oscar likes} all dustbins} &
  \nt{1}\der\sent{Oscar likes} \nt{2}\\
\sent{\align{Oscar likes} biscuits} &
  \nt{2}\der\sent{all dustbins}\\
& \nt{2}\der\sent{biscuits}\\
\hline
\end{tabular}
\end{fig}

This finding gives rise to the hypothesis (possibly unjustified)
that these two expressions are generated from the same
non-terminal. If enough traces of a whole group of expressions in
a whole group of contexts are found, the probability of this
hypothesis grows. In other words, grammar rules are only
introduced when enough evidence has been seen and thus only when
the probability of the hypothesis is high enough.

The difference with \abl's approach is that instead of inserting
hypotheses about constituents in the hypothesis space, the
unequal parts of the sentences are clustered (i.e.\ grouped) in
rewrite rules directly. However, \abl always stores the possible
constituents, whereas EMILE only induces grammar rules when
enough evidence has been found. EMILE never introduces
conflicting grammar rules; the grammar rules with the highest
probabilities are stored.

For a sentence of length $n$ the maximal number of different
contexts and expressions is $1/2 n(n+1)$.\footnote{Note that
EMILE's cluster routine does a more extensive search for patterns
than a k-gram routine that distinguishes only $n-(k-1)$ elements
in a sentence} The complexity of a routine that clusters all
contexts and expressions is polynomial in the number of contexts
and expressions.

The EMILE family of algorithms works efficiently for the class of
shallow context-free languages with characteristic contexts and
expressions provided that the sample is taken according to a
simple distribution~\citep{Adriaans:LSC-99}.  An
\index{Adriaans, P.W.}%
\emph{expression} of a type $T$ is \emph{characteristic} for $T$
\index{expression}%
\index{characteristic}%
if it only appears with contexts of type $T$. Similarly, a
\emph{context} of a type $T$ is \emph{characteristic} for $T$
\index{context}%
if it only appears with expressions of type $T$. In the
example one might see the context \sent{Oscar likes} as
characteristic for noun phrases and the phrases \sent{all
dustbins} and \sent{biscuits} as characteristic
for noun contexts. If more occurrences of the
characteristic contexts and types are found, the certainty that
these \emph{are} characteristic grows.

A distribution is simple if it is recursively enumerable. A class
of languages $C$ is \emph{shallow} if for each language $L$ it is
\index{shallow language}%
possible to find a grammar $G$, and a set of sentences $S$
inducing characteristic contexts and expressions for all the
types of $G$, such that the size of $S$ and the length of the
sentences of $S$ are logarithmic in the descriptive length of $L$
(relative to $C$).  Languages with characteristic contexts and
expressions for each syntactic type are called \emph{context- and
expression-separable}. A sample is \emph{characteristic} if it
\index{context separable}%
\index{expression separable}%
allows us to identify the right clusters that correspond with
non-terminals in the original grammar.

Samples generated by arbitrary probability distributions are very
likely to be non-characteristic. One can prove, however, that if
the sample is drawn according to a simple distribution and the
original grammar is shallow then the right clusters will be found
in a sample of polynomial size, i.e.\ one will have a
characteristic sample of polynomial size.

Natural languages seem to be context- and expression-separable
for the most part, i.e.\ if there are any types lacking
characteristic contexts or expressions,\footnote{After rewriting
types such as `verbs that are also nouns' as composites of basic
types.} these types are few in number, and rarely used.
Furthermore, there is no known example of a syntactic
construction in a natural language that cannot be expressed in a
short sentence. Hence the conjecture that natural languages are
(mostly) context- and expression-separable and shallow seems
tenable. This explains why EMILE works for natural language.

The EMILE 4.1 algorithm consists of two main stages:
\emph{clustering} and \emph{rule induction}.  In the clustering
phase all possible contexts and expressions of a sample are
gathered in a matrix. Starting from random seeds, clusters of
contexts and expressions, that form correct sentences, are
created.\footnote{A set of parameters and thresholds determines
the significance and the amount of noise in the clusters.} If
a group of contexts and expressions cluster together they receive
a type label. This creates a set of proto-rules. In the example,
\index{proto-rule}%
the proto-rules \nt{1} $\rightarrow$\sent{Oscar likes} \nt{2} and
\nt{2} $\rightarrow$ \sent{all dustbins} can be found (if there
is enough evidence for them). The sentence type \nt{1} can be
rewritten as \sent{Oscar likes} concatenated to an expression of
type \nt{2}.

A concise method for rule creation is used in the rule induction
phase.\footnote{EMILE 3.0 uses a much more elaborate, sound rule
induction algorithm, but it is impossible to implement this
routine efficiently.} In the rule induction phase, sentences in
\index{rule induction}%
the input are partially parsed using the set of proto-rules.
It introduces new grammar rules by applying the proto-rules.  New
rules are derived by substitution of types for characteristic
sub-expressions in typed expressions~\citep{Adriaans:LLC-92}.
\index{Adriaans, P.W.}%
Suppose for instance that the expression \sent{all dustbins} is
characteristic for type \nt{2}. It is then possible to form the
rule \sent{3} $\rightarrow$ \sent{cleans} \nt{2} \sent{with a
brush} from the rule \nt{3} $\rightarrow$ \sent{cleans all
dustbins with a brush}.

In \citep{Adriaans:LSC-99} it is shown that the EMILE 3.0
\index{Adriaans, P.W.}%
algorithm can PACS learn shallow context-free (or categorial)
languages with context- and expression separability in time
polynomial to the size of the grammar. EMILE 4.1 is an efficient
implementation of the main characteristics of 3.0.

%%%%%%%%%%%%%%%%%%%%%%%%%%%%%%%%%%%%%%%%%%%%%%%%%%%%%%%%%%%%%%%%%%%%%%%%%%%%%%%%
\thesissubsection{Theoretical comparison}
%%%%%%%%%%%%%%%%%%%%%%%%%%%%%%%%%%%%%%%%%%%%%%%%%%%%%%%%%%%%%%%%%%%%%%%%%%%%%%%%

While \abl directly (and greedily) structures sentences, EMILE
tries to find grammar rules in two steps. It first finds
proto-rules and using these proto-rules it introduces new grammar
rules.  Rules are only introduced when enough evidence has been
found.  This duality is actually the main difference of the two
systems.  Since \abl considers much more hypotheses, it results
in \abl being slower (i.e.\ taking more time and thus working on
smaller corpora) in contrast to EMILE, which is developed to work
on much larger corpora (say over 100,000 sentences).

The inner working of the algorithms is completely different.
EMILE finds a grammar rule when enough information is found to
support the rule. Evidence for grammar rules is found by
searching the matrix, which contains information on possible
contexts and expressions.  In other words, EMILE first finds a
set of proto rules.  The second phase uses these proto-rules to
search for occurrences of the right-hand side of a proto-rule in
the unstructured data, which indicates that the rule might have
been used to generate that sentence. This knowledge is used to
insert new grammar rules.

In contrast, \abl searches for unequal parts of sentences, since
these parts might have been generated from the same non-terminal
type (substitution class in EMILE's terminology). \abl remembers
\emph{all} possible substitution classes it finds and only when
all sentences have been considered and all hypotheses are found,
the ``best'' constituents are selected from the found hypotheses.

One more interesting feature worth mentioning is that EMILE (like
\abl) can learn recursive structures.
\index{recursive structure}%

%%%%%%%%%%%%%%%%%%%%%%%%%%%%%%%%%%%%%%%%%%%%%%%%%%%%%%%%%%%%%%%%%%%%%%%%%%%%%%%%
\thesissubsection{Numerical comparison}
%%%%%%%%%%%%%%%%%%%%%%%%%%%%%%%%%%%%%%%%%%%%%%%%%%%%%%%%%%%%%%%%%%%%%%%%%%%%%%%%

The two systems have been tested on two treebanks: the Air
Traffic Information System (ATIS) treebank and the Openbaar
\index{ATIS}%
\index{Air Traffic Information System}%
Vervoer Informatie Systeem (OVIS) treebank. Both treebanks have
\index{OVIS}%
\index{Openbaar Vervoer Informatie Systeem}%
been described in section~\vref{s:treebanks}. The only difference
here is that one-word sentences have been removed beforehand.
This explains the slightly different results of \abl in
table~\ref{tab:results1}.

The same evaluation approach as described in
chapter~\ref{ch:results} has been used. \abl and EMILE have both
been applied to the plain sentences of the two treebanks. The
structured sentences generated by \abl have been directly
compared against the structured sentences in the original corpus.
EMILE builds a context-free grammar. The plain sentences in the
original corpus are parsed using this grammar and the parsed
sentences are compared against the original tree structures.

\begin{tab}{Results of EMILE and \abl}
\label{tab:results1}
\vspace{1em}
\begin{tabular}
{|l|l||%
*{3}{rr|}%
}\hline
&&\multicolumn{2}{c|}{Recall}
&\multicolumn{2}{c|}{Precision}
&\multicolumn{2}{c|}{F-score}\\\hline\hline
ATIS & EMILE &
        16.81 & (0.69) & % Recall
        51.59 & (2.71) & % Precision
        25.35 & (1.00) \\ % F-Score
     & \abl   &
\textbf{25.77} & (0.22) & % Recall
\textbf{54.52} & (0.45) & % Precision
\textbf{35.00} & (0.29) \\ % F-Score
\hline\hline
OVIS & EMILE &
        36.89  & (0.77) & % Recall
        49.93  & (1.96) & % Precision
        41.43  & (3.21) \\ % F-Score
     & \abl   &
\textbf{53.59} & (0.07) & % Recall
\textbf{63.99} & (0.08) & % Precision
\textbf{58.33} & (0.06) \\ % F-Score
\hline
\end{tabular}
\end{tab}

An overview of the results of both systems on the two treebanks
can be found in table~\ref{tab:results1}. The figures in the
tables again represent the mean values of the metric followed by
their standard deviations (in brackets). Each result is computed
by applying the system ten times on the input corpus.

Note that the OVIS and ATIS corpora are certainly not
characteristic for the underlying grammars. It is therefore
impossible to learn a perfect grammar for these corpora from the
data in the corpora.\footnote{It is our hypothesis that one needs
a corpus of at least 50,000,000 sentences to get an acceptable
grammar of the English language on the basis of the EMILE
algorithm.}

As can be seen from the results, \abl outperforms EMILE on all
metrics on both treebanks. Since EMILE has been developed to work
on large corpora (much larger than the ATIS and OVIS treebanks),
the results are disappointing. However, it may well be the case
that EMILE will outperform the \abl system on such corpora.
\abl is a more greedy learner (it finds as many hypotheses as
possible and then disambiguates the hypothesis space), whereas
EMILE is much more cautious. Once a grammar rule has been
inserted in the grammar, it is considered correct.

Another explanation why \abl outperforms EMILE is that the EMILE
system has many parameters which influence for example the
greediness of the algorithm. By adjusting the parameters,
a different grammar may be found, which perhaps performs better
than this one. Several settings have been tried and the results
shown seem to be the best. This does not mean that there does not
exist a better setting  of the parameters.

Finally, the results of EMILE may be worse than those of \abl,
because the sentences had to be parsed with the grammar generated
by EMILE. A non-probabilistic parser is used to generate these
results. This may also explain the large standard deviations in
the results. If there are many derivations of the input
sentences, the system will select one at random. Since the
structured sentences are all parsed ten times, other derivations
may have been chosen, generating variable results.

%%%%%%%%%%%%%%%%%%%%%%%%%%%%%%%%%%%%%%%%%%%%%%%%%%%%%%%%%%%%%%%%%%%%%%%%%%%%%%%%
\thesissection{\abl in relation to the other systems}
%%%%%%%%%%%%%%%%%%%%%%%%%%%%%%%%%%%%%%%%%%%%%%%%%%%%%%%%%%%%%%%%%%%%%%%%%%%%%%%%

Now that several language learning systems have been described
and \abl is more closely compared against the EMILE system, this
section will relate \abl to established work in the field of grammar
induction. Each of the phases of \abl will be discussed briefly.

The two phases of \abl can roughly be compared to different
language learning methods. To our knowledge, no other language
learning system uses the edit distance algorithm to find possible
constituents. In this way, the alignment learning methods are
completely different from any other language learning technique,
but the idea behind alignment learning closely resembles that of
the first phase in the EMILE system. Both phases search for
substitutable subsentences.

Another way of looking at the alignment learning phase is that
the resulting (ambiguous) hypothesis space can be seen as a
collection of possible ways of compressing the input corpus. From
this point of view, the alignment learning phase resembles
systems that are based on the MDL principle.  The main difference
with \abl is that the hypothesis space contains a collection of
possible ways to compress the input corpus. Systems that use the
MDL principle usually find only the best compression (which does
not necessarily describe the best structure of the sentence).

The probabilistic selection learning methods are more closely
related to techniques used in other systems. The main difference
is that in \abl these techniques are used to disambiguate the
hypothesis space, where other systems use these techniques to
direct the learning system.  The probabilistic selection learning
instances select hypotheses based on the probability of the
possible constituents. A similar approach can be found in systems
that use distributional information to select the most probable
syntactic types such as the systems in \citep{Finch:SHOE92-230}
\index{Finch, S.}%
\index{Chater, N.}%
or \citep{Redington:98-22-425}. On the other hand, \abl assigns a
\index{Redington, M.}%
\index{Chater, N.}%
\index{Finch, S.}%
probability to the different hypotheses, which in a way is
similar to finding the best parse based on an SCFG
\citep{Baker:SCASA79-547}.
\index{Baker, J.K.}%

In the end, \abl can also be seen as a Bayesian learner. Using an
intermediate data structure (the hypothesis space), the system
finds the structured sentences such that:
\[
\arg\max_T \prod_{i=1}^n P(T_i|C_i)
\]
where $T$ is a list of $n$ trees with corresponding yields in $C$,
which is the list of sentences.

Finally, the \parseabl system has a grammar extraction phase
which makes use of the standard SCFG and STSG techniques.
\index{SCFG}%
\index{STSG}%
Extracting the grammars from the structured corpora generated by
the alignment and selection learning phases and also reparsing
the plain sentences is done using established methods.

%%%%%%%%%%%%%%%%%%%%%%%%%%%%%%%%%%%%%%%%%%%%%%%%%%%%%%%%%%%%%%%%%%%%%%%%%%%%%%%%
%%%%%%%%%%%%%%%%%%%%%%%%%%%%%%%%%%%%%%%%%%%%%%%%%%%%%%%%%%%%%%%%%%%%%%%%%%%%%%%%
\thesissection{Data-Oriented Parsing}
%%%%%%%%%%%%%%%%%%%%%%%%%%%%%%%%%%%%%%%%%%%%%%%%%%%%%%%%%%%%%%%%%%%%%%%%%%%%%%%%
%%%%%%%%%%%%%%%%%%%%%%%%%%%%%%%%%%%%%%%%%%%%%%%%%%%%%%%%%%%%%%%%%%%%%%%%%%%%%%%%
\label{s:DOP}
\index{Data-Oriented Parsing}%
\index{DOP}%

This section will relate \abl to the Data-Oriented Parsing (\dop)
framework \citep{Bod:ELS-95,Bod:BG-98}. Large parts of this
\index{Bod, R.}%
section can also be found in~\citep{vanZaanen:DOP02-??}.
\index{van Zaanen, M.M.}%

Data-Oriented Parsing and Alignment-Based Learning are two
completely different systems with different goals. The \dop
framework structures sentences based on known \emph{structured}
past experiences. \abl is a language learning system searching
for structure using \emph{unstructured} sentences only. However,
the global approach both choose to tackle their respective
problems is similar.

Both the \dop and \abl frameworks consist of two phases. The
first phase builds a search space of possible solutions and the
second phase searches this space to find the best solution.  In
the first phase, \dop considers all possible combinations of
subtrees in the tree grammar that lead to possible derivations of the
input sentence. The second phase then consists of finding the
best of these possibilities by computing the most probable parse,
effectively searching the ``derivation-space'', which contains
substructures of the sentences.

\abl has a similar setup. The first phase (alignment learning)
consists of building a search space of hypotheses by aligning
sentences to each other. The second phase (selection learning)
searches this space (using for example a statistical evaluation
function) to find the best set of hypotheses.

\abl and \dop are similar in remembering all possible solutions
in the search space for further processing later on.  The
advantage of proceeding in this way is that the final search
space contains more precise (statistical) information.  \abl and
\dop make definite choices using this more complete information,
in contrast to systems that choose between mutually exclusive
solutions at the time when they are found.

Note that \abl and \dop do not necessarily compute all solutions,
but all (or at least many) solutions are present in a compact
data structure. Taking into account all solutions is possible by
searching this data structure.

%%%%%%%%%%%%%%%%%%%%%%%%%%%%%%%%%%%%%%%%%%%%%%%%%%%%%%%%%%%%%%%%%%%%%%%%%%%%%%%%
\thesissubsection{Incorporating \abl in \dop}
%%%%%%%%%%%%%%%%%%%%%%%%%%%%%%%%%%%%%%%%%%%%%%%%%%%%%%%%%%%%%%%%%%%%%%%%%%%%%%%%
\label{s:ablindop}

Here we describe two extensions of the \dop system using \abl.
One way of extending \dop is to use \abl as a bootstrapping
method generating an initial tree grammar. The other way is to
have \abl running next to \dop to enhance \dop's robustness.

%%%%%%%%%%%%%%%%%%%%%%%%%%%%%%%%%%%%%%%%%%%%%%%%%%%%%%%%%%%%%%%%%%%%%%%%%%%%%%%%
\thesissubsubsection{Bootstrapping a tree grammar}
%%%%%%%%%%%%%%%%%%%%%%%%%%%%%%%%%%%%%%%%%%%%%%%%%%%%%%%%%%%%%%%%%%%%%%%%%%%%%%%%
\label{s:bootstrapping}

One of the main reasons for developing \abl was to allow a parser
to work on unstructured text without knowing the underlying
grammar beforehand. From this it follows directly that \abl can
be used to enhance \dop with a method to bootstrap an initial
tree grammar.  All \dop methods assume an initial tree grammar
containing subtrees.  These subtrees are normally extracted from
a structured corpus.  However, if no such corpus is available,
\dop cannot be used directly.

Normally, a structured corpus is built by hand. However, as
described in chapter~\ref{ch:learningalignment}, this is
expensive. For each language or domain, a new structured corpus
is needed and manually building such a corpus is not be feasible
due to time and cost restrictions.  Automatically building a
structured corpus circumvents these problems.  Unstructured
corpora are built more easily and applying an unsupervised
grammar bootstrapping system such as \abl to an unstructured
corpus is relatively fast and cheap. 

Figure~\ref{fig:abldop} gives an overview of the combined \abl
and \dop systems. The upper part describes how \abl is used to
build a structured corpus, while the lower part indicates how
\dop is used to parse a sentence.  Both systems are used as
usual, the figure merely illustrates how both systems can be
combined.  Starting out with an unstructured corpus, \abl
bootstraps a structured version of that corpus. From this,
subtrees are extracted using the regular method, which can then
be used for parsing. Note that subtrees extracted from the
sentences parsed by \dop can again be added to the tree grammar.

\begin{fig}{Using \abl to bootstrap a tree grammar for \dop}
\label{fig:abldop}
\vspace{1em}
\psset{framearc=.2,arrowscale=2}
\begin{psmatrix}[colsep=25pt]
&[name=plain]\psframebox[linestyle=dashed]{\parbox{2cm}{Plain\\corpus}}&
[name=abl]\psframebox{\parbox{2cm}{\abl\\}}&
[name=struc]\psframebox[linestyle=dashed]{\parbox{2cm}{Structured\\corpus}}\\
\empty[name=sent]\psframebox[linestyle=dashed]{\parbox{2cm}{Sentence\\}}&
[name=dop]\psframebox{\parbox{2cm}{\dop\\}}&
[name=treeb]\psframebox[linestyle=dashed]{\parbox{2cm}{Tree\\grammar}}&
[name=extract]\psframebox{\parbox{2cm}{Extract\\subtrees}}&
\psset{arrows=->}
\ncline{plain}{abl}
\ncline{abl}{struc}
\ncline{struc}{extract}
\ncline{extract}{treeb}
\ncline{treeb}{dop}
\ncline{sent}{dop}
\ncbar[angleA=270,angleB=270,arm=1cm]{dop}{extract}\naput{{Parsed sentence}}
\end{psmatrix}
\vspace{1cm}
\end{fig}

%%%%%%%%%%%%%%%%%%%%%%%%%%%%%%%%%%%%%%%%%%%%%%%%%%%%%%%%%%%%%%%%%%%%%%%%%%%%%%%%
\thesissubsubsection{Robustness of the parser}
%%%%%%%%%%%%%%%%%%%%%%%%%%%%%%%%%%%%%%%%%%%%%%%%%%%%%%%%%%%%%%%%%%%%%%%%%%%%%%%%
\label{s:robustness}

The standard \tdop model breaks down on sentences with unknown
words or unknown syntactic structures. One way of solving this
problem (in contrast to the \dop models described in
\citep{Bod:BG-98}) is to let \abl take care of these cases. The
\index{Bod, R.}%
main advantage is that even new syntactic structures (which were
not present in the original tree grammar) can be learned. To our
knowledge, no other \dop instantiation does this.

Figure~\ref{fig:robust} shows how, when \dop cannot parse a
sentence,\footnote{It is assumed that the unparsable sentence is
a correct sentence in the language and that the grammar (in the
form of \dop subtrees) is not complete.} the unparsable sentence
is passed to \abl. Applying \abl to the sentence results in a
structured version of the sentence. Since this structured
sentence resembles a parsed sentence, it can be the output of the
system. Another approach may be taken, where subtrees, extracted
from the sentence structured by \abl, are added to the tree
grammar and \dop will reparse the sentence (which will definitely
succeed).

\begin{fig}{Using \abl to adjust the tree grammar}
\label{fig:robust}
\vspace{1em}
\psset{framearc=.2,arrowscale=2}
\begin{psmatrix}[colsep=25pt]
&[name=unp]\psframebox[linestyle=dashed]{\parbox{2cm}{Unparsable\\sentence}}&
[name=abl]\psframebox{\parbox{2cm}{\abl\\}} &
[name=struc]\psframebox[linestyle=dashed]{\parbox{2cm}{Structured\\sentence}}\\
\empty
[name=sent]\psframebox[linestyle=dashed]{\parbox{2cm}{Sentence\\}}&
[name=dop]\psframebox{\parbox{2cm}{\dop\\}}&
[name=treeb]\psframebox[linestyle=dashed]{\parbox{2cm}{Tree\\grammar}}&
[name=extract]\psframebox{\parbox{2cm}{Extract\\subtrees}}
\psset{arrows=->}
\ncline{treeb}{dop}
\ncline[linestyle=dashed]{treeb}{abl}
\ncline{extract}{treeb}
\ncline{struc}{extract}
\ncline{abl}{struc}
\ncline{unp}{abl}
\ncline{dop}{unp}
\ncline{sent}{dop}
\ncbar[angleA=270,angleB=270,arm=1cm]{dop}{extract}\naput{{Parsed sentence}}
\end{psmatrix}
\vspace{1cm}
\end{fig}

The main problem with this approach is that \abl cannot learn
structure using the unparsable sentence only; \abl always needs
other sentences to align to.  Additionally, for any structure to
be found, \abl needs at least two sentences with at least one
word in common. There are several ways to solve this problem.

One way to find sentences for \abl to align is extracting them
from the tree grammar used by \dop. If complete tree structures
are present in the tree grammar, the yield of these trees (i.e.\
the plain sentences) can be used to align to the unparsable
sentence.  Furthermore, using the structure present in the trees
from the tree grammar, the correct type labels might be inserted
in the unparsable sentence.

If no completely lexicalised tree structures representing
sentences are present in the tree grammar (for example, because
they have been removed by pruning), sentences can still be found
by generating them from the subtrees. Using a smart generation
algorithm can assure that at least some words in the unparsable
sentence and the generated sentences are the same.

A completely different method of finding plain sentences which
\abl can use to align to is by running \abl parallel to
\dop. Each sentence \dop parses (correctly or incorrectly) is
also given to \abl. These sentences can then be used to learn
structure. When sentences are parsed with \dop, \abl builds its
own structured corpus which can be used to align unparsable
sentences to.

Additional information about the unparsable sentence can be
gathered when \abl initialises the structure of the unparsable
sentence with information from the (incomplete) chart \dop has
built. The incomplete chart contains structure based on the
subtrees in the treebank. These subtrees are a good indication of
part of the structure of the sentence even though the subtrees
cannot be combined into a complete derivation.

%%%%%%%%%%%%%%%%%%%%%%%%%%%%%%%%%%%%%%%%%%%%%%%%%%%%%%%%%%%%%%%%%%%%%%%%%%%%%%%%
\thesissubsection{Recursive definition}
%%%%%%%%%%%%%%%%%%%%%%%%%%%%%%%%%%%%%%%%%%%%%%%%%%%%%%%%%%%%%%%%%%%%%%%%%%%%%%%%

If \dop uses \abl (section~\ref{s:ablindop}) and \abl uses \dop
(section~\vref{s:extractstsg} and~\vref{s:selectionstsg}) at the
same time, there seems to be an infinite loop between the
systems, which is impossible to implement. However, when taking a
closer look, \dop is extended with \abl as a bootstrapping method
or to improve robustness. On the other hand, \abl is extended
with \dop as an improvement to the stochastic evaluation function
or to reparse sentences.  In the latter case, there is no need
for the robuster version of \dop. The \dop system that is used to
extend \abl is not the extended \dop system, so effectively there
is no recursive use between both systems.

%% end of file: compare.tex

%% file: extensions.tex
%%%%%%%%%%%%%%%%%%%%%%%%%%%%%%%%%%%%%%%%%%%%%%%%%%%%%%%%%%%%%%%%%%%%%%%%%%%%%%%%
%% Menno van Zaanen                                                           %%
%% menno@comp.leeds.ac.uk                                                     %%
%%%%%%%%%%%%%%%%%%%%%%%%%%%%%%%%%%%%%%%%%%%%%%%%%%%%%%%%%%%%%%%%%%%%%%%%%%%%%%%%
%% Filename: extensions.tex                                                   %%
%%%%%%%%%%%%%%%%%%%%%%%%%%%%%%%%%%%%%%%%%%%%%%%%%%%%%%%%%%%%%%%%%%%%%%%%%%%%%%%%

%%%%%%%%%%%%%%%%%%%%%%%%%%%%%%%%%%%%%%%%%%%%%%%%%%%%%%%%%%%%%%%%%%%%%%%%%%%%%%%%
%%%%%%%%%%%%%%%%%%%%%%%%%%%%%%%%%%%%%%%%%%%%%%%%%%%%%%%%%%%%%%%%%%%%%%%%%%%%%%%%
%%%%%%%%%%%%%%%%%%%%%%%%%%%%%%%%%%%%%%%%%%%%%%%%%%%%%%%%%%%%%%%%%%%%%%%%%%%%%%%%
\thesischaptertoc{Future Work:\\Extending the Framework}
{Future Work: Extending the Framework}
{Hmmm, especially enjoyed that one\ldots\\
 Let's see what's next\ldots\@}
{Offspring (Smash)}
\fancyhead[R]{\textsl{Extending the Framework}}
%%%%%%%%%%%%%%%%%%%%%%%%%%%%%%%%%%%%%%%%%%%%%%%%%%%%%%%%%%%%%%%%%%%%%%%%%%%%%%%%
%%%%%%%%%%%%%%%%%%%%%%%%%%%%%%%%%%%%%%%%%%%%%%%%%%%%%%%%%%%%%%%%%%%%%%%%%%%%%%%%
%%%%%%%%%%%%%%%%%%%%%%%%%%%%%%%%%%%%%%%%%%%%%%%%%%%%%%%%%%%%%%%%%%%%%%%%%%%%%%%%
\label{ch:extensions}

This chapter will describe several possible extensions of the
\index{extension}%
standard \abl system. At the moment, these extensions have not
yet been implemented.

First, three extensions of the alignment learning phase will be
described, followed by two possible extensions of the selection
learning phase. Next, two extensions that influence the entire
system are discussed, and finally something will be said about
applying the \abl system to other corpora.

%%%%%%%%%%%%%%%%%%%%%%%%%%%%%%%%%%%%%%%%%%%%%%%%%%%%%%%%%%%%%%%%%%%%%%%%%%%%%%%%
%%%%%%%%%%%%%%%%%%%%%%%%%%%%%%%%%%%%%%%%%%%%%%%%%%%%%%%%%%%%%%%%%%%%%%%%%%%%%%%%
\thesissection{Equal parts as hypotheses}
%%%%%%%%%%%%%%%%%%%%%%%%%%%%%%%%%%%%%%%%%%%%%%%%%%%%%%%%%%%%%%%%%%%%%%%%%%%%%%%%
%%%%%%%%%%%%%%%%%%%%%%%%%%%%%%%%%%%%%%%%%%%%%%%%%%%%%%%%%%%%%%%%%%%%%%%%%%%%%%%%
\index{hypothesis!equal part}%
\index{constituent!equal part}%

Even though the discussion in chapter~\ref{ch:learningalignment}
favoured assuming unequal parts of sentences as hypotheses, a
modified \abl system that, additionally, stores \emph{equal}
parts of sentences as hypothesis might yield better results.

The idea of the alignment learning phase is to find a good set of
hypotheses.  Ideally, the alignment learning phase inserts as
many correct and as few incorrect hypotheses into the
hypothesis universe as possible.  The selection learning phase,
which should select the best of these hypotheses, then has less
work in selecting the correct constituents and removing the
incorrect ones.

However, inserting too many hypotheses into the hypothesis
universe places a heavy burden on the selection learning phase,
since it will need to make a stricter selection based on a
hypothesis universe containing more noise. On the other hand,
since only the alignment learning phase inserts hypotheses into
the hypothesis universe, inserting too few hypotheses will
directly decrease the performance of the entire system.  There is
a trade-off between inserting more hypotheses (which implies that
many correct hypotheses are present in the hypothesis space, but
not all are correct) and inserting fewer hypotheses (where there
are fewer correct hypotheses inserted, but there is a larger
probability that the inserted hypotheses are correct).

Chapter~\ref{ch:learningalignment} showed that using equal parts
of sentences as hypotheses sometimes introduces correct
hypotheses, so adding these to the hypotheses universe may
increase the precision of the system in the end, while also
increasing the amount of work of the selection learning phase.

%%%%%%%%%%%%%%%%%%%%%%%%%%%%%%%%%%%%%%%%%%%%%%%%%%%%%%%%%%%%%%%%%%%%%%%%%%%%%%%%
%%%%%%%%%%%%%%%%%%%%%%%%%%%%%%%%%%%%%%%%%%%%%%%%%%%%%%%%%%%%%%%%%%%%%%%%%%%%%%%%
\thesissection{Weakening exact match}
%%%%%%%%%%%%%%%%%%%%%%%%%%%%%%%%%%%%%%%%%%%%%%%%%%%%%%%%%%%%%%%%%%%%%%%%%%%%%%%%
%%%%%%%%%%%%%%%%%%%%%%%%%%%%%%%%%%%%%%%%%%%%%%%%%%%%%%%%%%%%%%%%%%%%%%%%%%%%%%%%
\label{s:weakening}
\index{weakening exact match}%

The algorithms described so far are unable to learn any structure
when two sentences with completely distinct words are considered.
Since unequal parts of sentences are stored as hypotheses, only
the entire sentences (which have no words in common) are
hypotheses. In other words, for a hypothesis to be introduced,
there need to be equal words in the sentences. However, other
sentences in the corpus (which \emph{do} have words in common)
can be used to learn structure in the two distinct sentences.

Sometimes it is too strong a demand to require equal words in the
two sentences to find hypotheses; it is enough to have similar
words.  Imagine sentences~\ref{equivalent1}
and~\ref{equivalent2}, which are completely distinct. The
standard \abl learning methods would conclude that both are
sentences, but no more structure will be found. Now assume that
the algorithm knows that \textit{Book} and \textit{Show} are
words of the same type (representing verbs), it would find the
structures in sentences~\ref{resequivalent1}
and~\ref{resequivalent2}.

\eenumsentence{
\toplabel{equivalent}
\item \align{Book} a trip to Sesame Street
\label{equivalent1}
\item \align{Show} me Big Bird's house
\label{equivalent2}
}

\eenumsentence{
\toplabel{resequivalent}
\item \align{Book} {}[a trip to Sesame Street]\snt{1}
\label{resequivalent1}
\item \align{Show} {}[me Big Bird's house]\snt{1}
\label{resequivalent2}
}

An obvious way of implementing this is by using \emph{equivalence
classes} (for example the system as described in
\index{equivalence class}%
\citep{Redington:98-22-425}). Words that are closely related (in
\index{Redington, M.}%
\index{Chater, N.}%
\index{Finch, S.}%
a syntactic or semantic perspective) are grouped together in the
same class. Words that are in the same equivalence class are said
to be sufficiently equal, so the alignment algorithm may assume
they are equal and may thus link them. Sentences that do not have
words in common, but do have words in the same equivalence class
in common, can now be used to learn structure.

A great advantage of using equivalence classes is that they can be
learned in an unsupervised way. This means that when the
algorithm is extended with equivalence classes, it still does not
need to be initialised with structured training data.

Another way of looking at weakening the exact match is by
comparing it to the second phase of the EMILE system, the rule
induction. That phase introduces new grammar rules by applying
already known grammar rules to unstructured parts of sentences.
In other words, if there are grammar rules that rewrite
type~\nt{2} into \sent{Book} and into \sent{Show}, then the words
\sent{Book} and \sent{Show} are also possibly word groups of
type~\nt{2}, meaning that they are similar enough to be linked.
This again results in the sentences in~\ref{resequivalent}.

If equivalence classes or EMILE's rule induction phase are used
\index{rule induction}%
in the alignment learning phase, more hypotheses will be found
since more words in the sentences are seen as similar.  This
means that the selection learning phase of the algorithm has more
possible hypotheses to choose from.

%%%%%%%%%%%%%%%%%%%%%%%%%%%%%%%%%%%%%%%%%%%%%%%%%%%%%%%%%%%%%%%%%%%%%%%%%%%%%%%%
%%%%%%%%%%%%%%%%%%%%%%%%%%%%%%%%%%%%%%%%%%%%%%%%%%%%%%%%%%%%%%%%%%%%%%%%%%%%%%%%
\thesissection{Dual level constituents}
%%%%%%%%%%%%%%%%%%%%%%%%%%%%%%%%%%%%%%%%%%%%%%%%%%%%%%%%%%%%%%%%%%%%%%%%%%%%%%%%
%%%%%%%%%%%%%%%%%%%%%%%%%%%%%%%%%%%%%%%%%%%%%%%%%%%%%%%%%%%%%%%%%%%%%%%%%%%%%%%%
\label{s:duallevel}

In section~\vref{s:clustering} it was assumed that a hypothesis
in a certain context can only have one type. This assumption is
in line with Harris's procedure for finding substitutable
segments, but it introduces some problems.

Consider the sentences of~\ref{wellmeat} taken from the Penn
Treebank ATIS corpus.\footnote{A clearer example might be
\eenumsentence{
\toplabel{erniewellmeat}
\item \align{Ernie eats} biscuits
\item \align{Ernie eats} well
}}
When applying the \abl learning algorithm to these sentences, it
will determine that \sent{morning} and \sent{nonstop} are of the
same type, since they occur in the same context. However, in the
ATIS corpus, \sent{morning} is tagged as an \nt{NN} (a noun) and
\sent{nonstop} is a \nt{JJ} (an adjective).

\eenumsentence{
\toplabel{wellmeat}
\item \align{Show me the} [morning]\snt{1} \align{flights}
\item \align{Show me the} [nonstop]\snt{1} \align{flights}
}

On the other hand, one can argue that these words \emph{are} of
the same type, precisely because they occur in the same context.
Both words might be seen as some sort of modifying
phrase.\footnote{Although Harris's \emph{procedure} for finding
the substitutable segments breaks down in these cases, his
\emph{implication} does hold: \sent{nonstop} can be replaced by
for example \sent{cheap} (another adverb) and \sent{morning} can
be replaced by \sent{evening} (another noun).}

The assumption that word groups in the same context are always of
the same type is clearly not true. To solve this problem, merging
the types of hypotheses should be done more cautiously.

The example sentences of~\ref{wellmeat} show that there is a
difference between syntactic type and functional type of
constituents.  The words \sent{morning} and \sent{nonstop} have a
different syntactic type, a noun and an adjective respectively,
but both modify the noun \sent{flights}, i.e.\ they have the same
functional type. (The same applies for the sentences
in~\ref{erniewellmeat}.) \abl finds the functional type, while
the words are tagged according to their syntactic type, and thus
there is a discrepancy between the types.

The two different types are incorporated in the sentences as
shown in~\ref{wmsyntfunc}. Both hypotheses receive the \nt{1}
(functional) type because they occur in the same context. Each
hypothesis in the same context receives the same functional type.
The inner type (\nt{2} and \nt{3}) denotes the syntactic type of
the words. Hypotheses with the same yield always receive the same
syntactic type (which again is incorrect, but hopefully in the
end, this will even out).

\eenumsentence{
\toplabel{wmsyntfunc}
\item \align{Show me the} [[morning]\snt{2}]\snt{1} \align{flights}
\item \align{Show me the} [[nonstop]\snt{3}]\snt{1} \align{flights}
}

Since the overall use of the two words differs greatly, they occur in
different contexts. \sent{Morning} will in general occur in places
where nouns or noun phrases belong, while \sent{nonstop} will not.
This distinction can be used to differentiate between the two words.

When the alignment learning phase has finished, the merging of
syntactic and functional types is started. Only when the
distributions of two combination of a syntactic and a functional
type are similar enough (according to some criterion), they are
merged into the same (combined) type.  The distribution of
syntactic types within functional types can be used to find
combinations of syntactic and functional types that are similar
enough to be merged.  \sent{Morning} and \sent{nonstop} will then
receive different types, since the syntactic type of
\sent{morning} normally occurs within other functional types than
the syntactic type of \sent{nonstop}.

%%%%%%%%%%%%%%%%%%%%%%%%%%%%%%%%%%%%%%%%%%%%%%%%%%%%%%%%%%%%%%%%%%%%%%%%%%%%%%%%
%%%%%%%%%%%%%%%%%%%%%%%%%%%%%%%%%%%%%%%%%%%%%%%%%%%%%%%%%%%%%%%%%%%%%%%%%%%%%%%%
\thesissection{Alternative statistics in selection learning}
%%%%%%%%%%%%%%%%%%%%%%%%%%%%%%%%%%%%%%%%%%%%%%%%%%%%%%%%%%%%%%%%%%%%%%%%%%%%%%%%
%%%%%%%%%%%%%%%%%%%%%%%%%%%%%%%%%%%%%%%%%%%%%%%%%%%%%%%%%%%%%%%%%%%%%%%%%%%%%%%%
\label{s:parsingselectionlearning}
\index{selection learning!alternative statistics}%

Section~\vref{s:selectioninst} describes three different
selection learning methods. A probabilistic method performs best,
but those systems are very simple and make certain (incorrect)
assumptions.  This section will describe possible extensions or
improvements over the currently implemented systems. First, a
slightly modified probabilistic approach is discussed briefly,
followed by the description of an alternative approach based on
parsing.

%%%%%%%%%%%%%%%%%%%%%%%%%%%%%%%%%%%%%%%%%%%%%%%%%%%%%%%%%%%%%%%%%%%%%%%%%%%%%%%%
\thesissubsection{Smoothing}
%%%%%%%%%%%%%%%%%%%%%%%%%%%%%%%%%%%%%%%%%%%%%%%%%%%%%%%%%%%%%%%%%%%%%%%%%%%%%%%%
\index{selection learning!smoothing}%

One of the assumptions made when applying one of the probabilistic
selection learning methods is that the hypothesis universe contains all
possible hypotheses. In other words, it is assumed that the hypothesis
universe describes the complete population of hypotheses.

To loosen this assumption, the probabilities of the hypotheses
can be smoothed. Instead of assuming that the hypotheses in the
hypothesis universe are all existing hypotheses, it is seen as a
selection of the entire population. To account for this, a small
fraction of the probabilities of the hypotheses is shifted to the
unseen hypotheses.

There are several methods that can smooth the probabilities of the
hypotheses. \citet{Chen:ACL96-??,Chen:ESS-98} give a nice overview of
\index{Chen, S.F.}%
\index{Goodman, J.}%
the area of smoothing techniques.

%%%%%%%%%%%%%%%%%%%%%%%%%%%%%%%%%%%%%%%%%%%%%%%%%%%%%%%%%%%%%%%%%%%%%%%%%%%%%%%%
\thesissubsection{Selection learning through parsing}
%%%%%%%%%%%%%%%%%%%%%%%%%%%%%%%%%%%%%%%%%%%%%%%%%%%%%%%%%%%%%%%%%%%%%%%%%%%%%%%%
\index{selection learning!parsing}%

Instead of computing the probability of each possible combination
of (non-overlapping) constituents as described in
section~\vref{s:probsellearn}, it is also possible to
\emph{parse} the sentence with a grammar that is extracted from
the fuzzy trees, similar to the grammar extraction system as
described in section~\vref{s:grammarextract}. However, this
method is different from the \parseabl system, in that here the
parsing occurs in the selection learning phase. The grammar
extraction occurs directly after the alignment learning phase.

The main advantage of this selection learning method is that all
hypotheses are considered to be selected (or removed). This is in
contrast to the selection methods described earlier in this
thesis, where non-overlapping hypotheses are considered correct
and only overlapping hypotheses can be removed.

Since the final structure of the tree should be in a form that
could have been generated by a context-free grammar, the first
instantiation of selection learning by parsing extracts a
stochastic context-free grammar from the hypothesis space and
reparses the plain sentences using that grammar.  Similarly to
the two grammar extraction methods, another instantiation, using
the \dop system, which is based on the theory of stochastic tree
substitution grammars (STSG), will be discussed.

%%%%%%%%%%%%%%%%%%%%%%%%%%%%%%%%%%%%%%%%%%%%%%%%%%%%%%%%%%%%%%%%%%%%%%%%%%%%%%%%
\thesissubsubsection{Selection learning with an SCFG}
%%%%%%%%%%%%%%%%%%%%%%%%%%%%%%%%%%%%%%%%%%%%%%%%%%%%%%%%%%%%%%%%%%%%%%%%%%%%%%%%

The first system will extract an SCFG from the fuzzy trees
generated by the alignment learning phase. Using this grammar,
the plain sentences will be parsed and the resulting structure
will be the output of the selection learning phase.
Section~\vref{s:extractscfg} mentioned how to extract an SCFG
from a tree structure. However, the starting point for extracting
an SCFG in the grammar extraction phase is a tree structure, but
the starting point for selection learning is a fuzzy tree.  This
complicates things as shown in figure~\ref{fig:cfgfuzzy}.

\begin{fig}{Extracting an SCFG from a fuzzy tree}
\label{fig:cfgfuzzy}
[\snt{S}[\snt{X_1}[\snt{NP}Bert]\snt{NP}
[\snt{X_2}[\snt{V}sees]\snt{V}]\snt{X_1}
[\snt{NP}Ernie]\snt{NP}]\snt{X_2}]\snt{S}

\begin{multicols}{2}
\psset{nodesep=2pt,levelsep=35pt}
\pstree{\TR{\nt{S}}}
{\pstree{\TR{\nt{X_1}}}{\pstree{\TR{\nt{NP}}}{\TR{Bert}}
                        \pstree{\TR{\nt{V}}}{\TR{sees}}
                       }
 \pstree{\TR{\nt{NP}}}{\TR{Ernie}}
}

\pstree{\TR{\nt{S}}}
{
 \pstree{\TR{\nt{NP}}}{\TR{Bert}}
 \pstree{\TR{\nt{X_2}}}{\pstree{\TR{\nt{V}}}{\TR{sees}}
                        \pstree{\TR{\nt{NP}}}{\TR{Ernie}}
                       }
}
\end{multicols}

\begin{tabular}{l@{ \der\ }lr}
\nt{S}   & \nt{X_1} \nt{NP}  & 1/2=0.5 \\
\nt{S}   & \nt{NP} \nt{X_2}  & 1/2=0.5 \\
\nt{X_1} & \nt{NP} \nt{V}    & 1/1=1 \\
\nt{X_2} & \nt{V} \nt{NP}    & 1/1=1 \\
\nt{NP}  & Bert              & 1/2=0.5 or 2/4=0.5 \\
\nt{NP}  & Ernie             & 1/2=0.5 or 2/4=0.5 \\
\nt{V}   & sees              & 1/1=1 or 2/2=1
\end{tabular}

\end{fig}

The fuzzy tree, which is displayed as a structured sentence in the
figure mentioned above, denotes two (conflicting) tree structures
at the same time. When these two tree structures are extracted
and the regular grammar extraction method (as described in
section~\vref{s:extractscfg}) is used on these trees, then the
grammar shown in the figure is created. Using these grammar
rules, the tree structures encapsulated by the fuzzy tree can be
generated, so the structure in the fuzzy tree can be generated as
well.

The problem now is to compute the probabilities of the grammar
rules.  The first four grammar rules all occur once, but the next
two rules occur twice in the trees, so the probabilities are
$2/4=0.5$ (and $2/2=1$ for the last rule). However, in the
original fuzzy tree, the hypothesis only occurred once, so
actually the probabilities should be $1/2=0.5$ (and $1/1=1$ for
the final rule).

Since the final probabilities are the same in this case, it seems
as though there is no real problem. However, when for example the
grammar rule $\nt{NP} \rightarrow \sent{Bert}$ is found in
another fuzzy tree, the results will turn out differently. If the
\nt{NP} rules were counted four times, then the probability of
the new rule will be $3/5=0.6$, but if the original rules were
counted only two times, the probability should be $2/3=0.67$.

%%%%%%%%%%%%%%%%%%%%%%%%%%%%%%%%%%%%%%%%%%%%%%%%%%%%%%%%%%%%%%%%%%%%%%%%%%%%%%%%
\thesissubsubsection{Selection learning with an STSG}
%%%%%%%%%%%%%%%%%%%%%%%%%%%%%%%%%%%%%%%%%%%%%%%%%%%%%%%%%%%%%%%%%%%%%%%%%%%%%%%%
\label{s:selectionstsg}

Similarly to the approach taken in
section~\ref{s:grammarextract}, it is possible to extract a
stochastic tree substitution grammar instead of a stochastic
context free grammar to use for selection learning.  The
advantage of this type of grammar is that it can capture a wider
variety of stochastic dependencies between words in subtrees.

Extracting an SCFG from a fuzzy trees is not entirely without
problems, as shown in the previous section and extracting an STSG
has the same problem, only worse. The main problem with
extracting an SCFG is that grammar rules that are contained in
overlapping constituents can be counted in different ways. When
extracting subtrees, this occurs much more frequently.  On the
other hand, once this problem has been solved for the SCFG case,
it is also solved for the STSG case.

%%%%%%%%%%%%%%%%%%%%%%%%%%%%%%%%%%%%%%%%%%%%%%%%%%%%%%%%%%%%%%%%%%%%%%%%%%%%%%%%
%%%%%%%%%%%%%%%%%%%%%%%%%%%%%%%%%%%%%%%%%%%%%%%%%%%%%%%%%%%%%%%%%%%%%%%%%%%%%%%%
\thesissection{\abl with Data-Oriented Translation}
%%%%%%%%%%%%%%%%%%%%%%%%%%%%%%%%%%%%%%%%%%%%%%%%%%%%%%%%%%%%%%%%%%%%%%%%%%%%%%%%
%%%%%%%%%%%%%%%%%%%%%%%%%%%%%%%%%%%%%%%%%%%%%%%%%%%%%%%%%%%%%%%%%%%%%%%%%%%%%%%%
\index{DOT}%
\index{Data-Oriented Translation}%

Recently there has been research into a data-oriented approach to
machine translation
\citep{Poutsma:COLING00-635,Poutsma:DOT-00,Way:99-11-??}.  The
\index{Poutsma, A.}%
\index{Way, A.}%
idea of the systems (based on \dop) is to translate sentences by
parsing the source sentence using elementary trees extracted from
a bilingually annotated, paired treebank. Each of the elementary
\index{treebank!bilingual, paired}%
\index{structured corpus!bilingual, paired}%
subtrees in the source language is linked to a subtree in the
target language, so when a parse of the source language is found,
the parse and thus the translated sentence in the target language
follows automatically.

One of the main problems with \dop is that an initial set of
elementary subtrees is needed. With these machine translation
techniques, the problem is even worse, since
\emph{linked}\footnote{Subtrees are linked when they are
translation equivalent.} (bi-lingual) subtrees are needed.
Building such treebanks by hand is highly impractical. However,
in section~\vref{s:extractstsg}, it was shown that \abl can learn
a stochastic tree substitution grammar. This type of grammar is
the basis of the \dop framework. By adapting \abl slightly, it
can also be used to learn linked STSGs.

Instead of applying \abl to a set of sentences, \abl is given a
set of pairs of sentences, where one sentence of the pair
translates to the other sentence. If for example the sentences in
figure~\ref{fig:learnsentencepair} are given to \abl (the English
sentences on the left-hand side translate to the Dutch sentences
on the right-hand side), the hypotheses with types \nt{E_2} and
\nt{D_2} are introduced.

\begin{fig}{Learning structure in sentence pairs}
\label{fig:learnsentencepair}
\vspace{1em}
\begin{tabular}{l|l}
\multicolumn{1}{c|}{English}
  & \multicolumn{1}{c}{Dutch} \\\hline
{}[\align{Bert sees} [Ernie]\snt{E_2}]\snt{E_1}
  & {}[\align{Bert ziet} [Ernie]\snt{D_2}]\snt{D_1}\\
{}[\align{Bert sees} [Big Bird]\snt{E_2}]\snt{E_1} &
  {}[\align{Bert ziet} [Pino]\snt{D_2}]\snt{D_1}
\end{tabular}
\end{fig}

When a hypothesis is introduced in the source and target
sentences, a hypothetical link is also introduced between the
two. A link indicates that the two phrases below the linked nodes
are translation equivalent. Figure~\ref{fig:linkedtrees} shows
how the sentences of figure~\ref{fig:learnsentencepair} are
linked. The link between \nt{E_1} and \nt{D_1} indicates that the
two sentences are translation equivalent and the same applies to
the phrases below the \nt{E_2} and \nt{D_2} non-terminals.

\begin{fig}{Linked tree structures}
\label{fig:linkedtrees}
\vspace{1cm}
\pstree{\TR{\node{E1}{E_1}}}
{\TR{Bert}
 \TR{sees}
 \pstree{\TR{\node{E2}{E_2}}}
 {\TR{Ernie}
 }
}
\hspace{1cm}
\pstree{\TR{\node{D1}{D_1}}}
{\TR{Bert}
 \TR{ziet}
 \pstree{\TR{\node{D2}{D_2}}}
 {\TR{Ernie}
 }
}
\nlink{E1}{D1}
\nlink{E2}{D2}
\\
\vspace{1cm}
\pstree{\TR{\node{E1}{E_1}}}
{\TR{Bert}
 \TR{sees}
 \pstree{\TR{\node{E2}{E_2}}}
 {\TR{Big}
  \TR{Bird}
 }
}
\hspace{1cm}
\pstree{\TR{\node{D1}{D_1}}}
{\TR{Bert}
 \TR{ziet}
 \pstree{\TR{\node{D2}{D_2}}}
 {\TR{Pino}
 }
}
\nlink{E1}{D1}
\nlink{E2}{D2}
\end{fig}

When more than one pair of hypotheses is found in the sentences, all
possible combinations of hypothetical links are introduced. This
is needed since different languages may have a different word
order. For example the sentences as given in~\ref{likesplait}
(taken from~\citep{Poutsma:DOT-00}) show that the subject of the
\index{Poutsma, A.}%
first sentence occurs as the object of the second sentence. If
these sentences are aligned against sentences where the subjects
\emph{and} objects are different, it is unclear whether the
hypothesis containing \sent{John} should be linked against the
hypothesis \sent{Jean} or \sent{Marie}, since both sentences will
receive two hypotheses.

\eenumsentence{
\toplabel{likesplait}
\item John likes Mary
\item Marie pla\^\i t \`a Jean
}

After the alignment learning phase, the hypotheses need to be
disambiguated as usual, but since the system also introduces
ambiguous links, these need to be disambiguated as well.  The
best links can be chosen by computing the probability that a
hypothesis in one language is linked against the hypothesis in
the other language. In the end, links that occur more often will
be correct. For example, \sent{John} will be more often linked
with \sent{Jean} than with \sent{Marie}, when other sentences
containing \sent{John} and \sent{Jean} but not \sent{Marie} occur
in the corpus.

%%%%%%%%%%%%%%%%%%%%%%%%%%%%%%%%%%%%%%%%%%%%%%%%%%%%%%%%%%%%%%%%%%%%%%%%%%%%%%%%
%%%%%%%%%%%%%%%%%%%%%%%%%%%%%%%%%%%%%%%%%%%%%%%%%%%%%%%%%%%%%%%%%%%%%%%%%%%%%%%%
\thesissection{(Semi-)Supervised Alignment-Based Learning}
%%%%%%%%%%%%%%%%%%%%%%%%%%%%%%%%%%%%%%%%%%%%%%%%%%%%%%%%%%%%%%%%%%%%%%%%%%%%%%%%
%%%%%%%%%%%%%%%%%%%%%%%%%%%%%%%%%%%%%%%%%%%%%%%%%%%%%%%%%%%%%%%%%%%%%%%%%%%%%%%%
\index{SABL}%
\index{Supervised Alignment-Based Learning}%

Unsupervised grammar induction systems like \abl do not have any
knowledge about what the final treebank should look like, since
unsupervised systems are not guided towards the \emph{wanted}
structure. Although, they usually yield less than perfect results,
these systems are still useful, for example when building a
treebank of an unknown language, when no experts are available or
\index{treebank!build}%
\index{structured corpus!build}%
when results are needed quickly.

On the other hand, when experts \emph{are} available, when more
precise results are needed or when there are no pressing time
restrictions, the resulting treebank generated by an unsupervised
grammar induction system might not be satisfactory. One possible
way to use an unsupervised induction system for the generation of
high quality treebanks is to improve the quality of the generated
treebank by post-processing done by experts. As an example, the
Penn Treebank, which is probably the most widely used treebank to
date, was annotated in two phases (automatic structure induction
followed by manual post-processing) \citep{Marcus:93-19-313}.
\index{Marcus, M.}%
\index{Santorini, B.}%
\index{Marcinkiewicz, M.}%

A more interesting approach to building a structured corpus,
instead of choosing for one of the two possibilities of building
a treebank (using an induction system or annotating the treebank
\index{treebank!build}%
\index{structured corpus!build}%
by hand), would be to combine the best of both methods. This
should result in a system that suggests possible tree structures
for the expert to choose from \emph{and} it should learn from the
choices made by the expert in parallel.

The \abl system can be adapted into a system, called (Semi-)
Supervised \abl (\sabl), that indicates reasonably good tree
structures \emph{and} learns from the expert's choices. The only
changes needed in the algorithm occur in the selection learning
phase:
\begin{description}
\item [Select n-best hypotheses] Instead of selecting the best
hypotheses only, as in standard \abl, let the system select the n
(say 5) best hypotheses. These hypotheses are presented to the
expert, who chooses the correct one or, if the correct one is not
present in the set of n-best hypotheses, adds it manually.
\item [Learn from the expert's choice]
\abl's selection of the best hypotheses from the hypothesis universe
is guided by a probabilistic evaluation function. In order to learn
from the choices made by the expert, the probabilities of the chosen
hypotheses should be changed:
\begin{itemize}
\item If the correct hypothesis (i.e.\ the hypothesis chosen by the
expert) was already present in the hypothesis universe, the
probability of that hypothesis should be increased. When a
hypothesis has a high probability, it will have a higher chance
to be selected next time.
\item If the correct hypothesis was \emph{not} present in the
hypothesis universe, it should be inserted and the probabilities of
the hypotheses should be adjusted. Since it was the preferred
hypothesis, its probability should be increased as if the hypothesis
was present in the hypothesis universe already.
\end{itemize}
\end{description}

Varying the amount of increase in probability, changes the
learning properties of the system. A small amount of increase
makes the system a slow learner, while a large amount of increase
may over-fit the system.

Using an unsupervised grammar induction system and manual
annotation are two opposing methods in building a treebank. 
\index{treebank!build}%
\index{structured corpus!build}%
\sabl can be placed anywhere between the two extremes. By varying
how many of the proposed hypotheses are actually used, \sabl can
be adjusted to work anywhere in the continuum between hand
annotation and fully automatic structuring of sentences.

%%%%%%%%%%%%%%%%%%%%%%%%%%%%%%%%%%%%%%%%%%%%%%%%%%%%%%%%%%%%%%%%%%%%%%%%%%%%%%%%
%%%%%%%%%%%%%%%%%%%%%%%%%%%%%%%%%%%%%%%%%%%%%%%%%%%%%%%%%%%%%%%%%%%%%%%%%%%%%%%%
\thesissection{More corpora}
%%%%%%%%%%%%%%%%%%%%%%%%%%%%%%%%%%%%%%%%%%%%%%%%%%%%%%%%%%%%%%%%%%%%%%%%%%%%%%%%
%%%%%%%%%%%%%%%%%%%%%%%%%%%%%%%%%%%%%%%%%%%%%%%%%%%%%%%%%%%%%%%%%%%%%%%%%%%%%%%%
\index{corpus!other}%

Along with extending the \abl system, more extensive testing of
the current system is needed as well. Testing \abl on different
corpora will yield a deeper insight into the properties and
(possible) shortcomings of \abl.  Future research can take
several different directions when evaluating \abl. Interesting
future work will investigate
\begin{itemize}
\item the linguistic properties of \abl,
\item the performance of \abl in a larger domain,
\item the application of \abl on completely different data sets.
\end{itemize}

Chapter~\vref{ch:results} showed that \abl performs reasonably
well on corpora of mainly right branching languages. It clearly
outperformed the system that generates a randomly chosen left or
right branching structure.

It was claimed that \abl will perform equally well on corpora of
left or right branching languages. This claim needs to be tested on
corpora of for example Japanese. Since the \abl system does not
have a built-in preference for left or right branching
structures, it can be expected that \abl will perform equally
well on a corpus of a left or right branching language.

A right branching system outperforms \abl on an English or Dutch
corpus and a left branching system will probably outperform \abl
on a Japanese corpus. However, this is an unfair comparison,
since the left and right branching systems are biased towards the
language in the corpus (whereas \abl is not). The independent
random system as described in chapter~\ref{ch:results} can, like
\abl, be expected to perform similarly on a Japanese corpus.
Therefore, \abl will probably outperform the random system on a
corpus of a left branching language, too.

\abl has been tested on two corpora which both are taken from a
limited domain (that of flight information and public
transportation). Next to these two corpora, the \abl system has
been applied to the large domain WSJ corpus. This showed that it
\emph{is} practically possible to use this system on such a
corpus.

First tests on a larger domain corpus show the need for loosening
the exact match of words, as discussed in
section~\ref{s:weakening}. Since the size of the vocabulary in a
large domain is larger, it will help the system match
non-function words. This indicates that more research can be done
in this direction.

Finally, the system can also be applied to different types of
corpora. Although the original system is developed to find
syntactic structure in natural language sentences, it might be
interesting to see how well \abl can find structure in other
types of data, which can be (but are not limited to) for example:
\begin{description}
\item [Morphology]  There has been some work in the unsupervised
\index{corpus!morphology}%
learning of morphological structure, e.g.\
\citet{Clark:ACL/Student01-55,Gaussier:ULNLP99-??,Goldsmith:27-2-153}.
\index{Clark, A.}%
\index{Gaussier, E.}%
\index{Goldsmith, J.}%
It is also possible to apply \abl to a set of words, instead of a
set of plain sentences, to find inner-word structure.  \abl will
then align characters (phonemes, or plain letters) while the rest
of the system remains the same.

As an example, consider aligning the words \textipa{/r2nIN/}
(running) and \textipa{/wO:kIN/} (walking) as
shown in~\ref{morph} (taken from~\citep{Longman:LDCE-95}). From
this alignment, the syllables \textipa{/r2n/} (run),
\textipa{/wO:k/} (walk), and \textipa{/IN/} (-ing) can be found.
Similarly, applying the algorithm to compound words will
decompose these. When applying \abl to a collection of words, a
hierarchically deeper structure can be found.

\eenumsentence{
\toplabel{morph}
\item \textipa{r2n\align{IN}}\\
running
\item \textipa{wO:k\align{IN}}\\
walking
}

\item [Music] A central topic in musicology is to construct formal
\index{corpus!music}%
descriptions of musical structures. Where in linguistics it is
uncontroversial to use tree structures to describe the syntactic
structure of sentences, in musicology similar
structures\footnote{Recently, structured musical data has become
available \citep{Schaffrath:EFC-95}.} are used
\index{Schaffrath, H.}%
\citep{Sloboda:MM-85}.
\index{Sloboda, J.A.}%

Musical pieces can be structured according to different
viewpoints. \citet{Lerdahl:GTTM-83} recognise the following
\index{Lerdahl, F.}%
\index{Jackendoff, R.S.}%
components on which a musical piece can be structured:
\begin{description}
\item [Grouping structure] This component indicates how a musical
piece can be subdivided into sections, phrases and motives.
\item [Metrical structure] A musical piece contains strong and
weak beats, which are often repeated in a regular way within a
number of hierarchical levels. This component structures a
musical piece based on the metric structure within that piece.
\item [Time-span reduction] A musical piece can be structured
based on the pitches.  Each of the pitches can be placed in a
hierarchy of structural importance based on their position in
metrical and grouping structure.
\item [Prolongational reduction] The pitches in a musical piece
can also be structured in a hierarchy that ``expresses harmonic
and melodic tension and relaxation, continuity and progression.''
\end{description}
Next to these viewpoints, there are other dimensions of musical
structure, such as timbre, dynamics, and motivic-thematic
processes. However, these are not hierarchical in nature.

A good system that learns structure in music will need to be able
to recognise (the combinations of) these different viewpoints
with their corresponding parameters. If \abl is to be used in
this field, several adaptions are needed. Furthermore, musical
pieces are much longer than natural language sentences. Since the
\abl system relies on the edit distance algorithm, it has
difficulties with longer input.

\item [DNA structure] DNA (or RNA) is usually described by naming
\index{corpus!DNA structure}%
the bases in a DNA strand. Bases are the building blocks of DNA
and each base is denoted by a letter. Each base is one of
\sent{A}, \sent{G}, \sent{C}, or \sent{T}. A piece of DNA is thus
described by a list containing these four letters. In the end,
the parts of a DNA string of bases denoted by letters can be
combined to form larger molecules.

From the DNA molecules, RNA is extracted. RNA is a molecule that
is composed of (copies of) parts of the original DNA molecule.
The combinations of adjacent bases in the RNA can be seen as
blueprints for amino acids. These amino acids can then form
proteins. The main problem here is to find which parts of the DNA
molecules contain useful information and will be copied into an
RNA strand
\citep{Durbin:ACL01-1,Gusfield:AOSTAS-97,Sankoff:TWS-99,CLBS94}.
\index{Durbin, R.}%
\index{Gusfield, D.}%
\index{Sankoff, D.}%
\index{Kruskal, J.}%

Figure~\ref{fig:structuredna} illustrates how RNA bases combine
into amino acids, which again combine into a protein. This
hierarchy might be found when applying \abl to the bases of RNA.

\end{description}

\begin{fig}{Structure in RNA}
\label{fig:structuredna}
\vspace{1em}
\psset{nodesep=2pt,treesep=10pt,levelsep=35pt}
\pstree{\TR{Protein}}
{\pstree{\TR{Amino}}{\TR{Base}
                     \TR{Base}
                     \TR{Base}
                    }
 \pstree{\TR{Amino}}{\TR{Base}
                     \TR{Base}
                     \TR{Base}
                    }
 \pstree{\TR{Amino}}{\TR{Base}
                     \TR{Base}
                     \TR{Base}
                    }
 \pstree{\TR{Amino}}{\TR{Base}
                     \TR{Base}
                     \TR{Base}
                    }
}
\end{fig}

The main problem with these types of data (e.g.\ musical and
DNA/RNA) is that there are no clear ``sentences''. Whereas
musical data might be chunked into phrases, this is clearly more
difficult with DNA or RNA information. The current implementation
of \abl is inherently slow when applied to very large strings (of
for example over 10,000 symbols), since the time of the edit
distance algorithm is in the order of the squared length of the
string.

\clearpage

%%% NOTE!!!! hack the fancyheaders...
\fancyhead[R]{\textsl{\leftmark}}
%%% NOTE!!!! hack the fancyheaders...

%% end of file: extensions.tex

%% file: conclusion.tex
%%%%%%%%%%%%%%%%%%%%%%%%%%%%%%%%%%%%%%%%%%%%%%%%%%%%%%%%%%%%%%%%%%%%%%%%%%%%%%%%
%% Menno van Zaanen                                                           %%
%% menno@comp.leeds.ac.uk                                                     %%
%%%%%%%%%%%%%%%%%%%%%%%%%%%%%%%%%%%%%%%%%%%%%%%%%%%%%%%%%%%%%%%%%%%%%%%%%%%%%%%%
%% Filename: conclusion.tex                                                   %%
%%%%%%%%%%%%%%%%%%%%%%%%%%%%%%%%%%%%%%%%%%%%%%%%%%%%%%%%%%%%%%%%%%%%%%%%%%%%%%%%

%%%%%%%%%%%%%%%%%%%%%%%%%%%%%%%%%%%%%%%%%%%%%%%%%%%%%%%%%%%%%%%%%%%%%%%%%%%%%%%%
%%%%%%%%%%%%%%%%%%%%%%%%%%%%%%%%%%%%%%%%%%%%%%%%%%%%%%%%%%%%%%%%%%%%%%%%%%%%%%%%
%%%%%%%%%%%%%%%%%%%%%%%%%%%%%%%%%%%%%%%%%%%%%%%%%%%%%%%%%%%%%%%%%%%%%%%%%%%%%%%%
\thesischapter{Conclusion}
{De antwoorden zijn altijd al aanwezig.}
{Steve Vai (Passion and Warfare)}
%%%%%%%%%%%%%%%%%%%%%%%%%%%%%%%%%%%%%%%%%%%%%%%%%%%%%%%%%%%%%%%%%%%%%%%%%%%%%%%%
%%%%%%%%%%%%%%%%%%%%%%%%%%%%%%%%%%%%%%%%%%%%%%%%%%%%%%%%%%%%%%%%%%%%%%%%%%%%%%%%
%%%%%%%%%%%%%%%%%%%%%%%%%%%%%%%%%%%%%%%%%%%%%%%%%%%%%%%%%%%%%%%%%%%%%%%%%%%%%%%%
\label{ch:conclusion}

Alignment-Based Learning (\abl) is an unsupervised grammar
induction system that generates a labelled, bracketed version of
the unstructured input corpus. The goal of the system is to learn
syntactic structure in plain sentences using a minimum of
information. This implies that no \textit{a priori} knowledge of
the language of the input corpus (or of any other particular
language in general) is assumed, not even part-of-speech tags of
the words. This shows how an empiricist system can learn
syntactic structure in practice.

The system is a combination of several known techniques, which
are used in completely new ways. It relies heavily on
\citeauthor{Harris:SL-51}'s notion of substitutability and
\citeauthor{Wagner:74-21-168}'s edit distance algorithm.
Implementing a system based on the notion of substitutability
using the edit distance algorithm yields several new insights
into these established methods.

\citet{Harris:SL-51} states that substitutable segments are
\index{Harris, Z.S.}%
descriptively equivalent. This means that everything that can be
said about one segment can also be said about the other segment.
He also describes a method to find the substitutable segments by
comparing sentences: substitutable segments are parts of
utterances that can be substituted for each other in different
contexts. This idea is used as a starting point for the alignment
learning phase.

Harris never mentioned the practical problems with this approach.
The first problem to tackle is how to find the substitutable
parts of two sentences.  In \abl, the edit distance algorithm is
used for this task. This is reflected in the alignment learning
phase of the system.

The second problem of \citeauthor{Harris:SL-51}'s notion of
substitutability is that when searching for substitutable
segments, at some point conflicting (overlapping) substitutable
segments are found. The selection learning phase of \abl tries to
disambiguate between the possible (context-free) syntactic
structures found by the alignment learning phase.

\abl consists thus of two phases, alignment learning and
selection learning. The first phase generates a search space of
possible constituents and the second phase searches this space to
select the best constituents. The selected constituents are
stored in the structured output corpus.

During the alignment learning phase, pairs of sentences are
aligned against each other. This can be done in several different
ways. In this thesis three systems are described. The first uses
the instantiation of the edit distance algorithm
\citep{Wagner:74-21-168} which finds the longest common
\index{Wagner, R.A.}%
\index{Fischer, M.J.}%
subsequence between two sentences. The second is an adjusted
version of the longest common subsequence algorithm which prefers
not to link equal words in the two sentences that are relatively
far apart (i.e.\ one word in the beginning of the sentence and the
other word at the end of the other sentence).  The third system,
which does not use the edit distance algorithm, finds all
possible alignments if there are more than one.

Aligning sentences against each other uncovers parts of the
sentences that are equal (or unequal) in both. The unequal parts
of the sentences are stored as hypotheses, denoting possible
constituents. This step is in line with
\citeauthor{Harris:SL-51}'s notion of substitutability, where
parts of sentences that occur in the same context are recognised
as constituents.

The alignment learning phase may at some point introduce
hypotheses that overlap each other. Overlapping hypotheses are
unwanted, since their structure could never have been generated
by a context-free (or mildly context-sensitive) grammar.  The
selection learning phase selects hypotheses from the set of
hypothesis generated by the alignment learning phase, which
resolves all overlapping hypotheses. This can be done in several
different ways.

First, a non-probabilistic method is described where hypotheses
that are learned earlier are considered correct. The main
disadvantage of this system is that incorrectly learned
hypotheses can never be corrected, when they are learned early.
The other systems that have been implemented both have a
probabilistic basis. The probability of each (overlapping)
hypothesis is computed using counts from the hypotheses in the
set of all hypotheses generated by the alignment learning phase.
The probabilities of the separate hypotheses are combined and the
combination of (non-overlapping) hypotheses with the highest
combined probability is selected.

The probabilistic selection learning methods allow for a gradient
range of knowledge about hypotheses, instead of an absolute
yes/no distinction. This (potentially) solves many of the
problems that used to be attributed to Harris's notion of
substitutability (e.g.\ the problems introduced by
\citet{Chomsky:LSLT-75} and \citet{Pinker:LI-94} as discussed in
\index{Chomsky, N.}%
\index{Pinker, S.}%
section~\ref{s:criticism}).

The \abl system, consisting of the alignment and selection
learning phase, can be extended with a grammar extraction and
parsing phase. This system is called \parseabl. The output of
this system is a structured corpus (similar to the \abl system)
and a stochastic grammar (a context-free or tree substitution
grammar). Reparsing the plain sentences using an extracted
grammar does not improve the quality resulting structured corpus,
however.

No language \emph{dependent} assumptions are considered by the
system, however, it relies on some language \emph{independent}
assumptions.  First of all, Harris's idea of substitutability
gives a way of finding possible constituents. This results in a
richly structured version of the input sentences. For evaluation
purposes, an underlying context-free grammar constraint is
imposed on this data structure. Note that this is not a necessary
assumption for the system. It is not a feature of the system.

When applying the system to real-life data sets, some striking
features of the system arise.  The structured corpora generated
by \abl by applying it to the ATIS, OVIS and WSJ corpora all
contain recursive structures. This is interesting since \abl is
able to find recursive constituents by considering only a finite
number of sentences.

The \abl system has been applied to three corpora, the ATIS, OVIS
and WSJ corpus. On all corpora, \abl yielded encouraging results.
To our knowledge, this is the first time an unsupervised learning
system has been applied to the plain WSJ corpus. Additionally,
\abl has been compared to the EMILE system
\citep{Adriaans:LLC-92}, which it clearly outperforms. More
\index{Adriaans, P.W.}%
recently, \citet{Clark:ACL/CNLL01-105} has compared his system
\index{Clark, A.}%
against \abl. These results are not completely comparable, since
his system is bootstrapped using much more information. Even
then, \abl's results are competitive.

%% end of file: conclusion.tex

%% file: thesis.bbl
\begin{thebibliography}{}

\bibitem[{ACL}, 1995]{ACL95}
{ACL} (1995).
\newblock {\em Proceedings of the 33th Annual Meeting of the {}Association for
  Computational Linguistics ({ACL})}. Association for Computational Linguistics
  ({ACL}).

\bibitem[{ACL}, 1997]{ACL97}
{ACL} (1997).
\newblock {\em Proceedings of the 35th Annual Meeting of the {}Association for
  Computational Linguistics ({ACL}){} and the 8th Meeting of the {}European
  Chapter of the Association for Computational Linguistics ({EACL}){}; Madrid,
  Spain}. Association for Computational Linguistics ({ACL}).

\bibitem[Adriaans et~al., 2000]{Adriaans:SOFSEM00-173}
Adriaans, P., Trautwein, M., and Vervoort, M. (2000).
\newblock Towards high speed grammar induction on large text corpora.
\newblock In Hlav\'{a}\v{c}, V.~Feffrey, G. and Wiedermann, J., editors, {\em
  {SOFSEM} 2000: Theory and Practice of Informatics}, volume 1963 of {\em
  Lecture Notes in Computer Science}, pages 173--186. Springer-Verlag, Berlin
  Heidelberg, Germany.

\bibitem[Adriaans, 1992]{Adriaans:LLC-92}
Adriaans, P.~W. (1992).
\newblock {\em Language Learning from a Categorial Perspective}.
\newblock PhD thesis, University of Amsterdam, Amsterdam, the Netherlands.

\bibitem[Adriaans, 1999]{Adriaans:LSC-99}
Adriaans, P.~W. (1999).
\newblock Learning shallow context-free languages under simple distributions.
\newblock Technical Report PP-1999-13, Institute for Logic, Language and
  Computation ({ILLC}), Amsterdam, the Netherlands.

\bibitem[Allen, 1995]{Allen:NLU-95}
Allen, J. (1995).
\newblock {\em Natural Language Understanding}.
\newblock Benjamin/Cummings, Redwood City:CA, USA, 2nd edition.

\bibitem[Baker, 1979]{Baker:SCASA79-547}
Baker, J. (1979).
\newblock Trainable grammars for speech recognition.
\newblock In Wolf, J. and Klatt, D., editors, {\em Speech Communication Papers
  for the Ninety-seventh Meeting of the Acoustical Society of America}, pages
  547--550.

\bibitem[Bellman, 1957]{Bellman:DP-57}
Bellman, R.~E. (1957).
\newblock {\em Dynamic Programming}.
\newblock Princeton University Press, Princeton:NJ, USA.

\bibitem[Black et~al., 1991]{Black:SNL91-306}
Black, E., Abney, S., Flickinger, D., Gdaniec, C., Grishman, R., Harrison, P.,
  Hindle, D., Ingria, R., Jelinek, F., Klavans, J., Liberman, M., Marcus, M.,
  Roukos, S., Santorini, B., and Strzalkowski, T. (1991).
\newblock A procedure for quantitatively comparing the syntactic coverage of
  {E}nglish grammars.
\newblock In {\em Proceedings of a Workshop---Speech and Natural Language},
  pages 306--311.

\bibitem[Bloomfield, 1933]{Bloomfield:L-33}
Bloomfield, L. (1933).
\newblock {\em Language}.
\newblock George Allen \& Unwin Limited, London, UK, reprinted 1970 edition.

\bibitem[Bod, 1995]{Bod:ELS-95}
Bod, R. (1995).
\newblock {\em Enriching Linguistics with Statistics: Performance Models of
  Natural Language}.
\newblock PhD thesis, University of Amsterdam, Amsterdam, the Netherlands.

\bibitem[Bod, 1998]{Bod:BG-98}
Bod, R. (1998).
\newblock {\em Beyond Grammar---An Experience-Based Theory of Language},
  volume~88 of {\em {CSLI} Lecture Notes}.
\newblock Center for Study of Language and Information (CSLI) Publications,
  Stanford:CA, USA.

\bibitem[Bod, 2000]{Bod:COLING00-69}
Bod, R. (2000).
\newblock Parsing with the shortest derivation.
\newblock In \cite{COLING00}, pages 69--75.

\bibitem[Bonnema et~al., 1997]{Bonnema:ACL97-159}
Bonnema, R., Bod, R., and Scha, R. (1997).
\newblock A {DOP} model for semantic interpretation.
\newblock In \cite{ACL97}, pages 159--167.

\bibitem[Booth, 1969]{Booth:ASSA69-74}
Booth, T. (1969).
\newblock Probabilistic representation of formal languages.
\newblock In {\em Conference Record of 1969 Tenth Annual Symposium on Switching
  and Automata Theory}, pages 74--81.

\bibitem[Brent, 1999]{Brent:99-34-71}
Brent, M.~R. (1999).
\newblock An efficient, probabilistically sound algorithm for segmentation and
  word discovery.
\newblock {\em Machine Learning}, 34:71--105.

\bibitem[Brill, 1993]{Brill:ACL93-259}
Brill, E. (1993).
\newblock Automatic grammar induction and parsing free text: A
  transformation-based approach.
\newblock In {\em Proceedings of the 31th Annual Meeting of the {}Association
  for Computational Linguistics ({ACL}){}; Columbus:OH, USA}, pages 259--265.
  Association for Computational Linguistics ({ACL}).

\bibitem[Caraballo and Charniak, 1998]{Caraballo:98-24-275}
Caraballo, S.~A. and Charniak, E. (1998).
\newblock New figures of merit for best-first probabilistic chart parsing.
\newblock {\em Computational Linguistics}, 24(2):275--298.

\bibitem[Carroll, 1982]{Carroll:CIW-82-17}
Carroll, L. (1982).
\newblock Alice's adventures in wonderland.
\newblock In {\em The Complete Illustrated Works of Lewis Carroll}, pages
  17--114. Chancellor Press, London, UK, 1st (1993) edition.
\newblock First published in 1865.

\bibitem[Charniak, 1993]{Charniak:SLL-93}
Charniak, E. (1993).
\newblock {\em Statistical Language Learning}.
\newblock Massachusetts Institute of Technology Press, Cambridge:MA, USA and
  London, UK.

\bibitem[Charniak, 1997]{Charniak:NCAI97-598}
Charniak, E. (1997).
\newblock Statistical parsing with a context-free grammar and word statistics.
\newblock In {\em Proceedings of the Fourteenth National Conference on
  Artificial Intelligence}, pages 598--603. American Association for Artificial
  Intelligence ({AAAI}).

\bibitem[Chen, 1995]{Chen:ACL95-228}
Chen, S.~F. (1995).
\newblock Bayesian grammar induction for language modeling.
\newblock In \cite{ACL95}, pages 228--235.

\bibitem[Chen and Goodman, 1996]{Chen:ACL96-??}
Chen, S.~F. and Goodman, J. (1996).
\newblock An empirical study of smoothing techniques for language modeling.
\newblock In {\em Proceedings of the 34th Annual Meeting of the {}Association
  for Computational Linguistics ({ACL}){}; Santa Cruz:CA, USA}. Association for
  Computational Linguistics ({ACL}).

\bibitem[Chen and Goodman, 1998]{Chen:ESS-98}
Chen, S.~F. and Goodman, J. (1998).
\newblock An empirical study of smoothing techniques for language modeling.
\newblock Technical Report {TR}-10-98, Harvard University, Cambridge:MA, USA.

\bibitem[Chomsky, 1955]{Chomsky:LSLT-75}
Chomsky, N. (1955).
\newblock {\em The Logical Structure of Linguistic Theory}.
\newblock Plenum Press, New York:NY, USA, reprinted 1975 edition.

\bibitem[Chomsky, 1965]{Chomsky:ATS-65}
Chomsky, N. (1965).
\newblock {\em Aspects of the Theory of Syntax}.
\newblock Massachusetts Institute of Technology Press, Cambridge:MA, USA and
  London, UK.
\newblock 3rd paperback printing.

\bibitem[Chomsky, 1986]{Chomsky:KL-86}
Chomsky, N. (1986).
\newblock {\em Knowledge of Language --- Its Nature, Origin, and Use}.
\newblock Praeger Publishers, New York:NY, USA.

\bibitem[Clark, 2001a]{Clark:ACL/Student01-55}
Clark, A. (2001a).
\newblock Learning morphology with pair hidden markov models.
\newblock In {\em Proceedings of the Student Research Workshop of the 39th
  Annual Meeting of the {}Association for Computational Linguistics ({ACL}){}
  and the 10th Meeting of the {}European Chapter of the Association for
  Computational Linguistics ({EACL}){}; Toulouse, France}, pages 55--60.
  Association for Computational Linguistics ({ACL}).

\bibitem[Clark, 2001b]{Clark:ACL/CNLL01-105}
Clark, A. (2001b).
\newblock Unsupervised induction of stochastic context-free grammars using
  distributional clustering.
\newblock In \cite{ACL/CNLL01}, pages 105--112.

\bibitem[{COLING}, 2000]{COLING00}
{COLING} (2000).
\newblock {\em Proceedings of the 18th International Conference on
  Computational Linguistics ({COLING}); Saarbr{\"u}cken, Germany}. Association
  for Computational Linguistics ({ACL}).

\bibitem[Collins, 1997]{Collins:ACL97-16}
Collins, M. (1997).
\newblock Three generative, lexicalised models for statistical parsing.
\newblock In \cite{ACL97}, pages 16--23.

\bibitem[{CoNLL}, 2001]{ACL/CNLL01}
{CoNLL} (2001).
\newblock {\em Proceedings of the Workshop on Computational Natural Language
  Learning held at the 39th Annual Meeting of the {}Association for
  Computational Linguistics ({ACL}){} and the 10th Meeting of the {}European
  Chapter of the Association for Computational Linguistics ({EACL}){};
  Toulouse, France}. Association for Computational Linguistics ({ACL}).

\bibitem[Cook et~al., 1976]{Cook:76-10-59}
Cook, C.~M., Rosenfeld, A., and Aronson, A.~R. (1976).
\newblock Grammatical inference by hill climbing.
\newblock {\em Informational Sciences}, 10:59--80.

\bibitem[Cook and Holder, 1994]{Cook:94-1-231}
Cook, D.~J. and Holder, L.~B. (1994).
\newblock Substructure discovery using minimum description length and
  background knowledge.
\newblock {\em Journal of Artificial Intelligence Research}, 1:231--255.

\bibitem[de~Marcken, 1995]{deMarcken:ACL/Student95-??}
de~Marcken, C. (1995).
\newblock Acquiring a lexicon from unsegmented speech.
\newblock In {\em Proceedings of the Student Session of the 33th Annual Meeting
  of the {}Association for Computational Linguistics ({ACL})}. Association for
  Computational Linguistics ({ACL}).

\bibitem[de~Marcken, 1999]{deMarcken:UAL-99}
de~Marcken, C. (1999).
\newblock The unsupervised acquisition of a lexicon from continuous speech.
\newblock Technical Report AI Memo 1558, CBCL Memo 129, Massachusetts Institute
  of Technology, {}Artificial Intelligence Laboratory, {}Center for Biological
  and Computational Learning Department of Brain and Cognitive Sciences,
  Cambridge:MA, USA.

\bibitem[de~Marcken, 1996]{deMarcken:ULA-96}
de~Marcken, C.~G. (1996).
\newblock {\em Unsupervised Language Acquisition}.
\newblock PhD thesis, Massachusetts Institute of Technology, Cambridge:MA, USA.

\bibitem[D{\'e}jean, 2000]{Dejean:CONLL00-95}
D{\'e}jean, H. (2000).
\newblock {ALLiS}: a symbolic learning system for natural language learning.
\newblock In Cardie, C., Daelemans, W., N{\'e}dellec, C., and Tjong Kim~Sang,
  E., editors, {\em Proceedings of the Fourth Conference on Computational
  Natural Language Learning and of the Second Learning Language in Logic
  Workshop; Lisbon, Portugal}, pages 95--98.
\newblock Held in cooperation with ICGI-2000.

\bibitem[D{\"o}rnenburg, 1997]{Dornenburg:EEA-97}
D{\"o}rnenburg, E. (1997).
\newblock Extensions of the {EMILE} algorithm for inductive learning of
  context-free grammars.
\newblock Master's thesis, University of Dortmund, Dortmund, Germany.

\bibitem[Duda and Hart, 1973]{Duda:PCSA-73}
Duda, R.~O. and Hart, P.~E. (1973).
\newblock {\em Pattern Classification and Scene Analysis}.
\newblock John Wiley and Sons, New York:NY, USA.

\bibitem[Durbin, 2001]{Durbin:ACL01-1}
Durbin, R. (2001).
\newblock Interpreting the human genome sequence, using stochastic grammars.
\newblock In {\em Proceedings of the 39th Annual Meeting of the {}Association
  for Computational Linguistics ({ACL}){} and the 10th Meeting of the
  {}European Chapter of the Association for Computational Linguistics
  ({EACL}){}; Toulouse, France}. Association for Computational Linguistics
  ({ACL}).
\newblock Invited talk.

\bibitem[Finch and Chater, 1992]{Finch:SHOE92-230}
Finch, S. and Chater, N. (1992).
\newblock Bootstrapping syntactic categories using statistical methods.
\newblock In Daelemans, W. and Powers, D., editors, {\em Background and
  Experiments in Machine Learning of Natural Language: Proceedings First {SHOE}
  Workshop}, pages 230--235, Tilburg, the Netherlands. Institute for Language
  Technology and {AI} Tilburg University.

\bibitem[Frayn, 1965]{Frayn:TM-65}
Frayn, M. (1965).
\newblock {\em The Tin Men}.
\newblock Collins Fontana Books.

\bibitem[Frege, 1879]{Frege:BS-79}
Frege, G. (1879).
\newblock {\em Begriffsschrift, eine der {A}rithmetischen {N}achgebildete
  {F}ormelsprache des {R}eines {D}enkens}.
\newblock {V}erlag von Louis Nebert, Halle, Germany.

\bibitem[Gaussier, 1999]{Gaussier:ULNLP99-??}
Gaussier, E. (1999).
\newblock Unsupervised learning of derivational morphology from inflectional
  lexicons.
\newblock In \cite{ULNLP99}.

\bibitem[Gold, 1967]{Gold:67-10-447}
Gold, E. (1967).
\newblock Language identification in the limit.
\newblock {\em Information and Control}, 10:447--474.

\bibitem[Goldsmith, 2001]{Goldsmith:27-2-153}
Goldsmith, J. (2001).
\newblock Unsupervised learning of the morphology of a natural language.
\newblock {\em Computational Linguistics}, 27(2):153--198.

\bibitem[Goossens et~al., 1994]{Goossens:LC-94}
Goossens, M., Mittelbach, F., and Samarin, A. (1994).
\newblock {\em The {\LaTeX} Companion}.
\newblock Addison-Wesley Publishing Company, Reading:MA, USA.

\bibitem[Goossens et~al., 1997]{Goossens:LGC-97}
Goossens, M., Rahtz, S., and Mittelbach, F. (1997).
\newblock {\em The {\LaTeX} Graphics Companion}.
\newblock Addison-Wesley Publishing Company, Reading:MA, USA.

\bibitem[Grune and Jacobs, 1990]{Grune:PT-90}
Grune, D. and Jacobs, C. (1990).
\newblock {\em Parsing Techniques---A Practical Guide}.
\newblock Ellis Horwood Limited, Chichester, UK.
\newblock Printout by the authors.

\bibitem[Gr{\"u}nwald, 1994]{Grunwald:CSS-94-203}
Gr{\"u}nwald, P. (1994).
\newblock A minimum description length approach to grammar inference.
\newblock In Scheler, G., Wernter, S., and Riloff, E., editors, {\em
  Connectionist, Statistical and Symbolic Approaches to Learning for Natural
  Language}, volume 1004 of {\em Lecture Notes in {AI}}, pages 203--216.
  Springer-Verlag, Berlin Heidelberg, Germany.

\bibitem[Gusfield, 1997]{Gusfield:AOSTAS-97}
Gusfield, D. (1997).
\newblock {\em Algorithms on strings, trees, and sequences: computer science
  and computational biology}.
\newblock Cambridge University Press, Cambridge, UK.
\newblock Reprinted 1999 (with corrections).

\bibitem[Harris, 1951]{Harris:SL-51}
Harris, Z.~S. (1951).
\newblock {\em Structural Linguistics}.
\newblock University of Chicago Press, Chicago:IL, USA and London, UK, 7th
  (1966) edition.
\newblock Formerly Entitled: Methods in Structural Linguistics.

\bibitem[Honkela et~al., 1995]{Honkela:IANN95-3}
Honkela, T., Pulkki, V., and Kohonen, T. (1995).
\newblock Contextual relations of words in grimm tales, analyzed by
  self-organizing map.
\newblock In Fogerlman-Soulie, F. and Gallinari, P., editors, {\em Proceedings
  of the International Conference on Artificial Neural Networks; Paris,
  France}, pages 3--7.

\bibitem[Horning, 1969]{Horning:SGI-69}
Horning, J.~J. (1969).
\newblock {\em A study of grammatical inference}.
\newblock PhD thesis, Stanford University, Stanford:CA, USA.

\bibitem[Hubbard et~al., 1996]{Hubbard:96-4-313}
Hubbard, T. J.~P., Lesk, A.~M., and Tramontano, A. (1996).
\newblock Gathering them into the fold.
\newblock {\em Nature Structural Biology}, 3(4):313.

\bibitem[Huckle, 1995]{Huckle:ACL95-??}
Huckle, C.~C. (1995).
\newblock Grouping words using statistical context.
\newblock In \cite{ACL95}.

\bibitem[Hwa, 1999]{Hwa:ACL99-73}
Hwa, R. (1999).
\newblock Supervised grammar induction using training data with limited
  constituent information.
\newblock In {\em Proceedings of the 37th Annual Meeting of the {}Association
  for Computational Linguistics ({ACL}){}; Maryland:MD, USA}, pages 73--79.
  Association for Computational Linguistics ({ACL}).

\bibitem[Johnson, 1979]{Johnson:UPM-79}
Johnson, S.~C. (1979).
\newblock Yacc: Yet another compiler-compiler.
\newblock In {\em {UNIX} Programmer's Manual}, pages 353--387. Holt, Rinehart,
  and Winston, New York:NY, USA.
\newblock Vol. 2.

\bibitem[Jurafsky and Martin, 2000]{Jurafsky:SLP-00}
Jurafsky, D. and Martin, J.~H. (2000).
\newblock {\em Speech and Language Processing}.
\newblock Prentice Hall, Englewood Cliffs:NJ, USA.

\bibitem[Kay and Kummerfeld, 1989]{Kay:CPU-89}
Kay, J. and Kummerfeld, B. (1989).
\newblock {\em C Programming in a {UNIX} Environment}.
\newblock International Computer Science Series. Addison-Wesley Publishing
  Company, Reading:MA, USA.

\bibitem[Kehler and Stolcke, 1999a]{Kehler:ULNLP99-0}
Kehler, A. and Stolcke, A. (1999a).
\newblock Preface.
\newblock In \cite{ULNLP99}.

\bibitem[Kehler and Stolcke, 1999b]{ULNLP99}
Kehler, A. and Stolcke, A., editors (1999b).
\newblock {\em Proceedings of a Workshop---Unsupervised Learning in Natural
  Language Processing; Maryland:MD, USA}.

\bibitem[Kernighan and Ritchie, 1988]{Kernighan:CPL-88}
Kernighan, B.~W. and Ritchie, D.~M. (1988).
\newblock {\em The C Programming Language}.
\newblock Prentice Hall Software Series. Prentice Hall, Englewood Cliffs:NJ,
  USA, 2nd edition.
\newblock {ANSI} {C}.

\bibitem[Klein and Manning, 2001]{Klein:ACL/CNLL01-113}
Klein, D. and Manning, C.~D. (2001).
\newblock Distributional phrase structure indiction.
\newblock In \cite{ACL/CNLL01}, pages 113--120.

\bibitem[Knight and Yamada, 1999]{Knight:ULNLP99-37}
Knight, K. and Yamada, K. (1999).
\newblock A computational approach to deciphering unknown scripts.
\newblock In \cite{ULNLP99}, pages 37--44.

\bibitem[Knuth, 1986]{Knuth:TT-86}
Knuth, D.~E. (1986).
\newblock {\em The \TeX book}.
\newblock Addison-Wesley Publishing Company, Reading:MA, USA.
\newblock Reprinted with corrections 1993.

\bibitem[Knuth, 1996]{Knuth:SPCS-96-??}
Knuth, D.~E. (1996).
\newblock Are toy problems useful?
\newblock In {\em Selected Papers in Computer Science}. Center for Study of
  Language and Information (CSLI) Publications, Stanford:CA, USA.

\bibitem[Lamport, 1994]{Lamport:LADPS-94}
Lamport, L. (1994).
\newblock {\em {\LaTeX}: A Document Preparation System}.
\newblock Addison-Wesley Publishing Company, Reading:MA, USA, 2nd edition.

\bibitem[Lari and Young, 1990]{Lari:90-4-35}
Lari, K. and Young, S. (1990).
\newblock The estimation of stochastic context-free grammars using the
  inside-outside algorithm.
\newblock {\em Computer Speech and Language}, 4(35--56).

\bibitem[Lee, 1996]{Lee:LCFL-96}
Lee, L. (1996).
\newblock Learning of context-free languages: A survey of the literature.
\newblock Technical Report TR-12-96, Harvard University, Cambridge:MA, USA.

\bibitem[Lerdahl and Jackendoff, 1983]{Lerdahl:GTTM-83}
Lerdahl, F. and Jackendoff, R.~S. (1983).
\newblock {\em A generative theory of tonal music}.
\newblock Massachusetts Institute of Technology Press, Cambridge:MA, USA and
  London, UK.

\bibitem[Levenshtein, 1965]{Levenshtein:65-163-845}
Levenshtein, V.~I. (1965).
\newblock Binary codes capable of correcting deletions, insertions, and
  reversals.
\newblock {\em Doklady Akademii Nauk {SSR}}, 163(4):845--848.
\newblock Original in Russian.

\bibitem[Li and Vit{\'a}nyi, 1991]{Li:91-5-911}
Li, M. and Vit{\'a}nyi, P. M.~B. (1991).
\newblock Learning simple concepts under simple distributions.
\newblock {\em SIAM Journal of Computing}, 20(5):911--935.

\bibitem[Lippman, 1991]{Lippman:CP-91}
Lippman, S.~B. (1991).
\newblock {\em {C}++ Primer}.
\newblock Addison-Wesley Publishing Company, Reading:MA, USA, 2nd edition.
\newblock Reprinted with corrections {}February{} 1993.

\bibitem[Longman, 1995]{Longman:LDCE-95}
Longman (1995).
\newblock {\em Longman Dictionary of Contemporary {E}nglish}.
\newblock Longman Group Ltd, Essex, UK, 3rd edition.

\bibitem[Losee, 1996]{Losee:95-??-??}
Losee, R.~M. (1996).
\newblock Learning syntactic rules and tags with genetic algorithm for
  information retrieval and filtering: An empirical basis for grammatical
  rules.
\newblock {\em Information Processing \& Management}, 32(2):185--197.

\bibitem[Magerman and Marcus, 1990]{Magerman:AAAI90-984}
Magerman, D.~M. and Marcus, M.~P. (1990).
\newblock Parsing a natural language using mutual information statistics.
\newblock In {\em Proceedings of the Eighth National Conference on Artificial
  Intelligence}, pages 984--989. American Association for Artificial
  Intelligence ({AAAI}).

\bibitem[Marcus et~al., 1993]{Marcus:93-19-313}
Marcus, M., Santorini, B., and Marcinkiewicz, M. (1993).
\newblock Building a large annotated corpus of {E}nglish: the {P}enn treebank.
\newblock {\em Computational Linguistics}, 19(2):313--330.

\bibitem[Montague, 1974]{Montague:FP-74-247}
Montague, R. (1974).
\newblock The proper treatment of quantification in ordinary {E}nglish.
\newblock In Thomason, R.~H., editor, {\em Formal Philosophy --- Selected
  Papers of Richard Montague}, chapter~8, pages 247--270. Yale University
  Press, New Haven:CT, USA and London, UK.

\bibitem[Nakamura and Ishiwata, 2000]{Nakamura:ICGI00-186}
Nakamura, K. and Ishiwata, T. (2000).
\newblock Synthesizing context free grammars from sample strings based on
  inductive {CYK} algorithm.
\newblock In \cite{ICGI00}, pages 186--195.

\bibitem[Nevado et~al., 2000]{Nevado:ICGI00-196}
Nevado, F., S{\'a}nchez, J.-A., and Bened{\'\i}, J.-M. (2000).
\newblock Combination of estimation algorithms and grammatical inference
  techniques to learn stochastic context-free grammars.
\newblock In \cite{ICGI00}, pages 196--206.

\bibitem[Nevill-Manning and Witten, 1997]{Nevill-Manning:97-7-67}
Nevill-Manning, C.~G. and Witten, I.~H. (1997).
\newblock Identifying hierarchical structure in sequences: A linear-time
  algorithm.
\newblock {\em Journal of Artificial Intelligence Research}, 7:67--82.

\bibitem[Oliveira, 2000]{ICGI00}
Oliveira, A.~L., editor (2000).
\newblock {\em Grammatical Inference: Algorithms and Applications ({ICGI});
  Lisbon, Portugal}.

\bibitem[Osborne, 1994]{Osborne:LUB-94}
Osborne, M. (1994).
\newblock {\em Learning Unification-Based Natural Language Grammars}.
\newblock PhD thesis, University of York, York, UK.

\bibitem[Pereira and Schabes, 1992]{Pereira:ACL92-128}
Pereira, F. and Schabes, Y. (1992).
\newblock Inside-outside reestimation from partially bracketed corpora.
\newblock In {\em Proceedings of the 30th Annual Meeting of the {}Association
  for Computational Linguistics ({ACL}){}; Newark:NJ, USA}, pages 128--135.
  Association for Computational Linguistics ({ACL}).

\bibitem[Pinker, 1984]{Pinker:LLLD-84}
Pinker, S. (1984).
\newblock {\em Language learnability and language development}.
\newblock Harvard University Press, Cambridge:MA.

\bibitem[Pinker, 1994]{Pinker:LI-94}
Pinker, S. (1994).
\newblock {\em The Language Instinct---The New Science of Language and Mind}.
\newblock Penguin Books, Harmondsworth, Middlesex, UK.

\bibitem[Poutsma, 2000a]{Poutsma:COLING00-635}
Poutsma, A. (2000a).
\newblock {D}ata-{O}riented {T}ranslation.
\newblock In \cite{COLING00}, pages 635--641.

\bibitem[Poutsma, 2000b]{Poutsma:DOT-00}
Poutsma, A. (2000b).
\newblock {D}ata-{O}riented {T}ranslation---using the {D}ata-{O}riented
  {P}arsing framework for machine translation.
\newblock Master's thesis, University of Amsterdam, Amsterdam, the Netherlands.

\bibitem[Powers, 1997]{MLNL97}
Powers, D.~M., editor (1997).
\newblock {\em Tutorial Notes---Machine Learning of Natural Language; Madrid,
  Spain}. Association for Computational Linguistics ({ACL}).

\bibitem[Redington et~al., 1998]{Redington:98-22-425}
Redington, M., Chater, N., and Finch, S. (1998).
\newblock Distributional information: A powerful cue for acquiring syntactic
  categories.
\newblock {\em Cognitive Science}, 22(4):425--469.

\bibitem[Russell and Norvig, 1995]{Russell:AI-95}
Russell, S. and Norvig, P. (1995).
\newblock {\em Artificial intelligence: a modern approach}.
\newblock Prentice Hall, Englewood Cliffs:NJ, USA.

\bibitem[Sadler and Vendelmans, 1990]{Sadler:COLING90-449}
Sadler, V. and Vendelmans, R. (1990).
\newblock Pilot implementation of a bilingual knowledge bank.
\newblock In {\em Proceedings of the 13th International Conference on
  Computational Linguistics ({COLING}); Helsinki, Finland}, pages 449--451.
  Association for Computational Linguistics ({ACL}).

\bibitem[Sakakibara, 1992]{Sakakibara:92-97-23}
Sakakibara, Y. (1992).
\newblock Efficient learning of context-free grammars from positive structural
  examples.
\newblock {\em Information and Computation}, 97:23--60.

\bibitem[Sakakibara and Muramatsu, 2000]{Sakakibara:ICGI00-229}
Sakakibara, Y. and Muramatsu, H. (2000).
\newblock Learning context-free grammars from partially structured examples.
\newblock In \cite{ICGI00}, pages 229--240.

\bibitem[Sampson, 1995]{Sampson:EC-95}
Sampson, G. (1995).
\newblock {\em {E}nglish for the Computer --- The {SUSANNE} Corpus and Analytic
  Scheme}.
\newblock Clarendon Press (Oxford University Press), New York:NY, USA.

\bibitem[Sampson, 1997]{Sampson:EE-97}
Sampson, G. (1997).
\newblock {\em Educating Eve---The `Language Instinct' Debate}.
\newblock Cassell, London, UK and New York:NY, USA.
\newblock Reprinted in paperback with minor changes 1999.

\bibitem[Sampson, 2000]{Sampson:00-5-53}
Sampson, G. (2000).
\newblock A proposal for improving the measurement of parse accuracy.
\newblock {\em International Journal of Corpus Linguistics}, 5(1):53--68.

\bibitem[Sankoff and Kruskal, 1999]{Sankoff:TWS-99}
Sankoff, D. and Kruskal, J. (1999).
\newblock {\em Time Warps, String Edits, and Macromolecules---The Theory and
  Practice of Sequence Comparison}.
\newblock The David Hume Series (Philosophy and Cognitive Science Reissues).
  Center for Study of Language and Information (CSLI) Publications,
  Stanford:CA, USA.

\bibitem[Schaffrath, 1995]{Schaffrath:EFC-95}
Schaffrath, H. (1995).
\newblock The {E}ssen folksong collection in the humdrum kern format.
\newblock D. Huron (ed.). Menlo Park:CA, USA. Center for Computer Assisted
  Research in the Humanities.

\bibitem[Scholtes and Bloembergen, 1992]{Scholtes:IJCNN92-??}
Scholtes, J.~C. and Bloembergen, S. (1992).
\newblock Corpus based parsing with a self-organizing neural net.
\newblock In {\em Proceedings of the International Joint Conference on Neural
  Networks ({IJCNN}); Beijing, P.R. China}.

\bibitem[Searls, 1994]{CLBS94}
Searls, D., editor (1994).
\newblock {\em The Computational Linguistics of Biological Sequences (ACL'94
  Tutorial Notes)}. Association for Computational Linguistics ({ACL}).

\bibitem[Sima'an, 1999]{Simaan:LED-99}
Sima'an, K. (1999).
\newblock {\em Learning Efficient Disambiguation}.
\newblock PhD thesis, Universteit Utrecht, Utrecht, the Netherlands.

\bibitem[Simon, 1969]{Simon:SA-69}
Simon, H.~A. (1969).
\newblock {\em The Sciences of the Artificial}.
\newblock Massachusetts Institute of Technology Press, Cambridge:MA, USA and
  London, UK, 1st edition.

\bibitem[Sloboda, 1985]{Sloboda:MM-85}
Sloboda, J.~A. (1985).
\newblock {\em The Musical Mind}, volume~5 of {\em Oxford Psychology Series}.
\newblock Oxford University Press, New York:NY, USA.

\bibitem[Stolcke, 1994]{Stolcke:BLP-94}
Stolcke, A. (1994).
\newblock {\em Bayesian Learning of Probabilistic Language Models}.
\newblock PhD thesis, University of California, Berkeley:CA, USA.

\bibitem[Stolcke and Omohundro, 1994]{Stolcke:ICGI94-106}
Stolcke, A. and Omohundro, S. (1994).
\newblock Inducing probabilistic grammars by bayesian model merging.
\newblock In {\em Proceedings of the Second International Conference on Grammar
  Inference and Applications; Alicante, Spain}, pages 106--118.

\bibitem[Stroustrup, 1997]{Stroustrup:CPL-97}
Stroustrup, B. (1997).
\newblock {\em The {C++} Programming Language}.
\newblock Addison-Wesley Publishing Company, Reading:MA, USA, 3rd edition.

\bibitem[Valiant, 1984]{Valiant:84-27-1134}
Valiant, L. (1984).
\newblock A theory of the learnable.
\newblock {\em Communications of the Association for Computing Machinery},
  27(11):1134--1142.

\bibitem[van Zaanen, 1997]{vanZaanen:ECD-97}
van Zaanen, M. (1997).
\newblock Error correction using {DOP}.
\newblock Master's thesis, Vrije Universiteit, Amsterdam, the Netherlands.

\bibitem[van Zaanen, 1999a]{vanZaanen:CLIN99-235}
van Zaanen, M. (1999a).
\newblock Bootstrapping structure using similarity.
\newblock In Monachesi, P., editor, {\em Computational Linguistics in the
  Netherlands 1999---Selected Papers from the Tenth {CLIN} Meeting}, pages
  235--245, Utrecht, the Netherlands. Universteit Utrecht.

\bibitem[van Zaanen, 1999b]{vanZaanen:CLUK99-1}
van Zaanen, M. (1999b).
\newblock Error correction using {DOP}.
\newblock In \uppercase{d}e Roeck, A., editor, {\em Proceedings of the Second
  UK Special Interest Group for Computational Linguistics ({CLUK2}) (Second
  Issue)}, pages 1--12, Colchester, UK. University of Essex.

\bibitem[van Zaanen, 2000a]{vanZaanen:COLING00-961}
van Zaanen, M. (2000a).
\newblock {ABL}: {A}lignment-{B}ased {L}earning.
\newblock In \cite{COLING00}, pages 961--967.

\bibitem[van Zaanen, 2000b]{vanZaanen:ICML00-1063}
van Zaanen, M. (2000b).
\newblock Bootstrapping syntax and recursion using {A}lignment-{B}ased
  {L}earning.
\newblock In Langley, P., editor, {\em Proceedings of the Seventeenth
  International Conference on Machine Learning}, pages 1063--1070, Stanford:CA,
  USA. Stanford University.

\bibitem[van Zaanen, 2000c]{vanZaanen:CLUK00-75}
van Zaanen, M. (2000c).
\newblock Learning structure using {A}lignment {B}ased {L}earning.
\newblock In Kilgarriff, A., Pearce, D., and Tiberius, C., editors, {\em
  Proceedings of the Third Annual Doctoral Research Colloquium ({CLUK})}, pages
  75--82. Universities of Brighton and Sussex.

\bibitem[van Zaanen, 2001]{vanZaanen:BTG-01}
van Zaanen, M. (2001).
\newblock Building treebanks using a grammar induction system.
\newblock Technical Report TR2001.06, University of Leeds, Leeds, UK.

\bibitem[van Zaanen, 2002]{vanZaanen:DOP02-??}
van Zaanen, M. (2002).
\newblock {A}lignment-{B}ased {L}earning versus {D}ata-{O}riented {P}arsing.
\newblock In Bod, R., Sima'an, K., and Scha, R., editors, {\em {D}ata
  {O}riented {P}arsing}. Center for Study of Language and Information (CSLI)
  Publications, Stanford:CA, USA.
\newblock to be published.

\bibitem[van Zaanen and Adriaans, 2001a]{vanZaanen:BNAIC01-??}
van Zaanen, M. and Adriaans, P. (2001a).
\newblock {A}lignment-{B}ased {L}earning versus {EMILE}: A comparison.
\newblock In {\em Proceedings of the Belgian-Dutch Conference on Artificial
  Intelligence ({BNAIC}); Amsterdam, the Netherlands}.
\newblock to be published.

\bibitem[van Zaanen and Adriaans, 2001b]{vanZaanen:CTU-01}
van Zaanen, M. and Adriaans, P. (2001b).
\newblock Comparing two unsupervised grammar induction systems:
  {A}lignment-{B}ased {L}earning vs. {EMILE}.
\newblock Technical Report TR2001.05, University of Leeds, Leeds, UK.

\bibitem[Vervoort, 2000]{Vervoort:GWG-00}
Vervoort, M.~R. (2000).
\newblock {\em Games, Walks and Grammars}.
\newblock PhD thesis, University of Amsterdam, Amsterdam, the Netherlands.

\bibitem[Viterbi, 1967]{Viterbi:67-13-260}
Viterbi, A. (1967).
\newblock Error bounds for convolutional codes and an asymptotically optimum
  decoding algorithm.
\newblock {\em Institute of Electrical and Electronics Engineers Transactions
  on Information Theory}, 13:260--269.

\bibitem[Wagner and Fischer, 1974]{Wagner:74-21-168}
Wagner, R.~A. and Fischer, M.~J. (1974).
\newblock The string-to-string correction problem.
\newblock {\em Journal of the Association for Computing Machinery},
  21(1):168--173.

\bibitem[Way, 1999]{Way:99-11-??}
Way, A. (1999).
\newblock A hybrid architecture for robust {MT} using {LFG-DOP}.
\newblock {\em Journal of Experimental and Theoretical Artificial
  Intelligence}, 11(4).
\newblock Special Issue on Memory-Based Language Processing.

\bibitem[Wolff, 1975]{Wolff:75-66-79}
Wolff, J. (1975).
\newblock An algorithm for the segmentation of an artificial language analogue.
\newblock {\em British Journal of Psychology}, 66(1):79--90.

\bibitem[Wolff, 1977]{Wolff:77-68-97}
Wolff, J. (1977).
\newblock The discovery of segments in natural language.
\newblock {\em British Journal of Psychology}, 68:97--106.

\bibitem[Wolff, 1996]{Wolff:CLP-96-67}
Wolff, J. (1996).
\newblock Learning and reasoning as information compression by multiple
  alignment, unification and search.
\newblock In Gammerman, A., editor, {\em Computational Learning and
  Probabilistic Reasoning}, chapter~4, pages 67--85. John Wiley \& Sons, Ltd.,
  Chichester, UK.

\bibitem[Wolff, 1980]{Wolff:80-23-255}
Wolff, J.~G. (1980).
\newblock Language acquisition and the discovery of phrase structure.
\newblock {\em Language and Speech}, 23(3):255--269.

\bibitem[Wolff, 1982]{Wolff:82-2-57}
Wolff, J.~G. (1982).
\newblock Language acquisition, data compression and generalization.
\newblock {\em Language and Communication}, 2(1):57--89.

\bibitem[Wolff, 1988]{Wolff:CPL-88-179}
Wolff, J.~G. (1988).
\newblock Learning syntax and meanings through optimization and distributional
  analysis.
\newblock In Levy, Y., Schlesinger, I., and Braine, M., editors, {\em
  Categories and Processes in Language Acquisition}, chapter~7, pages 179--215.
  Lawrence Erlbaum, Hillsdale:NJ, USA.

\bibitem[Wolff, 1998a]{Wolff:PICE-98}
Wolff, J.~G. (1998a).
\newblock Parsing as information compresion by multiple alignment, unification
  and search: Examples.
\newblock Technical report, University of Wales, Bangor, UK.

\bibitem[Wolff, 1998b]{Wolff:PICS-98}
Wolff, J.~G. (1998b).
\newblock Parsing as information compresion by multiple alignment, unification
  and search: {SP}52.
\newblock Technical report, University of Wales, Bangor, UK.

\bibitem[Younger, 1967]{Younger:67-10-189}
Younger, D. (1967).
\newblock Recognition and parsing of context-free languages in time $n^3$.
\newblock {\em Information and Control}, 10(2):189--208.

\end{thebibliography}
